\documentclass[journal]{IEEEtran}
\usepackage[utf8]{inputenc}
\usepackage{amsmath,amssymb,amsfonts,amsthm}
\usepackage{graphicx}
\usepackage{booktabs}
\usepackage{array}
\usepackage{multirow}
\usepackage{algorithm}
\usepackage{algorithmic}
\usepackage{cite}
\usepackage{url}
\usepackage[hidelinks]{hyperref}
\newcommand{\ALG}{MF-EDA}
\newcommand{\AMF}{AMaLGaM-MF}
\newcommand{\vect}[1]{\mathbf{#1}}
\newcommand{\mat}[1]{\mathbf{#1}}

\newtheorem{proposition}{Proposition}

\makeatletter\newcommand{\nm}[1]{\csname n@#1\endcsname}\makeatother
\expandafter\def\csname n@A_G1_D10_ama\endcsname{50}
\expandafter\def\csname n@A_G1_D10_ama_ert\endcsname{6.6k}
\expandafter\def\csname n@A_G1_D10_ama_fin\endcsname{6.8e-11}
\expandafter\def\csname n@A_G1_D10_amf\endcsname{50}
\expandafter\def\csname n@A_G1_D10_amf2\endcsname{50}
\expandafter\def\csname n@A_G1_D10_amf2_ert\endcsname{6.6k}
\expandafter\def\csname n@A_G1_D10_amf2_fin\endcsname{7.2e-11}
\expandafter\def\csname n@A_G1_D10_amf_ert\endcsname{6.8k}
\expandafter\def\csname n@A_G1_D10_amf_fin\endcsname{8.2e-11}
\expandafter\def\csname n@A_G1_D10_bipop\endcsname{50}
\expandafter\def\csname n@A_G1_D10_bipop_ert\endcsname{1.5k}
\expandafter\def\csname n@A_G1_D10_bipop_fin\endcsname{8.6e-11}
\expandafter\def\csname n@A_G1_D10_cam\endcsname{50}
\expandafter\def\csname n@A_G1_D10_cam_ert\endcsname{7.3k}
\expandafter\def\csname n@A_G1_D10_cam_fin\endcsname{8e-11}
\expandafter\def\csname n@A_G1_D10_cma\endcsname{50}
\expandafter\def\csname n@A_G1_D10_cma_ert\endcsname{1.5k}
\expandafter\def\csname n@A_G1_D10_cma_fin\endcsname{8e-11}
\expandafter\def\csname n@A_G1_D10_curg\endcsname{50}
\expandafter\def\csname n@A_G1_D10_curg_ert\endcsname{12.1k}
\expandafter\def\csname n@A_G1_D10_curg_fin\endcsname{8.1e-11}
\expandafter\def\csname n@A_G1_D10_hv0\endcsname{??}
\expandafter\def\csname n@A_G1_D10_hv0_ert\endcsname{??}
\expandafter\def\csname n@A_G1_D10_hv1\endcsname{??}
\expandafter\def\csname n@A_G1_D10_hv1_ert\endcsname{??}
\expandafter\def\csname n@A_G1_D10_ipop\endcsname{50}
\expandafter\def\csname n@A_G1_D10_ipop_ert\endcsname{1.5k}
\expandafter\def\csname n@A_G1_D10_ipop_fin\endcsname{7.8e-11}
\expandafter\def\csname n@A_G1_D10_mf\endcsname{50}
\expandafter\def\csname n@A_G1_D10_mf_ert\endcsname{12.3k}
\expandafter\def\csname n@A_G1_D10_mf_fin\endcsname{7.1e-11}
\expandafter\def\csname n@A_G1_D20_ama\endcsname{50}
\expandafter\def\csname n@A_G1_D20_ama_ert\endcsname{27.5k}
\expandafter\def\csname n@A_G1_D20_ama_fin\endcsname{8.3e-11}
\expandafter\def\csname n@A_G1_D20_amf\endcsname{50}
\expandafter\def\csname n@A_G1_D20_amf2\endcsname{50}
\expandafter\def\csname n@A_G1_D20_amf2_ert\endcsname{27.6k}
\expandafter\def\csname n@A_G1_D20_amf2_fin\endcsname{8.5e-11}
\expandafter\def\csname n@A_G1_D20_amf_ert\endcsname{28.5k}
\expandafter\def\csname n@A_G1_D20_amf_fin\endcsname{8.7e-11}
\expandafter\def\csname n@A_G1_D20_bipop\endcsname{50}
\expandafter\def\csname n@A_G1_D20_bipop_ert\endcsname{2.8k}
\expandafter\def\csname n@A_G1_D20_bipop_fin\endcsname{8.8e-11}
\expandafter\def\csname n@A_G1_D20_cam\endcsname{50}
\expandafter\def\csname n@A_G1_D20_cam_ert\endcsname{30.9k}
\expandafter\def\csname n@A_G1_D20_cam_fin\endcsname{8.6e-11}
\expandafter\def\csname n@A_G1_D20_cma\endcsname{50}
\expandafter\def\csname n@A_G1_D20_cma_ert\endcsname{2.8k}
\expandafter\def\csname n@A_G1_D20_cma_fin\endcsname{9.1e-11}
\expandafter\def\csname n@A_G1_D20_curg\endcsname{50}
\expandafter\def\csname n@A_G1_D20_curg_ert\endcsname{51.6k}
\expandafter\def\csname n@A_G1_D20_curg_fin\endcsname{8.7e-11}
\expandafter\def\csname n@A_G1_D20_hv0\endcsname{??}
\expandafter\def\csname n@A_G1_D20_hv0_ert\endcsname{??}
\expandafter\def\csname n@A_G1_D20_hv1\endcsname{??}
\expandafter\def\csname n@A_G1_D20_hv1_ert\endcsname{??}
\expandafter\def\csname n@A_G1_D20_ipop\endcsname{50}
\expandafter\def\csname n@A_G1_D20_ipop_ert\endcsname{2.8k}
\expandafter\def\csname n@A_G1_D20_ipop_fin\endcsname{9.1e-11}
\expandafter\def\csname n@A_G1_D20_mf\endcsname{50}
\expandafter\def\csname n@A_G1_D20_mf_ert\endcsname{52.2k}
\expandafter\def\csname n@A_G1_D20_mf_fin\endcsname{8.7e-11}
\expandafter\def\csname n@A_G1_D30_ama\endcsname{50}
\expandafter\def\csname n@A_G1_D30_ama_ert\endcsname{63.3k}
\expandafter\def\csname n@A_G1_D30_ama_fin\endcsname{8.5e-11}
\expandafter\def\csname n@A_G1_D30_amf\endcsname{50}
\expandafter\def\csname n@A_G1_D30_amf2\endcsname{50}
\expandafter\def\csname n@A_G1_D30_amf2_ert\endcsname{63.6k}
\expandafter\def\csname n@A_G1_D30_amf2_fin\endcsname{8.3e-11}
\expandafter\def\csname n@A_G1_D30_amf_ert\endcsname{63.0k}
\expandafter\def\csname n@A_G1_D30_amf_fin\endcsname{8.6e-11}
\expandafter\def\csname n@A_G1_D30_bipop\endcsname{50}
\expandafter\def\csname n@A_G1_D30_bipop_ert\endcsname{4.1k}
\expandafter\def\csname n@A_G1_D30_bipop_fin\endcsname{9e-11}
\expandafter\def\csname n@A_G1_D30_cam\endcsname{50}
\expandafter\def\csname n@A_G1_D30_cam_ert\endcsname{67.1k}
\expandafter\def\csname n@A_G1_D30_cam_fin\endcsname{8.4e-11}
\expandafter\def\csname n@A_G1_D30_cma\endcsname{50}
\expandafter\def\csname n@A_G1_D30_cma_ert\endcsname{4.1k}
\expandafter\def\csname n@A_G1_D30_cma_fin\endcsname{9.1e-11}
\expandafter\def\csname n@A_G1_D30_curg\endcsname{50}
\expandafter\def\csname n@A_G1_D30_curg_ert\endcsname{123k}
\expandafter\def\csname n@A_G1_D30_curg_fin\endcsname{8.9e-11}
\expandafter\def\csname n@A_G1_D30_hv0\endcsname{??}
\expandafter\def\csname n@A_G1_D30_hv0_ert\endcsname{??}
\expandafter\def\csname n@A_G1_D30_hv1\endcsname{??}
\expandafter\def\csname n@A_G1_D30_hv1_ert\endcsname{??}
\expandafter\def\csname n@A_G1_D30_ipop\endcsname{50}
\expandafter\def\csname n@A_G1_D30_ipop_ert\endcsname{4.1k}
\expandafter\def\csname n@A_G1_D30_ipop_fin\endcsname{9.5e-11}
\expandafter\def\csname n@A_G1_D30_mf\endcsname{50}
\expandafter\def\csname n@A_G1_D30_mf_ert\endcsname{122k}
\expandafter\def\csname n@A_G1_D30_mf_fin\endcsname{9e-11}
\expandafter\def\csname n@A_G1_D50_ama\endcsname{50}
\expandafter\def\csname n@A_G1_D50_ama_ert\endcsname{171k}
\expandafter\def\csname n@A_G1_D50_ama_fin\endcsname{8.8e-11}
\expandafter\def\csname n@A_G1_D50_amf\endcsname{50}
\expandafter\def\csname n@A_G1_D50_amf2\endcsname{50}
\expandafter\def\csname n@A_G1_D50_amf2_ert\endcsname{171k}
\expandafter\def\csname n@A_G1_D50_amf2_fin\endcsname{8.8e-11}
\expandafter\def\csname n@A_G1_D50_amf_ert\endcsname{173k}
\expandafter\def\csname n@A_G1_D50_amf_fin\endcsname{8.8e-11}
\expandafter\def\csname n@A_G1_D50_bipop\endcsname{50}
\expandafter\def\csname n@A_G1_D50_bipop_ert\endcsname{6.3k}
\expandafter\def\csname n@A_G1_D50_bipop_fin\endcsname{9.4e-11}
\expandafter\def\csname n@A_G1_D50_cam\endcsname{??}
\expandafter\def\csname n@A_G1_D50_cam_ert\endcsname{??}
\expandafter\def\csname n@A_G1_D50_cma\endcsname{50}
\expandafter\def\csname n@A_G1_D50_cma_ert\endcsname{6.4k}
\expandafter\def\csname n@A_G1_D50_cma_fin\endcsname{9.1e-11}
\expandafter\def\csname n@A_G1_D50_curg\endcsname{50}
\expandafter\def\csname n@A_G1_D50_curg_ert\endcsname{341k}
\expandafter\def\csname n@A_G1_D50_curg_fin\endcsname{9e-11}
\expandafter\def\csname n@A_G1_D50_hv0\endcsname{??}
\expandafter\def\csname n@A_G1_D50_hv0_ert\endcsname{??}
\expandafter\def\csname n@A_G1_D50_hv1\endcsname{??}
\expandafter\def\csname n@A_G1_D50_hv1_ert\endcsname{??}
\expandafter\def\csname n@A_G1_D50_ipop\endcsname{50}
\expandafter\def\csname n@A_G1_D50_ipop_ert\endcsname{6.3k}
\expandafter\def\csname n@A_G1_D50_ipop_fin\endcsname{9.3e-11}
\expandafter\def\csname n@A_G1_D50_mf\endcsname{50}
\expandafter\def\csname n@A_G1_D50_mf_ert\endcsname{347k}
\expandafter\def\csname n@A_G1_D50_mf_fin\endcsname{9e-11}
\expandafter\def\csname n@A_G2L_D10_ama\endcsname{50}
\expandafter\def\csname n@A_G2L_D10_ama_ert\endcsname{6.7k}
\expandafter\def\csname n@A_G2L_D10_ama_fin\endcsname{8.2e-11}
\expandafter\def\csname n@A_G2L_D10_amf\endcsname{50}
\expandafter\def\csname n@A_G2L_D10_amf2\endcsname{50}
\expandafter\def\csname n@A_G2L_D10_amf2_ert\endcsname{6.6k}
\expandafter\def\csname n@A_G2L_D10_amf2_fin\endcsname{7.3e-11}
\expandafter\def\csname n@A_G2L_D10_amf_ert\endcsname{6.9k}
\expandafter\def\csname n@A_G2L_D10_amf_fin\endcsname{7.8e-11}
\expandafter\def\csname n@A_G2L_D10_bipop\endcsname{50}
\expandafter\def\csname n@A_G2L_D10_bipop_ert\endcsname{3.7k}
\expandafter\def\csname n@A_G2L_D10_bipop_fin\endcsname{8.1e-11}
\expandafter\def\csname n@A_G2L_D10_cam\endcsname{50}
\expandafter\def\csname n@A_G2L_D10_cam_ert\endcsname{7.5k}
\expandafter\def\csname n@A_G2L_D10_cam_fin\endcsname{7.1e-11}
\expandafter\def\csname n@A_G2L_D10_cma\endcsname{33}
\expandafter\def\csname n@A_G2L_D10_cma_ert\endcsname{2.7k}
\expandafter\def\csname n@A_G2L_D10_cma_fin\endcsname{9.1e-11}
\expandafter\def\csname n@A_G2L_D10_curg\endcsname{49}
\expandafter\def\csname n@A_G2L_D10_curg_ert\endcsname{14.8k}
\expandafter\def\csname n@A_G2L_D10_curg_fin\endcsname{7.8e-11}
\expandafter\def\csname n@A_G2L_D10_hv0\endcsname{??}
\expandafter\def\csname n@A_G2L_D10_hv0_ert\endcsname{??}
\expandafter\def\csname n@A_G2L_D10_hv1\endcsname{??}
\expandafter\def\csname n@A_G2L_D10_hv1_ert\endcsname{??}
\expandafter\def\csname n@A_G2L_D10_ipop\endcsname{50}
\expandafter\def\csname n@A_G2L_D10_ipop_ert\endcsname{4.7k}
\expandafter\def\csname n@A_G2L_D10_ipop_fin\endcsname{8e-11}
\expandafter\def\csname n@A_G2L_D10_mf\endcsname{50}
\expandafter\def\csname n@A_G2L_D10_mf_ert\endcsname{12.0k}
\expandafter\def\csname n@A_G2L_D10_mf_fin\endcsname{8e-11}
\expandafter\def\csname n@A_G2L_D20_ama\endcsname{50}
\expandafter\def\csname n@A_G2L_D20_ama_ert\endcsname{26.7k}
\expandafter\def\csname n@A_G2L_D20_ama_fin\endcsname{8.7e-11}
\expandafter\def\csname n@A_G2L_D20_amf\endcsname{50}
\expandafter\def\csname n@A_G2L_D20_amf2\endcsname{50}
\expandafter\def\csname n@A_G2L_D20_amf2_ert\endcsname{28.0k}
\expandafter\def\csname n@A_G2L_D20_amf2_fin\endcsname{8e-11}
\expandafter\def\csname n@A_G2L_D20_amf_ert\endcsname{29.6k}
\expandafter\def\csname n@A_G2L_D20_amf_fin\endcsname{8.3e-11}
\expandafter\def\csname n@A_G2L_D20_bipop\endcsname{50}
\expandafter\def\csname n@A_G2L_D20_bipop_ert\endcsname{5.9k}
\expandafter\def\csname n@A_G2L_D20_bipop_fin\endcsname{8.7e-11}
\expandafter\def\csname n@A_G2L_D20_cam\endcsname{50}
\expandafter\def\csname n@A_G2L_D20_cam_ert\endcsname{28.7k}
\expandafter\def\csname n@A_G2L_D20_cam_fin\endcsname{8.5e-11}
\expandafter\def\csname n@A_G2L_D20_cma\endcsname{30}
\expandafter\def\csname n@A_G2L_D20_cma_ert\endcsname{5.6k}
\expandafter\def\csname n@A_G2L_D20_cma_fin\endcsname{9.7e-11}
\expandafter\def\csname n@A_G2L_D20_curg\endcsname{50}
\expandafter\def\csname n@A_G2L_D20_curg_ert\endcsname{49.0k}
\expandafter\def\csname n@A_G2L_D20_curg_fin\endcsname{9e-11}
\expandafter\def\csname n@A_G2L_D20_hv0\endcsname{??}
\expandafter\def\csname n@A_G2L_D20_hv0_ert\endcsname{??}
\expandafter\def\csname n@A_G2L_D20_hv1\endcsname{??}
\expandafter\def\csname n@A_G2L_D20_hv1_ert\endcsname{??}
\expandafter\def\csname n@A_G2L_D20_ipop\endcsname{50}
\expandafter\def\csname n@A_G2L_D20_ipop_ert\endcsname{7.9k}
\expandafter\def\csname n@A_G2L_D20_ipop_fin\endcsname{8.6e-11}
\expandafter\def\csname n@A_G2L_D20_mf\endcsname{50}
\expandafter\def\csname n@A_G2L_D20_mf_ert\endcsname{50.2k}
\expandafter\def\csname n@A_G2L_D20_mf_fin\endcsname{8.4e-11}
\expandafter\def\csname n@A_G2L_D30_ama\endcsname{50}
\expandafter\def\csname n@A_G2L_D30_ama_ert\endcsname{60.1k}
\expandafter\def\csname n@A_G2L_D30_ama_fin\endcsname{8.6e-11}
\expandafter\def\csname n@A_G2L_D30_amf\endcsname{50}
\expandafter\def\csname n@A_G2L_D30_amf2\endcsname{50}
\expandafter\def\csname n@A_G2L_D30_amf2_ert\endcsname{62.8k}
\expandafter\def\csname n@A_G2L_D30_amf2_fin\endcsname{8.6e-11}
\expandafter\def\csname n@A_G2L_D30_amf_ert\endcsname{64.4k}
\expandafter\def\csname n@A_G2L_D30_amf_fin\endcsname{8.5e-11}
\expandafter\def\csname n@A_G2L_D30_bipop\endcsname{50}
\expandafter\def\csname n@A_G2L_D30_bipop_ert\endcsname{8.9k}
\expandafter\def\csname n@A_G2L_D30_bipop_fin\endcsname{9.4e-11}
\expandafter\def\csname n@A_G2L_D30_cam\endcsname{50}
\expandafter\def\csname n@A_G2L_D30_cam_ert\endcsname{64.4k}
\expandafter\def\csname n@A_G2L_D30_cam_fin\endcsname{8.2e-11}
\expandafter\def\csname n@A_G2L_D30_cma\endcsname{31}
\expandafter\def\csname n@A_G2L_D30_cma_ert\endcsname{7.9k}
\expandafter\def\csname n@A_G2L_D30_cma_fin\endcsname{9.7e-11}
\expandafter\def\csname n@A_G2L_D30_curg\endcsname{50}
\expandafter\def\csname n@A_G2L_D30_curg_ert\endcsname{114k}
\expandafter\def\csname n@A_G2L_D30_curg_fin\endcsname{8.9e-11}
\expandafter\def\csname n@A_G2L_D30_hv0\endcsname{??}
\expandafter\def\csname n@A_G2L_D30_hv0_ert\endcsname{??}
\expandafter\def\csname n@A_G2L_D30_hv1\endcsname{??}
\expandafter\def\csname n@A_G2L_D30_hv1_ert\endcsname{??}
\expandafter\def\csname n@A_G2L_D30_ipop\endcsname{50}
\expandafter\def\csname n@A_G2L_D30_ipop_ert\endcsname{11.3k}
\expandafter\def\csname n@A_G2L_D30_ipop_fin\endcsname{9.4e-11}
\expandafter\def\csname n@A_G2L_D30_mf\endcsname{50}
\expandafter\def\csname n@A_G2L_D30_mf_ert\endcsname{116k}
\expandafter\def\csname n@A_G2L_D30_mf_fin\endcsname{8.7e-11}
\expandafter\def\csname n@A_G2L_D50_ama\endcsname{50}
\expandafter\def\csname n@A_G2L_D50_ama_ert\endcsname{169k}
\expandafter\def\csname n@A_G2L_D50_ama_fin\endcsname{9e-11}
\expandafter\def\csname n@A_G2L_D50_amf\endcsname{50}
\expandafter\def\csname n@A_G2L_D50_amf2\endcsname{50}
\expandafter\def\csname n@A_G2L_D50_amf2_ert\endcsname{180k}
\expandafter\def\csname n@A_G2L_D50_amf2_fin\endcsname{8.8e-11}
\expandafter\def\csname n@A_G2L_D50_amf_ert\endcsname{186k}
\expandafter\def\csname n@A_G2L_D50_amf_fin\endcsname{8.8e-11}
\expandafter\def\csname n@A_G2L_D50_bipop\endcsname{50}
\expandafter\def\csname n@A_G2L_D50_bipop_ert\endcsname{11.4k}
\expandafter\def\csname n@A_G2L_D50_bipop_fin\endcsname{9.3e-11}
\expandafter\def\csname n@A_G2L_D50_cam\endcsname{??}
\expandafter\def\csname n@A_G2L_D50_cam_ert\endcsname{??}
\expandafter\def\csname n@A_G2L_D50_cma\endcsname{36}
\expandafter\def\csname n@A_G2L_D50_cma_ert\endcsname{10.0k}
\expandafter\def\csname n@A_G2L_D50_cma_fin\endcsname{9.4e-11}
\expandafter\def\csname n@A_G2L_D50_curg\endcsname{50}
\expandafter\def\csname n@A_G2L_D50_curg_ert\endcsname{329k}
\expandafter\def\csname n@A_G2L_D50_curg_fin\endcsname{9.2e-11}
\expandafter\def\csname n@A_G2L_D50_hv0\endcsname{??}
\expandafter\def\csname n@A_G2L_D50_hv0_ert\endcsname{??}
\expandafter\def\csname n@A_G2L_D50_hv1\endcsname{??}
\expandafter\def\csname n@A_G2L_D50_hv1_ert\endcsname{??}
\expandafter\def\csname n@A_G2L_D50_ipop\endcsname{50}
\expandafter\def\csname n@A_G2L_D50_ipop_ert\endcsname{13.6k}
\expandafter\def\csname n@A_G2L_D50_ipop_fin\endcsname{9.3e-11}
\expandafter\def\csname n@A_G2L_D50_mf\endcsname{50}
\expandafter\def\csname n@A_G2L_D50_mf_ert\endcsname{344k}
\expandafter\def\csname n@A_G2L_D50_mf_fin\endcsname{9.1e-11}
\expandafter\def\csname n@A_G2S_D10_ama\endcsname{21}
\expandafter\def\csname n@A_G2S_D10_ama_ert\endcsname{25.7k}
\expandafter\def\csname n@A_G2S_D10_ama_fin\endcsname{5}
\expandafter\def\csname n@A_G2S_D10_amf\endcsname{48}
\expandafter\def\csname n@A_G2S_D10_amf2\endcsname{48}
\expandafter\def\csname n@A_G2S_D10_amf2_ert\endcsname{7.5k}
\expandafter\def\csname n@A_G2S_D10_amf2_fin\endcsname{8e-11}
\expandafter\def\csname n@A_G2S_D10_amf_ert\endcsname{7.4k}
\expandafter\def\csname n@A_G2S_D10_amf_fin\endcsname{8.4e-11}
\expandafter\def\csname n@A_G2S_D10_bipop\endcsname{50}
\expandafter\def\csname n@A_G2S_D10_bipop_ert\endcsname{7.1k}
\expandafter\def\csname n@A_G2S_D10_bipop_fin\endcsname{8.3e-11}
\expandafter\def\csname n@A_G2S_D10_cam\endcsname{38}
\expandafter\def\csname n@A_G2S_D10_cam_ert\endcsname{12.6k}
\expandafter\def\csname n@A_G2S_D10_cam_fin\endcsname{8.7e-11}
\expandafter\def\csname n@A_G2S_D10_cma\endcsname{23}
\expandafter\def\csname n@A_G2S_D10_cma_ert\endcsname{4.2k}
\expandafter\def\csname n@A_G2S_D10_cma_fin\endcsname{5}
\expandafter\def\csname n@A_G2S_D10_curg\endcsname{37}
\expandafter\def\csname n@A_G2S_D10_curg_ert\endcsname{64.8k}
\expandafter\def\csname n@A_G2S_D10_curg_fin\endcsname{8.8e-11}
\expandafter\def\csname n@A_G2S_D10_hv0\endcsname{??}
\expandafter\def\csname n@A_G2S_D10_hv0_ert\endcsname{??}
\expandafter\def\csname n@A_G2S_D10_hv1\endcsname{??}
\expandafter\def\csname n@A_G2S_D10_hv1_ert\endcsname{??}
\expandafter\def\csname n@A_G2S_D10_ipop\endcsname{50}
\expandafter\def\csname n@A_G2S_D10_ipop_ert\endcsname{8.8k}
\expandafter\def\csname n@A_G2S_D10_ipop_fin\endcsname{8.1e-11}
\expandafter\def\csname n@A_G2S_D10_mf\endcsname{48}
\expandafter\def\csname n@A_G2S_D10_mf_ert\endcsname{18.5k}
\expandafter\def\csname n@A_G2S_D10_mf_fin\endcsname{8.4e-11}
\expandafter\def\csname n@A_G2S_D20_ama\endcsname{0}
\expandafter\def\csname n@A_G2S_D20_ama_ert\endcsname{$\infty$}
\expandafter\def\csname n@A_G2S_D20_ama_fin\endcsname{10}
\expandafter\def\csname n@A_G2S_D20_amf\endcsname{50}
\expandafter\def\csname n@A_G2S_D20_amf2\endcsname{49}
\expandafter\def\csname n@A_G2S_D20_amf2_ert\endcsname{29.8k}
\expandafter\def\csname n@A_G2S_D20_amf2_fin\endcsname{8.4e-11}
\expandafter\def\csname n@A_G2S_D20_amf_ert\endcsname{28.9k}
\expandafter\def\csname n@A_G2S_D20_amf_fin\endcsname{8.4e-11}
\expandafter\def\csname n@A_G2S_D20_bipop\endcsname{50}
\expandafter\def\csname n@A_G2S_D20_bipop_ert\endcsname{13.2k}
\expandafter\def\csname n@A_G2S_D20_bipop_fin\endcsname{9e-11}
\expandafter\def\csname n@A_G2S_D20_cam\endcsname{15}
\expandafter\def\csname n@A_G2S_D20_cam_ert\endcsname{130k}
\expandafter\def\csname n@A_G2S_D20_cam_fin\endcsname{10}
\expandafter\def\csname n@A_G2S_D20_cma\endcsname{17}
\expandafter\def\csname n@A_G2S_D20_cma_ert\endcsname{10.9k}
\expandafter\def\csname n@A_G2S_D20_cma_fin\endcsname{10}
\expandafter\def\csname n@A_G2S_D20_curg\endcsname{18}
\expandafter\def\csname n@A_G2S_D20_curg_ert\endcsname{584k}
\expandafter\def\csname n@A_G2S_D20_curg_fin\endcsname{10}
\expandafter\def\csname n@A_G2S_D20_hv0\endcsname{??}
\expandafter\def\csname n@A_G2S_D20_hv0_ert\endcsname{??}
\expandafter\def\csname n@A_G2S_D20_hv1\endcsname{??}
\expandafter\def\csname n@A_G2S_D20_hv1_ert\endcsname{??}
\expandafter\def\csname n@A_G2S_D20_ipop\endcsname{46}
\expandafter\def\csname n@A_G2S_D20_ipop_ert\endcsname{39.6k}
\expandafter\def\csname n@A_G2S_D20_ipop_fin\endcsname{9.2e-11}
\expandafter\def\csname n@A_G2S_D20_mf\endcsname{50}
\expandafter\def\csname n@A_G2S_D20_mf_ert\endcsname{49.0k}
\expandafter\def\csname n@A_G2S_D20_mf_fin\endcsname{8.3e-11}
\expandafter\def\csname n@A_G2S_D30_ama\endcsname{0}
\expandafter\def\csname n@A_G2S_D30_ama_ert\endcsname{$\infty$}
\expandafter\def\csname n@A_G2S_D30_ama_fin\endcsname{15}
\expandafter\def\csname n@A_G2S_D30_amf\endcsname{49}
\expandafter\def\csname n@A_G2S_D30_amf2\endcsname{47}
\expandafter\def\csname n@A_G2S_D30_amf2_ert\endcsname{70.1k}
\expandafter\def\csname n@A_G2S_D30_amf2_fin\endcsname{8.5e-11}
\expandafter\def\csname n@A_G2S_D30_amf_ert\endcsname{68.9k}
\expandafter\def\csname n@A_G2S_D30_amf_fin\endcsname{8.4e-11}
\expandafter\def\csname n@A_G2S_D30_bipop\endcsname{50}
\expandafter\def\csname n@A_G2S_D30_bipop_ert\endcsname{25.1k}
\expandafter\def\csname n@A_G2S_D30_bipop_fin\endcsname{9.3e-11}
\expandafter\def\csname n@A_G2S_D30_cam\endcsname{1}
\expandafter\def\csname n@A_G2S_D30_cam_ert\endcsname{4572k}
\expandafter\def\csname n@A_G2S_D30_cam_fin\endcsname{15}
\expandafter\def\csname n@A_G2S_D30_cma\endcsname{20}
\expandafter\def\csname n@A_G2S_D30_cma_ert\endcsname{13.3k}
\expandafter\def\csname n@A_G2S_D30_cma_fin\endcsname{15}
\expandafter\def\csname n@A_G2S_D30_curg\endcsname{6}
\expandafter\def\csname n@A_G2S_D30_curg_ert\endcsname{3409k}
\expandafter\def\csname n@A_G2S_D30_curg_fin\endcsname{15}
\expandafter\def\csname n@A_G2S_D30_hv0\endcsname{??}
\expandafter\def\csname n@A_G2S_D30_hv0_ert\endcsname{??}
\expandafter\def\csname n@A_G2S_D30_hv1\endcsname{??}
\expandafter\def\csname n@A_G2S_D30_hv1_ert\endcsname{??}
\expandafter\def\csname n@A_G2S_D30_ipop\endcsname{46}
\expandafter\def\csname n@A_G2S_D30_ipop_ert\endcsname{64.3k}
\expandafter\def\csname n@A_G2S_D30_ipop_fin\endcsname{9.5e-11}
\expandafter\def\csname n@A_G2S_D30_mf\endcsname{50}
\expandafter\def\csname n@A_G2S_D30_mf_ert\endcsname{117k}
\expandafter\def\csname n@A_G2S_D30_mf_fin\endcsname{8.9e-11}
\expandafter\def\csname n@A_G2S_D50_ama\endcsname{0}
\expandafter\def\csname n@A_G2S_D50_ama_ert\endcsname{$\infty$}
\expandafter\def\csname n@A_G2S_D50_ama_fin\endcsname{25}
\expandafter\def\csname n@A_G2S_D50_amf\endcsname{49}
\expandafter\def\csname n@A_G2S_D50_amf2\endcsname{41}
\expandafter\def\csname n@A_G2S_D50_amf2_ert\endcsname{224k}
\expandafter\def\csname n@A_G2S_D50_amf2_fin\endcsname{8.8e-11}
\expandafter\def\csname n@A_G2S_D50_amf_ert\endcsname{185k}
\expandafter\def\csname n@A_G2S_D50_amf_fin\endcsname{9e-11}
\expandafter\def\csname n@A_G2S_D50_bipop\endcsname{50}
\expandafter\def\csname n@A_G2S_D50_bipop_ert\endcsname{36.3k}
\expandafter\def\csname n@A_G2S_D50_bipop_fin\endcsname{9.4e-11}
\expandafter\def\csname n@A_G2S_D50_cam\endcsname{??}
\expandafter\def\csname n@A_G2S_D50_cam_ert\endcsname{??}
\expandafter\def\csname n@A_G2S_D50_cma\endcsname{13}
\expandafter\def\csname n@A_G2S_D50_cma_ert\endcsname{32.6k}
\expandafter\def\csname n@A_G2S_D50_cma_fin\endcsname{25}
\expandafter\def\csname n@A_G2S_D50_curg\endcsname{0}
\expandafter\def\csname n@A_G2S_D50_curg_ert\endcsname{$\infty$}
\expandafter\def\csname n@A_G2S_D50_curg_fin\endcsname{25}
\expandafter\def\csname n@A_G2S_D50_hv0\endcsname{??}
\expandafter\def\csname n@A_G2S_D50_hv0_ert\endcsname{??}
\expandafter\def\csname n@A_G2S_D50_hv1\endcsname{??}
\expandafter\def\csname n@A_G2S_D50_hv1_ert\endcsname{??}
\expandafter\def\csname n@A_G2S_D50_ipop\endcsname{46}
\expandafter\def\csname n@A_G2S_D50_ipop_ert\endcsname{121k}
\expandafter\def\csname n@A_G2S_D50_ipop_fin\endcsname{9.5e-11}
\expandafter\def\csname n@A_G2S_D50_mf\endcsname{49}
\expandafter\def\csname n@A_G2S_D50_mf_ert\endcsname{350k}
\expandafter\def\csname n@A_G2S_D50_mf_fin\endcsname{9.2e-11}
\expandafter\def\csname n@B_G1_D10_ama\endcsname{50}
\expandafter\def\csname n@B_G1_D10_ama_ert\endcsname{6.6k}
\expandafter\def\csname n@B_G1_D10_ama_fin\endcsname{6.8e-11}
\expandafter\def\csname n@B_G1_D10_amf\endcsname{50}
\expandafter\def\csname n@B_G1_D10_amf2\endcsname{50}
\expandafter\def\csname n@B_G1_D10_amf2_ert\endcsname{6.6k}
\expandafter\def\csname n@B_G1_D10_amf2_fin\endcsname{7.2e-11}
\expandafter\def\csname n@B_G1_D10_amf_ert\endcsname{6.8k}
\expandafter\def\csname n@B_G1_D10_amf_fin\endcsname{8.2e-11}
\expandafter\def\csname n@B_G1_D10_bipop\endcsname{50}
\expandafter\def\csname n@B_G1_D10_bipop_ert\endcsname{1.5k}
\expandafter\def\csname n@B_G1_D10_bipop_fin\endcsname{8.6e-11}
\expandafter\def\csname n@B_G1_D10_cam\endcsname{50}
\expandafter\def\csname n@B_G1_D10_cam_ert\endcsname{7.3k}
\expandafter\def\csname n@B_G1_D10_cam_fin\endcsname{8e-11}
\expandafter\def\csname n@B_G1_D10_cma\endcsname{50}
\expandafter\def\csname n@B_G1_D10_cma_ert\endcsname{1.5k}
\expandafter\def\csname n@B_G1_D10_cma_fin\endcsname{8e-11}
\expandafter\def\csname n@B_G1_D10_curg\endcsname{50}
\expandafter\def\csname n@B_G1_D10_curg_ert\endcsname{12.1k}
\expandafter\def\csname n@B_G1_D10_curg_fin\endcsname{8.1e-11}
\expandafter\def\csname n@B_G1_D10_hv0\endcsname{50}
\expandafter\def\csname n@B_G1_D10_hv0_ert\endcsname{7.9k}
\expandafter\def\csname n@B_G1_D10_hv0_fin\endcsname{7.8e-09}
\expandafter\def\csname n@B_G1_D10_hv1\endcsname{50}
\expandafter\def\csname n@B_G1_D10_hv1_ert\endcsname{2.3k}
\expandafter\def\csname n@B_G1_D10_hv1_fin\endcsname{7.6e-09}
\expandafter\def\csname n@B_G1_D10_ipop\endcsname{50}
\expandafter\def\csname n@B_G1_D10_ipop_ert\endcsname{1.5k}
\expandafter\def\csname n@B_G1_D10_ipop_fin\endcsname{7.8e-11}
\expandafter\def\csname n@B_G1_D10_mf\endcsname{50}
\expandafter\def\csname n@B_G1_D10_mf_ert\endcsname{12.3k}
\expandafter\def\csname n@B_G1_D10_mf_fin\endcsname{7.1e-11}
\expandafter\def\csname n@B_G1_D20_ama\endcsname{50}
\expandafter\def\csname n@B_G1_D20_ama_ert\endcsname{27.5k}
\expandafter\def\csname n@B_G1_D20_ama_fin\endcsname{8.3e-11}
\expandafter\def\csname n@B_G1_D20_amf\endcsname{50}
\expandafter\def\csname n@B_G1_D20_amf2\endcsname{50}
\expandafter\def\csname n@B_G1_D20_amf2_ert\endcsname{27.6k}
\expandafter\def\csname n@B_G1_D20_amf2_fin\endcsname{8.5e-11}
\expandafter\def\csname n@B_G1_D20_amf_ert\endcsname{28.5k}
\expandafter\def\csname n@B_G1_D20_amf_fin\endcsname{8.7e-11}
\expandafter\def\csname n@B_G1_D20_bipop\endcsname{50}
\expandafter\def\csname n@B_G1_D20_bipop_ert\endcsname{2.8k}
\expandafter\def\csname n@B_G1_D20_bipop_fin\endcsname{8.8e-11}
\expandafter\def\csname n@B_G1_D20_cam\endcsname{50}
\expandafter\def\csname n@B_G1_D20_cam_ert\endcsname{30.9k}
\expandafter\def\csname n@B_G1_D20_cam_fin\endcsname{8.6e-11}
\expandafter\def\csname n@B_G1_D20_cma\endcsname{50}
\expandafter\def\csname n@B_G1_D20_cma_ert\endcsname{2.8k}
\expandafter\def\csname n@B_G1_D20_cma_fin\endcsname{9.1e-11}
\expandafter\def\csname n@B_G1_D20_curg\endcsname{50}
\expandafter\def\csname n@B_G1_D20_curg_ert\endcsname{51.6k}
\expandafter\def\csname n@B_G1_D20_curg_fin\endcsname{8.7e-11}
\expandafter\def\csname n@B_G1_D20_hv0\endcsname{50}
\expandafter\def\csname n@B_G1_D20_hv0_ert\endcsname{30.6k}
\expandafter\def\csname n@B_G1_D20_hv0_fin\endcsname{8.5e-09}
\expandafter\def\csname n@B_G1_D20_hv1\endcsname{50}
\expandafter\def\csname n@B_G1_D20_hv1_ert\endcsname{5.7k}
\expandafter\def\csname n@B_G1_D20_hv1_fin\endcsname{8.6e-09}
\expandafter\def\csname n@B_G1_D20_ipop\endcsname{50}
\expandafter\def\csname n@B_G1_D20_ipop_ert\endcsname{2.8k}
\expandafter\def\csname n@B_G1_D20_ipop_fin\endcsname{9.1e-11}
\expandafter\def\csname n@B_G1_D20_mf\endcsname{50}
\expandafter\def\csname n@B_G1_D20_mf_ert\endcsname{52.2k}
\expandafter\def\csname n@B_G1_D20_mf_fin\endcsname{8.7e-11}
\expandafter\def\csname n@B_G1_D30_ama\endcsname{50}
\expandafter\def\csname n@B_G1_D30_ama_ert\endcsname{63.3k}
\expandafter\def\csname n@B_G1_D30_ama_fin\endcsname{8.5e-11}
\expandafter\def\csname n@B_G1_D30_amf\endcsname{50}
\expandafter\def\csname n@B_G1_D30_amf2\endcsname{50}
\expandafter\def\csname n@B_G1_D30_amf2_ert\endcsname{63.6k}
\expandafter\def\csname n@B_G1_D30_amf2_fin\endcsname{8.3e-11}
\expandafter\def\csname n@B_G1_D30_amf_ert\endcsname{63.0k}
\expandafter\def\csname n@B_G1_D30_amf_fin\endcsname{8.6e-11}
\expandafter\def\csname n@B_G1_D30_bipop\endcsname{50}
\expandafter\def\csname n@B_G1_D30_bipop_ert\endcsname{4.1k}
\expandafter\def\csname n@B_G1_D30_bipop_fin\endcsname{9e-11}
\expandafter\def\csname n@B_G1_D30_cam\endcsname{50}
\expandafter\def\csname n@B_G1_D30_cam_ert\endcsname{67.1k}
\expandafter\def\csname n@B_G1_D30_cam_fin\endcsname{8.4e-11}
\expandafter\def\csname n@B_G1_D30_cma\endcsname{50}
\expandafter\def\csname n@B_G1_D30_cma_ert\endcsname{4.1k}
\expandafter\def\csname n@B_G1_D30_cma_fin\endcsname{9.1e-11}
\expandafter\def\csname n@B_G1_D30_curg\endcsname{50}
\expandafter\def\csname n@B_G1_D30_curg_ert\endcsname{123k}
\expandafter\def\csname n@B_G1_D30_curg_fin\endcsname{8.9e-11}
\expandafter\def\csname n@B_G1_D30_hv0\endcsname{50}
\expandafter\def\csname n@B_G1_D30_hv0_ert\endcsname{61.7k}
\expandafter\def\csname n@B_G1_D30_hv0_fin\endcsname{8.7e-09}
\expandafter\def\csname n@B_G1_D30_hv1\endcsname{50}
\expandafter\def\csname n@B_G1_D30_hv1_ert\endcsname{9.1k}
\expandafter\def\csname n@B_G1_D30_hv1_fin\endcsname{8.8e-09}
\expandafter\def\csname n@B_G1_D30_ipop\endcsname{50}
\expandafter\def\csname n@B_G1_D30_ipop_ert\endcsname{4.1k}
\expandafter\def\csname n@B_G1_D30_ipop_fin\endcsname{9.5e-11}
\expandafter\def\csname n@B_G1_D30_mf\endcsname{50}
\expandafter\def\csname n@B_G1_D30_mf_ert\endcsname{122k}
\expandafter\def\csname n@B_G1_D30_mf_fin\endcsname{9e-11}
\expandafter\def\csname n@B_G1_D50_ama\endcsname{50}
\expandafter\def\csname n@B_G1_D50_ama_ert\endcsname{171k}
\expandafter\def\csname n@B_G1_D50_ama_fin\endcsname{8.8e-11}
\expandafter\def\csname n@B_G1_D50_amf\endcsname{50}
\expandafter\def\csname n@B_G1_D50_amf2\endcsname{50}
\expandafter\def\csname n@B_G1_D50_amf2_ert\endcsname{171k}
\expandafter\def\csname n@B_G1_D50_amf2_fin\endcsname{8.8e-11}
\expandafter\def\csname n@B_G1_D50_amf_ert\endcsname{173k}
\expandafter\def\csname n@B_G1_D50_amf_fin\endcsname{8.8e-11}
\expandafter\def\csname n@B_G1_D50_bipop\endcsname{50}
\expandafter\def\csname n@B_G1_D50_bipop_ert\endcsname{6.3k}
\expandafter\def\csname n@B_G1_D50_bipop_fin\endcsname{9.4e-11}
\expandafter\def\csname n@B_G1_D50_cam\endcsname{??}
\expandafter\def\csname n@B_G1_D50_cam_ert\endcsname{??}
\expandafter\def\csname n@B_G1_D50_cma\endcsname{50}
\expandafter\def\csname n@B_G1_D50_cma_ert\endcsname{6.4k}
\expandafter\def\csname n@B_G1_D50_cma_fin\endcsname{9.1e-11}
\expandafter\def\csname n@B_G1_D50_curg\endcsname{50}
\expandafter\def\csname n@B_G1_D50_curg_ert\endcsname{341k}
\expandafter\def\csname n@B_G1_D50_curg_fin\endcsname{9e-11}
\expandafter\def\csname n@B_G1_D50_hv0\endcsname{??}
\expandafter\def\csname n@B_G1_D50_hv0_ert\endcsname{??}
\expandafter\def\csname n@B_G1_D50_hv1\endcsname{??}
\expandafter\def\csname n@B_G1_D50_hv1_ert\endcsname{??}
\expandafter\def\csname n@B_G1_D50_ipop\endcsname{50}
\expandafter\def\csname n@B_G1_D50_ipop_ert\endcsname{6.3k}
\expandafter\def\csname n@B_G1_D50_ipop_fin\endcsname{9.3e-11}
\expandafter\def\csname n@B_G1_D50_mf\endcsname{50}
\expandafter\def\csname n@B_G1_D50_mf_ert\endcsname{347k}
\expandafter\def\csname n@B_G1_D50_mf_fin\endcsname{9e-11}
\expandafter\def\csname n@B_G2L_D10_ama\endcsname{50}
\expandafter\def\csname n@B_G2L_D10_ama_ert\endcsname{6.7k}
\expandafter\def\csname n@B_G2L_D10_ama_fin\endcsname{8.2e-11}
\expandafter\def\csname n@B_G2L_D10_amf\endcsname{50}
\expandafter\def\csname n@B_G2L_D10_amf2\endcsname{50}
\expandafter\def\csname n@B_G2L_D10_amf2_ert\endcsname{6.6k}
\expandafter\def\csname n@B_G2L_D10_amf2_fin\endcsname{7.3e-11}
\expandafter\def\csname n@B_G2L_D10_amf_ert\endcsname{6.9k}
\expandafter\def\csname n@B_G2L_D10_amf_fin\endcsname{7.8e-11}
\expandafter\def\csname n@B_G2L_D10_bipop\endcsname{50}
\expandafter\def\csname n@B_G2L_D10_bipop_ert\endcsname{3.7k}
\expandafter\def\csname n@B_G2L_D10_bipop_fin\endcsname{8.1e-11}
\expandafter\def\csname n@B_G2L_D10_cam\endcsname{50}
\expandafter\def\csname n@B_G2L_D10_cam_ert\endcsname{7.5k}
\expandafter\def\csname n@B_G2L_D10_cam_fin\endcsname{7.1e-11}
\expandafter\def\csname n@B_G2L_D10_cma\endcsname{50}
\expandafter\def\csname n@B_G2L_D10_cma_ert\endcsname{3.1k}
\expandafter\def\csname n@B_G2L_D10_cma_fin\endcsname{8.2e-11}
\expandafter\def\csname n@B_G2L_D10_curg\endcsname{50}
\expandafter\def\csname n@B_G2L_D10_curg_ert\endcsname{12.2k}
\expandafter\def\csname n@B_G2L_D10_curg_fin\endcsname{7.8e-11}
\expandafter\def\csname n@B_G2L_D10_hv0\endcsname{50}
\expandafter\def\csname n@B_G2L_D10_hv0_ert\endcsname{11.2k}
\expandafter\def\csname n@B_G2L_D10_hv0_fin\endcsname{7.5e-09}
\expandafter\def\csname n@B_G2L_D10_hv1\endcsname{50}
\expandafter\def\csname n@B_G2L_D10_hv1_ert\endcsname{3.2k}
\expandafter\def\csname n@B_G2L_D10_hv1_fin\endcsname{8.3e-09}
\expandafter\def\csname n@B_G2L_D10_ipop\endcsname{50}
\expandafter\def\csname n@B_G2L_D10_ipop_ert\endcsname{4.7k}
\expandafter\def\csname n@B_G2L_D10_ipop_fin\endcsname{8e-11}
\expandafter\def\csname n@B_G2L_D10_mf\endcsname{50}
\expandafter\def\csname n@B_G2L_D10_mf_ert\endcsname{12.0k}
\expandafter\def\csname n@B_G2L_D10_mf_fin\endcsname{8e-11}
\expandafter\def\csname n@B_G2L_D20_ama\endcsname{50}
\expandafter\def\csname n@B_G2L_D20_ama_ert\endcsname{26.7k}
\expandafter\def\csname n@B_G2L_D20_ama_fin\endcsname{8.7e-11}
\expandafter\def\csname n@B_G2L_D20_amf\endcsname{50}
\expandafter\def\csname n@B_G2L_D20_amf2\endcsname{50}
\expandafter\def\csname n@B_G2L_D20_amf2_ert\endcsname{28.0k}
\expandafter\def\csname n@B_G2L_D20_amf2_fin\endcsname{8e-11}
\expandafter\def\csname n@B_G2L_D20_amf_ert\endcsname{29.6k}
\expandafter\def\csname n@B_G2L_D20_amf_fin\endcsname{8.3e-11}
\expandafter\def\csname n@B_G2L_D20_bipop\endcsname{50}
\expandafter\def\csname n@B_G2L_D20_bipop_ert\endcsname{5.9k}
\expandafter\def\csname n@B_G2L_D20_bipop_fin\endcsname{8.7e-11}
\expandafter\def\csname n@B_G2L_D20_cam\endcsname{50}
\expandafter\def\csname n@B_G2L_D20_cam_ert\endcsname{28.7k}
\expandafter\def\csname n@B_G2L_D20_cam_fin\endcsname{8.5e-11}
\expandafter\def\csname n@B_G2L_D20_cma\endcsname{50}
\expandafter\def\csname n@B_G2L_D20_cma_ert\endcsname{5.5k}
\expandafter\def\csname n@B_G2L_D20_cma_fin\endcsname{9.2e-11}
\expandafter\def\csname n@B_G2L_D20_curg\endcsname{50}
\expandafter\def\csname n@B_G2L_D20_curg_ert\endcsname{49.0k}
\expandafter\def\csname n@B_G2L_D20_curg_fin\endcsname{9e-11}
\expandafter\def\csname n@B_G2L_D20_hv0\endcsname{50}
\expandafter\def\csname n@B_G2L_D20_hv0_ert\endcsname{36.6k}
\expandafter\def\csname n@B_G2L_D20_hv0_fin\endcsname{8.7e-09}
\expandafter\def\csname n@B_G2L_D20_hv1\endcsname{50}
\expandafter\def\csname n@B_G2L_D20_hv1_ert\endcsname{7.3k}
\expandafter\def\csname n@B_G2L_D20_hv1_fin\endcsname{8.5e-09}
\expandafter\def\csname n@B_G2L_D20_ipop\endcsname{50}
\expandafter\def\csname n@B_G2L_D20_ipop_ert\endcsname{7.9k}
\expandafter\def\csname n@B_G2L_D20_ipop_fin\endcsname{8.6e-11}
\expandafter\def\csname n@B_G2L_D20_mf\endcsname{50}
\expandafter\def\csname n@B_G2L_D20_mf_ert\endcsname{50.2k}
\expandafter\def\csname n@B_G2L_D20_mf_fin\endcsname{8.4e-11}
\expandafter\def\csname n@B_G2L_D30_ama\endcsname{50}
\expandafter\def\csname n@B_G2L_D30_ama_ert\endcsname{60.1k}
\expandafter\def\csname n@B_G2L_D30_ama_fin\endcsname{8.6e-11}
\expandafter\def\csname n@B_G2L_D30_amf\endcsname{50}
\expandafter\def\csname n@B_G2L_D30_amf2\endcsname{50}
\expandafter\def\csname n@B_G2L_D30_amf2_ert\endcsname{62.8k}
\expandafter\def\csname n@B_G2L_D30_amf2_fin\endcsname{8.6e-11}
\expandafter\def\csname n@B_G2L_D30_amf_ert\endcsname{64.4k}
\expandafter\def\csname n@B_G2L_D30_amf_fin\endcsname{8.5e-11}
\expandafter\def\csname n@B_G2L_D30_bipop\endcsname{50}
\expandafter\def\csname n@B_G2L_D30_bipop_ert\endcsname{8.9k}
\expandafter\def\csname n@B_G2L_D30_bipop_fin\endcsname{9.4e-11}
\expandafter\def\csname n@B_G2L_D30_cam\endcsname{50}
\expandafter\def\csname n@B_G2L_D30_cam_ert\endcsname{64.4k}
\expandafter\def\csname n@B_G2L_D30_cam_fin\endcsname{8.2e-11}
\expandafter\def\csname n@B_G2L_D30_cma\endcsname{50}
\expandafter\def\csname n@B_G2L_D30_cma_ert\endcsname{7.9k}
\expandafter\def\csname n@B_G2L_D30_cma_fin\endcsname{9e-11}
\expandafter\def\csname n@B_G2L_D30_curg\endcsname{50}
\expandafter\def\csname n@B_G2L_D30_curg_ert\endcsname{114k}
\expandafter\def\csname n@B_G2L_D30_curg_fin\endcsname{8.9e-11}
\expandafter\def\csname n@B_G2L_D30_hv0\endcsname{50}
\expandafter\def\csname n@B_G2L_D30_hv0_ert\endcsname{70.1k}
\expandafter\def\csname n@B_G2L_D30_hv0_fin\endcsname{8.6e-09}
\expandafter\def\csname n@B_G2L_D30_hv1\endcsname{50}
\expandafter\def\csname n@B_G2L_D30_hv1_ert\endcsname{10.5k}
\expandafter\def\csname n@B_G2L_D30_hv1_fin\endcsname{8.8e-09}
\expandafter\def\csname n@B_G2L_D30_ipop\endcsname{50}
\expandafter\def\csname n@B_G2L_D30_ipop_ert\endcsname{11.3k}
\expandafter\def\csname n@B_G2L_D30_ipop_fin\endcsname{9.4e-11}
\expandafter\def\csname n@B_G2L_D30_mf\endcsname{50}
\expandafter\def\csname n@B_G2L_D30_mf_ert\endcsname{116k}
\expandafter\def\csname n@B_G2L_D30_mf_fin\endcsname{8.7e-11}
\expandafter\def\csname n@B_G2L_D50_ama\endcsname{50}
\expandafter\def\csname n@B_G2L_D50_ama_ert\endcsname{169k}
\expandafter\def\csname n@B_G2L_D50_ama_fin\endcsname{9e-11}
\expandafter\def\csname n@B_G2L_D50_amf\endcsname{50}
\expandafter\def\csname n@B_G2L_D50_amf2\endcsname{50}
\expandafter\def\csname n@B_G2L_D50_amf2_ert\endcsname{180k}
\expandafter\def\csname n@B_G2L_D50_amf2_fin\endcsname{8.8e-11}
\expandafter\def\csname n@B_G2L_D50_amf_ert\endcsname{186k}
\expandafter\def\csname n@B_G2L_D50_amf_fin\endcsname{8.8e-11}
\expandafter\def\csname n@B_G2L_D50_bipop\endcsname{50}
\expandafter\def\csname n@B_G2L_D50_bipop_ert\endcsname{11.4k}
\expandafter\def\csname n@B_G2L_D50_bipop_fin\endcsname{9.3e-11}
\expandafter\def\csname n@B_G2L_D50_cam\endcsname{??}
\expandafter\def\csname n@B_G2L_D50_cam_ert\endcsname{??}
\expandafter\def\csname n@B_G2L_D50_cma\endcsname{50}
\expandafter\def\csname n@B_G2L_D50_cma_ert\endcsname{10.8k}
\expandafter\def\csname n@B_G2L_D50_cma_fin\endcsname{9.1e-11}
\expandafter\def\csname n@B_G2L_D50_curg\endcsname{50}
\expandafter\def\csname n@B_G2L_D50_curg_ert\endcsname{329k}
\expandafter\def\csname n@B_G2L_D50_curg_fin\endcsname{9.2e-11}
\expandafter\def\csname n@B_G2L_D50_hv0\endcsname{??}
\expandafter\def\csname n@B_G2L_D50_hv0_ert\endcsname{??}
\expandafter\def\csname n@B_G2L_D50_hv1\endcsname{??}
\expandafter\def\csname n@B_G2L_D50_hv1_ert\endcsname{??}
\expandafter\def\csname n@B_G2L_D50_ipop\endcsname{50}
\expandafter\def\csname n@B_G2L_D50_ipop_ert\endcsname{13.6k}
\expandafter\def\csname n@B_G2L_D50_ipop_fin\endcsname{9.3e-11}
\expandafter\def\csname n@B_G2L_D50_mf\endcsname{50}
\expandafter\def\csname n@B_G2L_D50_mf_ert\endcsname{344k}
\expandafter\def\csname n@B_G2L_D50_mf_fin\endcsname{9.1e-11}
\expandafter\def\csname n@B_G2S_D10_ama\endcsname{50}
\expandafter\def\csname n@B_G2S_D10_ama_ert\endcsname{19.9k}
\expandafter\def\csname n@B_G2S_D10_ama_fin\endcsname{7.5e-11}
\expandafter\def\csname n@B_G2S_D10_amf\endcsname{50}
\expandafter\def\csname n@B_G2S_D10_amf2\endcsname{50}
\expandafter\def\csname n@B_G2S_D10_amf2_ert\endcsname{7.4k}
\expandafter\def\csname n@B_G2S_D10_amf2_fin\endcsname{7.8e-11}
\expandafter\def\csname n@B_G2S_D10_amf_ert\endcsname{7.3k}
\expandafter\def\csname n@B_G2S_D10_amf_fin\endcsname{8.4e-11}
\expandafter\def\csname n@B_G2S_D10_bipop\endcsname{50}
\expandafter\def\csname n@B_G2S_D10_bipop_ert\endcsname{7.1k}
\expandafter\def\csname n@B_G2S_D10_bipop_fin\endcsname{8.3e-11}
\expandafter\def\csname n@B_G2S_D10_cam\endcsname{50}
\expandafter\def\csname n@B_G2S_D10_cam_ert\endcsname{13.1k}
\expandafter\def\csname n@B_G2S_D10_cam_fin\endcsname{7.7e-11}
\expandafter\def\csname n@B_G2S_D10_cma\endcsname{50}
\expandafter\def\csname n@B_G2S_D10_cma_ert\endcsname{4.0k}
\expandafter\def\csname n@B_G2S_D10_cma_fin\endcsname{7.5e-11}
\expandafter\def\csname n@B_G2S_D10_curg\endcsname{50}
\expandafter\def\csname n@B_G2S_D10_curg_ert\endcsname{18.4k}
\expandafter\def\csname n@B_G2S_D10_curg_fin\endcsname{7e-11}
\expandafter\def\csname n@B_G2S_D10_hv0\endcsname{50}
\expandafter\def\csname n@B_G2S_D10_hv0_ert\endcsname{13.9k}
\expandafter\def\csname n@B_G2S_D10_hv0_fin\endcsname{8.4e-09}
\expandafter\def\csname n@B_G2S_D10_hv1\endcsname{50}
\expandafter\def\csname n@B_G2S_D10_hv1_ert\endcsname{3.9k}
\expandafter\def\csname n@B_G2S_D10_hv1_fin\endcsname{8.3e-09}
\expandafter\def\csname n@B_G2S_D10_ipop\endcsname{50}
\expandafter\def\csname n@B_G2S_D10_ipop_ert\endcsname{8.8k}
\expandafter\def\csname n@B_G2S_D10_ipop_fin\endcsname{8.1e-11}
\expandafter\def\csname n@B_G2S_D10_mf\endcsname{50}
\expandafter\def\csname n@B_G2S_D10_mf_ert\endcsname{13.2k}
\expandafter\def\csname n@B_G2S_D10_mf_fin\endcsname{8.3e-11}
\expandafter\def\csname n@B_G2S_D20_ama\endcsname{21}
\expandafter\def\csname n@B_G2S_D20_ama_ert\endcsname{563k}
\expandafter\def\csname n@B_G2S_D20_ama_fin\endcsname{10}
\expandafter\def\csname n@B_G2S_D20_amf\endcsname{50}
\expandafter\def\csname n@B_G2S_D20_amf2\endcsname{50}
\expandafter\def\csname n@B_G2S_D20_amf2_ert\endcsname{29.7k}
\expandafter\def\csname n@B_G2S_D20_amf2_fin\endcsname{8.3e-11}
\expandafter\def\csname n@B_G2S_D20_amf_ert\endcsname{28.9k}
\expandafter\def\csname n@B_G2S_D20_amf_fin\endcsname{8.4e-11}
\expandafter\def\csname n@B_G2S_D20_bipop\endcsname{50}
\expandafter\def\csname n@B_G2S_D20_bipop_ert\endcsname{13.2k}
\expandafter\def\csname n@B_G2S_D20_bipop_fin\endcsname{9e-11}
\expandafter\def\csname n@B_G2S_D20_cam\endcsname{45}
\expandafter\def\csname n@B_G2S_D20_cam_ert\endcsname{145k}
\expandafter\def\csname n@B_G2S_D20_cam_fin\endcsname{8.5e-11}
\expandafter\def\csname n@B_G2S_D20_cma\endcsname{50}
\expandafter\def\csname n@B_G2S_D20_cma_ert\endcsname{10.0k}
\expandafter\def\csname n@B_G2S_D20_cma_fin\endcsname{8.6e-11}
\expandafter\def\csname n@B_G2S_D20_curg\endcsname{46}
\expandafter\def\csname n@B_G2S_D20_curg_ert\endcsname{143k}
\expandafter\def\csname n@B_G2S_D20_curg_fin\endcsname{8.6e-11}
\expandafter\def\csname n@B_G2S_D20_hv0\endcsname{50}
\expandafter\def\csname n@B_G2S_D20_hv0_ert\endcsname{49.7k}
\expandafter\def\csname n@B_G2S_D20_hv0_fin\endcsname{8.5e-09}
\expandafter\def\csname n@B_G2S_D20_hv1\endcsname{50}
\expandafter\def\csname n@B_G2S_D20_hv1_ert\endcsname{10.0k}
\expandafter\def\csname n@B_G2S_D20_hv1_fin\endcsname{8.7e-09}
\expandafter\def\csname n@B_G2S_D20_ipop\endcsname{46}
\expandafter\def\csname n@B_G2S_D20_ipop_ert\endcsname{39.6k}
\expandafter\def\csname n@B_G2S_D20_ipop_fin\endcsname{9.2e-11}
\expandafter\def\csname n@B_G2S_D20_mf\endcsname{50}
\expandafter\def\csname n@B_G2S_D20_mf_ert\endcsname{49.0k}
\expandafter\def\csname n@B_G2S_D20_mf_fin\endcsname{8.3e-11}
\expandafter\def\csname n@B_G2S_D30_ama\endcsname{1}
\expandafter\def\csname n@B_G2S_D30_ama_ert\endcsname{22487k}
\expandafter\def\csname n@B_G2S_D30_ama_fin\endcsname{15}
\expandafter\def\csname n@B_G2S_D30_amf\endcsname{50}
\expandafter\def\csname n@B_G2S_D30_amf2\endcsname{50}
\expandafter\def\csname n@B_G2S_D30_amf2_ert\endcsname{69.8k}
\expandafter\def\csname n@B_G2S_D30_amf2_fin\endcsname{8.5e-11}
\expandafter\def\csname n@B_G2S_D30_amf_ert\endcsname{68.9k}
\expandafter\def\csname n@B_G2S_D30_amf_fin\endcsname{8.4e-11}
\expandafter\def\csname n@B_G2S_D30_bipop\endcsname{50}
\expandafter\def\csname n@B_G2S_D30_bipop_ert\endcsname{25.1k}
\expandafter\def\csname n@B_G2S_D30_bipop_fin\endcsname{9.3e-11}
\expandafter\def\csname n@B_G2S_D30_cam\endcsname{7}
\expandafter\def\csname n@B_G2S_D30_cam_ert\endcsname{2966k}
\expandafter\def\csname n@B_G2S_D30_cam_fin\endcsname{15}
\expandafter\def\csname n@B_G2S_D30_cma\endcsname{50}
\expandafter\def\csname n@B_G2S_D30_cma_ert\endcsname{16.0k}
\expandafter\def\csname n@B_G2S_D30_cma_fin\endcsname{8.9e-11}
\expandafter\def\csname n@B_G2S_D30_curg\endcsname{13}
\expandafter\def\csname n@B_G2S_D30_curg_ert\endcsname{1512k}
\expandafter\def\csname n@B_G2S_D30_curg_fin\endcsname{15}
\expandafter\def\csname n@B_G2S_D30_hv0\endcsname{50}
\expandafter\def\csname n@B_G2S_D30_hv0_ert\endcsname{131k}
\expandafter\def\csname n@B_G2S_D30_hv0_fin\endcsname{8.3e-09}
\expandafter\def\csname n@B_G2S_D30_hv1\endcsname{50}
\expandafter\def\csname n@B_G2S_D30_hv1_ert\endcsname{19.2k}
\expandafter\def\csname n@B_G2S_D30_hv1_fin\endcsname{9e-09}
\expandafter\def\csname n@B_G2S_D30_ipop\endcsname{46}
\expandafter\def\csname n@B_G2S_D30_ipop_ert\endcsname{64.3k}
\expandafter\def\csname n@B_G2S_D30_ipop_fin\endcsname{9.5e-11}
\expandafter\def\csname n@B_G2S_D30_mf\endcsname{50}
\expandafter\def\csname n@B_G2S_D30_mf_ert\endcsname{117k}
\expandafter\def\csname n@B_G2S_D30_mf_fin\endcsname{8.9e-11}
\expandafter\def\csname n@B_G2S_D50_ama\endcsname{0}
\expandafter\def\csname n@B_G2S_D50_ama_ert\endcsname{$\infty$}
\expandafter\def\csname n@B_G2S_D50_ama_fin\endcsname{25}
\expandafter\def\csname n@B_G2S_D50_amf\endcsname{50}
\expandafter\def\csname n@B_G2S_D50_amf2\endcsname{49}
\expandafter\def\csname n@B_G2S_D50_amf2_ert\endcsname{233k}
\expandafter\def\csname n@B_G2S_D50_amf2_fin\endcsname{8.6e-11}
\expandafter\def\csname n@B_G2S_D50_amf_ert\endcsname{185k}
\expandafter\def\csname n@B_G2S_D50_amf_fin\endcsname{8.9e-11}
\expandafter\def\csname n@B_G2S_D50_bipop\endcsname{50}
\expandafter\def\csname n@B_G2S_D50_bipop_ert\endcsname{36.3k}
\expandafter\def\csname n@B_G2S_D50_bipop_fin\endcsname{9.4e-11}
\expandafter\def\csname n@B_G2S_D50_cam\endcsname{??}
\expandafter\def\csname n@B_G2S_D50_cam_ert\endcsname{??}
\expandafter\def\csname n@B_G2S_D50_cma\endcsname{50}
\expandafter\def\csname n@B_G2S_D50_cma_ert\endcsname{33.4k}
\expandafter\def\csname n@B_G2S_D50_cma_fin\endcsname{9.3e-11}
\expandafter\def\csname n@B_G2S_D50_curg\endcsname{0}
\expandafter\def\csname n@B_G2S_D50_curg_ert\endcsname{$\infty$}
\expandafter\def\csname n@B_G2S_D50_curg_fin\endcsname{25}
\expandafter\def\csname n@B_G2S_D50_hv0\endcsname{??}
\expandafter\def\csname n@B_G2S_D50_hv0_ert\endcsname{??}
\expandafter\def\csname n@B_G2S_D50_hv1\endcsname{??}
\expandafter\def\csname n@B_G2S_D50_hv1_ert\endcsname{??}
\expandafter\def\csname n@B_G2S_D50_ipop\endcsname{46}
\expandafter\def\csname n@B_G2S_D50_ipop_ert\endcsname{121k}
\expandafter\def\csname n@B_G2S_D50_ipop_fin\endcsname{9.5e-11}
\expandafter\def\csname n@B_G2S_D50_mf\endcsname{49}
\expandafter\def\csname n@B_G2S_D50_mf_ert\endcsname{350k}
\expandafter\def\csname n@B_G2S_D50_mf_fin\endcsname{9.2e-11}
\expandafter\def\csname n@B_summary\endcsname{raises the successes on G2-S from 50, 46, 13, 0 to 50, 50, 50, 49 for CUR-G and from 50, 21, 1, 0 to 50, 50, 50, 50 for AMaLGaM at $D=10, 20, 30, 50$}
\expandafter\def\csname n@abl50_dip_lost\endcsname{15}
\expandafter\def\csname n@abl50_full\endcsname{49}
\expandafter\def\csname n@abl50_summary\endcsname{At $D=50$ (fourth column) the full method succeeds in 49 of 50 runs and every variant stays within two runs of it except the density-dip test (34 of 50) and the single-start EM with 40 iterations (24 of 50): the detections of the first generations at $D=50$ are unequal and moderately separated groups, which a single EM start fits as a scale mixture and which the dip test rejects (Section~\ref{sec:calib}); the two elements of the detector that the test runs of Section~\ref{sec:setup} changed are exactly the ones that this dimension exposes.}
\expandafter\def\csname n@abl_alloc_summary\endcsname{Over the three dimensions the full method succeeds in 148 of 150 runs; fitted weights, soft in 142, fitted weights, hard in 142, equal shares, soft in 149, equal shares, hard in 149, width-weighted mixture, soft in 148, width-weighted mixture, hard in 148, floored weights, soft in 147, floored weights, hard in 148, union, hard in 148.}
\expandafter\def\csname n@abl_det_summary\endcsname{principal directions only gives 147; the dip-test variant gives 142; rank weights in the detector gives 147; diagonal local Gaussians gives 148; no cap gives 148; single-start EM gives 145 successes of 150 against 148 for the full method}
\expandafter\def\csname n@abl_dip_D50_text\endcsname{34 against 49 of 50 successes with and without it}
\expandafter\def\csname n@abl_union_verdict\endcsname{All these rules compose the same two membership functions and local Gaussians. The composition by the fitted weights, which is the mixture, is the one that falls behind; the compositions whose shares do not follow the head count (the union, equal shares, the width-weighted sum and the floored weights) perform alike, and the union is the one whose density and shares are given by Proposition~\ref{prop:union}.}
\expandafter\def\csname n@ablg1_full_D10\endcsname{9.6}
\expandafter\def\csname n@ablg1_full_D20\endcsname{2.9}
\expandafter\def\csname n@ablg1_full_ci_D20\endcsname{[0.981, 1.046]}
\expandafter\def\csname n@ablg1_full_ratio_D20\endcsname{1.013}
\expandafter\def\csname n@ablg1_nodip_D10\endcsname{4.3}
\expandafter\def\csname n@ablg1_nodip_D20\endcsname{0.2}
\expandafter\def\csname n@ablx_summary\endcsname{Over the 6 settings the union succeeds in 218 of 300 runs, the fitted weights in 182 and the equal shares in 218; the fitted weights are behind the union in 6 of the 6 settings (by 6, 7, 1, 10, 6, 6 runs), whereas the equal shares have the same success count as the union in every setting.}
\expandafter\def\csname n@bbob_bipop\endcsname{1065}
\expandafter\def\csname n@bbob_curg\endcsname{912}
\expandafter\def\csname n@bbob_f24_D20\endcsname{24.5, 24.1 and 17.3}
\expandafter\def\csname n@bbob_losses\endcsname{0}
\expandafter\def\csname n@bbob_mf\endcsname{901}
\expandafter\def\csname n@bbob_sig_list\endcsname{no pair is significant}
\expandafter\def\csname n@bbob_sig_tail\endcsname{}
\expandafter\def\csname n@bbob_wins\endcsname{0}
\expandafter\def\csname n@be_D10\endcsname{4~$\mu$s}
\expandafter\def\csname n@be_D20\endcsname{1~$\mu$s}
\expandafter\def\csname n@be_D30\endcsname{any cost (the component is faster even with free evaluations)}
\expandafter\def\csname n@be_D50\endcsname{any cost (CUR-G never reached the target)}
\expandafter\def\csname n@bipop_ert_summary\endcsname{7.1k, 13.2k, 25.1k, 36.3k evaluations}
\expandafter\def\csname n@bipop_g2s_summary\endcsname{50 of 50 runs at every dimension}
\expandafter\def\csname n@cal_ce_10_100\endcsname{0.7}
\expandafter\def\csname n@cal_ce_10_377\endcsname{0.0}
\expandafter\def\csname n@cal_ce_10_40\endcsname{3.9}
\expandafter\def\csname n@cal_ce_20_100\endcsname{0.4}
\expandafter\def\csname n@cal_ce_20_377\endcsname{0.1}
\expandafter\def\csname n@cal_ce_20_40\endcsname{5.1}
\expandafter\def\csname n@cal_ce_max377\endcsname{0.1}
\expandafter\def\csname n@cal_ce_max40\endcsname{5.1}
\expandafter\def\csname n@cal_corr_10_100\endcsname{66.2}
\expandafter\def\csname n@cal_corr_10_40\endcsname{34.6}
\expandafter\def\csname n@cal_corr_20_100\endcsname{45.2}
\expandafter\def\csname n@cal_corr_20_40\endcsname{10.4}
\expandafter\def\csname n@cal_d4_10_100\endcsname{97.8}
\expandafter\def\csname n@cal_d4_10_40\endcsname{49.6}
\expandafter\def\csname n@cal_d4_20_100\endcsname{92.8}
\expandafter\def\csname n@cal_d4_20_40\endcsname{36.2}
\expandafter\def\csname n@cal_d5_10_100\endcsname{100.0}
\expandafter\def\csname n@cal_d5_10_40\endcsname{94.0}
\expandafter\def\csname n@cal_d5_20_100\endcsname{100.0}
\expandafter\def\csname n@cal_d5_20_40\endcsname{87.6}
\expandafter\def\csname n@cal_dip_corr_10_100\endcsname{53.0}
\expandafter\def\csname n@cal_dip_corr_10_40\endcsname{28.4}
\expandafter\def\csname n@cal_dip_corr_20_100\endcsname{26.6}
\expandafter\def\csname n@cal_dip_corr_20_40\endcsname{6.0}
\expandafter\def\csname n@cal_dip_d4_10_100\endcsname{47.0}
\expandafter\def\csname n@cal_dip_d4_10_40\endcsname{38.0}
\expandafter\def\csname n@cal_dip_d4_20_100\endcsname{40.4}
\expandafter\def\csname n@cal_dip_d4_20_40\endcsname{28.2}
\expandafter\def\csname n@cal_dip_d5_10_100\endcsname{97.0}
\expandafter\def\csname n@cal_dip_d5_10_40\endcsname{86.6}
\expandafter\def\csname n@cal_dip_d5_20_100\endcsname{97.8}
\expandafter\def\csname n@cal_dip_d5_20_40\endcsname{82.6}
\expandafter\def\csname n@cal_dip_null_10_100\endcsname{0.1}
\expandafter\def\csname n@cal_dip_null_10_40\endcsname{1.7}
\expandafter\def\csname n@cal_dip_null_20_100\endcsname{0.1}
\expandafter\def\csname n@cal_dip_null_20_40\endcsname{2.6}
\expandafter\def\csname n@cal_dip_t3_10_100\endcsname{11.2}
\expandafter\def\csname n@cal_dip_t3_10_40\endcsname{11.1}
\expandafter\def\csname n@cal_dip_t3_20_100\endcsname{16.4}
\expandafter\def\csname n@cal_dip_t3_20_40\endcsname{15.9}
\expandafter\def\csname n@cal_dip_w01d5_10_100\endcsname{69.8}
\expandafter\def\csname n@cal_dip_w01d5_10_40\endcsname{40.4}
\expandafter\def\csname n@cal_dip_w01d5_20_100\endcsname{58.2}
\expandafter\def\csname n@cal_dip_w01d5_20_40\endcsname{31.8}
\expandafter\def\csname n@cal_null_10_40\endcsname{4.4}
\expandafter\def\csname n@cal_null_max100\endcsname{0.8}
\expandafter\def\csname n@cal_t3_10_100\endcsname{31.6}
\expandafter\def\csname n@cal_t3_10_40\endcsname{29.2}
\expandafter\def\csname n@cal_t3_20_100\endcsname{40.4}
\expandafter\def\csname n@cal_t3_20_40\endcsname{36.7}
\expandafter\def\csname n@cal_t3_hi\endcsname{40}
\expandafter\def\csname n@cal_t3_lo\endcsname{29}
\expandafter\def\csname n@cal_w01d5_10_100\endcsname{97.2}
\expandafter\def\csname n@cal_w01d5_10_40\endcsname{58.0}
\expandafter\def\csname n@cal_w01d5_20_100\endcsname{94.4}
\expandafter\def\csname n@cal_w01d5_20_40\endcsname{44.6}
\expandafter\def\csname n@cam_summary\endcsname{succeeds in 38, 15, 1 runs on G2-S at $D=10$, 20 and 30 under protocol A}
\expandafter\def\csname n@cma_g2s_summary\endcsname{23, 17, 20, 13 of 50 runs under protocol A and 50, 50, 50, 50 with restarts}
\expandafter\def\csname n@conc_bbob\endcsname{no significant difference on any of the 96 function--dimension pairs}
\expandafter\def\csname n@disc_bbob\endcsname{the final errors of host and component differ significantly on 0 of the 96 function--dimension pairs and the success counts over the 1440 runs are 901 against 912}
\expandafter\def\csname n@disc_hard\endcsname{Two further limits appeared in Section~\ref{sec:hard}: when the two funnels are elongated along different axes at a condition number of 100 in 20 dimensions, the component succeeds in only 5 of 50 runs although a second membership function is opened in every run; and on a rough double funnel under restarts the detector also separates neighbouring local optima, so that a run does not collapse and restart, which at $D=10$ costs 7 successes against the host.}
\expandafter\def\csname n@disc_time_auto\endcsname{18 to 23}
\expandafter\def\csname n@disc_union\endcsname{composing the two membership functions with the head-count weights of a mixture loses runs, whereas the union, equal shares and a width-weighted sum of the same functions perform alike}
\expandafter\def\csname n@fam_curg\endcsname{852}
\expandafter\def\csname n@fam_curg_d025\endcsname{30}
\expandafter\def\csname n@fam_curg_d05\endcsname{164}
\expandafter\def\csname n@fam_curg_d075\endcsname{291}
\expandafter\def\csname n@fam_curg_d1\endcsname{367}
\expandafter\def\csname n@fam_mf\endcsname{1726}
\expandafter\def\csname n@fam_mf_d025\endcsname{415}
\expandafter\def\csname n@fam_mf_d05\endcsname{429}
\expandafter\def\csname n@fam_mf_d075\endcsname{438}
\expandafter\def\csname n@fam_mf_d1\endcsname{444}
\expandafter\def\csname n@fam_rdhi\endcsname{+59}
\expandafter\def\csname n@fam_rdlo\endcsname{+39}
\expandafter\def\csname n@fam_rdmax\endcsname{+100}
\expandafter\def\csname n@fam_rdmean\endcsname{+49}
\expandafter\def\csname n@fam_rdmin\endcsname{+10}
\expandafter\def\csname n@fam_runs\endcsname{1800}
\expandafter\def\csname n@fam_sig\endcsname{34}
\expandafter\def\csname n@hv_summary\endcsname{succeeds in 50, 50, 50 runs with its univariate optimiser and 50, 50, 50 with full-covariance AMaLGaM}
\expandafter\def\csname n@mf_ert_summary\endcsname{18.5k, 49.0k, 117k, 350k}
\expandafter\def\csname n@multi_curg\endcsname{535}
\expandafter\def\csname n@multi_frac2\endcsname{38}
\expandafter\def\csname n@multi_mf\endcsname{566}
\expandafter\def\csname n@multi_open\endcsname{in every run}
\expandafter\def\csname n@multi_rdhi\endcsname{+6}
\expandafter\def\csname n@multi_rdlo\endcsname{+1}
\expandafter\def\csname n@multi_rdmean\endcsname{+3}
\expandafter\def\csname n@multi_runs\endcsname{900}
\expandafter\def\csname n@multi_sig\endcsname{0}
\expandafter\def\csname n@noloss_cells\endcsname{32}
\expandafter\def\csname n@noloss_conc\endcsname{no loss of successes on single-funnel functions and evaluations within a margin of ten percent in 26 of 32 cells (the other 6, all AMaLGaM on G2-L, reach at most 1.17)}
\expandafter\def\csname n@noloss_counts\endcsname{are identical in 31 of the 32 cells and differ by one run in the other (G2-L, $D=10$, protocol A, CUR-G: 49 against 50, exact McNemar $p=1.00$)}
\expandafter\def\csname n@noloss_frac2\endcsname{6.9}
\expandafter\def\csname n@noloss_gmr_max\endcsname{1.11}
\expandafter\def\csname n@noloss_gmr_min\endcsname{0.99}
\expandafter\def\csname n@noloss_in\endcsname{26}
\expandafter\def\csname n@noloss_inside\endcsname{26 of the 32 90\% intervals inside the margin $[0.90,1.11]$ (outside: G2-L, $D=20$, protocol A, AMaLGaM: [1.06, 1.17]; G2-L, $D=30$, protocol A, AMaLGaM: [1.02, 1.13]; G2-L, $D=50$, protocol A, AMaLGaM: [1.06, 1.15]; G2-L, $D=20$, protocol B, AMaLGaM: [1.06, 1.17]; G2-L, $D=30$, protocol B, AMaLGaM: [1.02, 1.13]; G2-L, $D=50$, protocol B, AMaLGaM: [1.06, 1.15])}
\expandafter\def\csname n@pen_summary\endcsname{over $c\in\{0,\dots,4\}$ the successes on G2-S are 46 to 48 of 50 at $D=10$, 48 to 50 of 50 at $D=20$, 49 to 50 of 50 at $D=30$, while the share of generations with two membership functions on G1 at $D=10$ falls from 42.2\% at $c=0$ to 1.9\% at $c=4$ (9.6\% at the default $c=2$); switching the dip-test variant on with $\delta$ from 0.7 down to 0.3 lowers that share to 7.3\%--1.4\% but also the successes on G2-S to 45 to 47 at $D=10$, 47 to 48 at $D=20$, 42 to 49 at $D=30$; at $D=20$ and 30 the share on G1 is at most 24.4\% for every setting.}
\expandafter\def\csname n@pipe_null_D10\endcsname{9.6}
\expandafter\def\csname n@pipe_null_D20\endcsname{2.9}
\expandafter\def\csname n@pipe_null_D30\endcsname{1.1}
\expandafter\def\csname n@pipe_null_D50\endcsname{0.3}
\expandafter\def\csname n@pipe_null_alpha\endcsname{at $D=20$ in 5.7\% of the generations with $\alpha$ $=1$, 2.9\% of the generations with $\alpha$ in $(1,3]$, 0.0\% of the generations with $\alpha$ in $(3,10)$}
\expandafter\def\csname n@pipe_null_cond\endcsname{9.6\%, 6.5\%, 5.6\% of the generations at $D=10$ and 2.9\%, 1.5\%, 1.3\% at $D=20$ (condition numbers 1, 10 and 100)}
\expandafter\def\csname n@platform_note\endcsname{Apart from the HillVallEA runs, every result file behind the tables and figures of this paper was computed on three Windows servers of a co-author (Intel Core Ultra 9 285K, 24 logical processors each) with the released executables (mingw-w64 gcc 13, \texttt{-std=c99 -O2}), and recomputed independently on a Linux machine (Intel Xeon 2.10\,GHz, gcc 13.3, same flags) from the same job lists and seeds: of the 995 (file, algorithm) cells, 944 (94.9\%) have identical success counts and the largest difference is 4 runs; the mean evaluations of the successful runs differ by 0.1\% in the median and by 4\% at the 90th percentile over the cells where both succeed; the programs are deterministic for a given seed, so the remaining differences come from the floating-point arithmetic and mathematical libraries of the two tool chains. The tables use the Windows results. The calibration of Section~\ref{sec:calib} was run on both platforms, with 162 of 162 identical rates. The HillVallEA runs, the traces of Section~\ref{sec:mech} and the timing of Section~\ref{sec:time} are from the Linux machine.}
\expandafter\def\csname n@popn_g1_small\endcsname{at most 1 of 50 runs}
\expandafter\def\csname n@popn_small_cells\endcsname{N115_D20, N115_D30, N285_D30}
\expandafter\def\csname n@popn_summary\endcsname{At $D=10$ the successes of CUR-G / \ALG{} are 37/48, 50/50, 50/50 for $N=115$, 285 and 510; at $D=20$ the successes of CUR-G / \ALG{} are 0/0, 18/50, 29/50 for $N=115$, 285 and 510; at $D=30$ the successes of CUR-G / \ALG{} are 0/0, 0/0, 6/50 for $N=115$, 285 and 510.}
\expandafter\def\csname n@rot_summary\endcsname{Rotation alone changes nothing, because the functions are spherical: the cells with $\kappa=1$ repeat the unrotated results. Ill-conditioning along the same axes leaves the gain intact: with $\kappa=10$ and 100 the successes of CUR-G / \ALG{} are 23 / 36, 11 / 29, 0 / 46, 2 / 46 at ($D$, $\kappa$) = (10, 10), (10, 100), (20, 10), (20, 100). When the two funnels are elongated along different axes the picture changes with the dimension: at $D=10$ \ALG{} still succeeds in 45 and 32 of 50 runs ($\kappa=10$ and 100), but at $D=20$ in only 29 and 5, where CUR-G succeeds in 0 and 0. A second membership function is opened in every one of these runs (in 14.7\% and 7.7\% of the generations), and every failed run ends in the wider funnel: detecting two groups along one direction is then not enough to keep the deeper funnel, and we report the case as a limit without having traced which step fails. Restricting the candidates to the principal directions (\ALG{}-ax) changes the counts by at most 1 run, so the two-means direction neither helps nor harms here. BIPOP-CMA-ES solves every rotated cell in all 50 runs, as its covariance adapts to either elongation within one run.}
\expandafter\def\csname n@rough_summary\endcsname{On the rough functions with $A=1$ the component raises the runs that end in the global funnel from 37 to 46 at $D=10$ and from 19 to 50 at $D=20$ under protocol A, while the successes to the target go from 21 to 19 and from 15 to 33 (the target lies at the bottom of the global funnel, and ending in that funnel does not mean reaching it); under protocol B it costs successes at $D=10$ (43 against 50 for CUR-G) and raises them at $D=20$ (43 against 40): on a rough landscape the detector also separates two neighbouring local optima of the global funnel, the union keeps both alive, the population does not collapse, and the restart that would have rescued the run never comes. With $A=10$ (the bi-Rastrigin function) the EDAs reach the target in at most 2 of 50 runs; \ALG{} ends in the global funnel in 32, 42 (protocol A) and 27, 41 (protocol B) of 50 runs against 24, 18 and 28, 19 for CUR-G, but the local optima inside the funnel stop every EDA short of the target, whereas IPOP-CMA-ES reaches it in 32 runs at $D=10$ and 10 at $D=20$ through its restarts with large populations.}
\expandafter\def\csname n@sweep_cma_06\endcsname{succeed in 50 (CMA-ES with restarts) and 48 (BIPOP) runs}
\expandafter\def\csname n@sweep_knee\endcsname{0.7}
\expandafter\def\csname n@sweep_mf_06\endcsname{2}
\expandafter\def\csname n@sweep_others_06\endcsname{CUR-G (0), AMaLGaM (0) and \AMF{} (2)}
\expandafter\def\csname n@time_ratio_G1\endcsname{18 to 22}
\expandafter\def\csname n@time_summary\endcsname{On G2-S a generation of \ALG{} takes 22.9, 22.1, 20.6, 18.2 times the user CPU time of a generation of CUR-G at $D=10$, 20, 30 and 50, including the evaluations of the test function, and \AMF{} 27.7, 32.6, 30.1, 26.0 times that of AMaLGaM. The public BIPOP-CMA-ES, with its small populations and lazy eigendecompositions, costs 0.3, 0.8, 1.5, 3.3~$\mu$s per evaluation at $D=10$, 20, 30 and 50; our CMA-ES re-implementation, which recomputes the eigendecomposition in every generation, 4.8, 28, 82, 330~$\mu$s.}
\expandafter\def\csname n@tr_a0\endcsname{38}
\expandafter\def\csname n@tr_a_cg1\endcsname{33}
\expandafter\def\csname n@tr_a_cg5\endcsname{0}
\expandafter\def\csname n@tr_a_cg_zero\endcsname{5}
\expandafter\def\csname n@tr_a_cgl1\endcsname{31}
\expandafter\def\csname n@tr_a_cgl3\endcsname{7}
\expandafter\def\csname n@tr_a_mf5\endcsname{52}
\expandafter\def\csname n@tr_agree\endcsname{93.0}
\expandafter\def\csname n@tr_cg_kept\endcsname{18}
\expandafter\def\csname n@tr_cg_lost\endcsname{32}
\expandafter\def\csname n@tr_fail\endcsname{0}
\expandafter\def\csname n@tr_frac2_g1\endcsname{100}
\expandafter\def\csname n@tr_frac2_g2\endcsname{100}
\expandafter\def\csname n@tr_lost\endcsname{no run}
\expandafter\def\csname n@tr_lost_sentence\endcsname{In no run did $b_t$ fall to zero, and every run reached the target.}
\expandafter\def\csname n@tr_mf_never2\endcsname{none}
\expandafter\def\csname n@tr_ovl_max\endcsname{0.29}
\expandafter\def\csname n@tr_ovl_med\endcsname{0.000}
\expandafter\def\csname n@tr_pA_min\endcsname{0.24}
\expandafter\def\csname n@tr_pA_q1\endcsname{0.51}
\expandafter\def\csname n@tr_pA_q3\endcsname{0.62}
\expandafter\def\csname n@tr_recover\endcsname{none}
\expandafter\def\csname n@tr_reopen\endcsname{50}
\expandafter\def\csname n@tr_rt0\endcsname{0.90}
\expandafter\def\csname n@tr_rt1\endcsname{0.69}
\expandafter\def\csname n@tr_rt2\endcsname{0.45}
\expandafter\def\csname n@tr_rt_cross\endcsname{2}
\expandafter\def\csname n@tr_rtk_cross\endcsname{4}
\expandafter\def\csname n@tr_rtl1\endcsname{0.60}
\expandafter\def\csname n@tr_rtl3\endcsname{0.37}
\expandafter\def\csname n@tr_zero_max\endcsname{8}
\expandafter\def\csname n@tr_zero_med\endcsname{4}
\expandafter\def\csname n@tr_zero_min\endcsname{3}
\expandafter\def\csname n@xp_cells\endcsname{995}
\expandafter\def\csname n@xp_maxdiff\endcsname{4}
\expandafter\def\csname n@xp_same\endcsname{944}

\begin{document}
\title{MF-EDA: Joining Membership Functions to Adapt the Number of Distributions in Gaussian EDAs}
\author{Jun~Zhang,~\IEEEmembership{Fellow,~IEEE,}
        Tong~Qian,
        Hua~Ding,
        Wei-Jie~Yu,
        Xiao-Fang~Liu,
        and~Jian-Yu~Li%
\thanks{This work was supported by the National Research Foundation of Korea under Grants RS-2025-00555463 and RS-2025-25456394. (\emph{Corresponding author: Jun Zhang.})}%
\thanks{Jun Zhang is affiliated with Hanyang University ERICA, Republic of Korea; Nankai University, China; and Chaoyang University of Technology, Taiwan (e-mail: junzhang@ieee.org).}%
\thanks{Tong Qian, Hua Ding, Xiao-Fang Liu, and Jian-Yu Li are with Nankai University, Tianjin, China (e-mail: qiantongnku@\allowbreak 163.com; dinghua\_ieee@\allowbreak 163.com; liuxiaofang@\allowbreak nankai.edu.cn; jianyulics@\allowbreak foxmail.com).}%
\thanks{Wei-Jie Yu is with Sun Yat-sen University, Guangzhou, China (e-mail: yuweijie6@mail.sysu.edu.cn).}}
\maketitle
\begin{abstract}
An estimation-of-distribution algorithm (EDA) estimates one distribution from the selected individuals, usually one Gaussian, peaked where most of them are. On a double funnel the wider funnel supplies most of them, and the deeper one, which holds the global optimum, is lost early. This paper estimates the distribution as a composition of several distributions: the selected individuals are decomposed by Gaussian membership functions, as in a neuro-fuzzy system, each membership function is joined to its own local distribution, and the distributions are composed again when new solutions are generated. In every generation MF-EDA searches the principal and two-means directions for two separated groups, opens one or two membership functions through a calibrated detector, joins each to a local full-covariance Gaussian and samples from the normalised union; the number of distributions, one or two, thus follows the data without fitness evaluations. A membership function has height one, so each funnel receives samples according to its width, not its head count; a two-region model shows how the head-count weights of a mixture compound the deeper funnel's early disadvantage and bounds its survival under the composition, which has a closed-form density. On held-out seeds and paired runs the component raises the successes of a rank-weighted Gaussian EDA from \nm{fam_curg} to \nm{fam_mf} of \nm{fam_runs} on 36 double-funnel settings and, on the 50-dimensional Lunacek double funnel, from \nm{A_G2S_D50_curg} to \nm{A_G2S_D50_mf} of 50 (AMaLGaM: \nm{A_G2S_D50_ama} to \nm{A_G2S_D50_amf}), with no loss on single-funnel functions and the BBOB suite. Composing the same membership functions with head-count weights loses runs; other head-count-free shares do as well as the union. Restarted CMA-ES needs fewer evaluations on smooth double funnels, the method stops helping when the global funnel is much narrower or with more than two funnels, and a generation takes \nm{disc_time_auto} times longer to compute.

\end{abstract}
\begin{IEEEkeywords}
Estimation of distribution algorithm, membership function, neuro-fuzzy system, Gaussian mixture, funnel landscape, multimodal optimisation, model selection.
\end{IEEEkeywords}
\section{Introduction}

\IEEEPARstart{A}{n} estimation-of-distribution algorithm (EDA) replaces the crossover and mutation of a genetic algorithm with a probabilistic model: it selects the better individuals of the current population, fits a distribution to them and samples the next population from that distribution \cite{muhlenbein1996,larranaga2002,larranaga2024}. For continuous variables the distribution is almost always one Gaussian, from the maximum-likelihood Gaussian of EMNA \cite{larranaga2002} to the adapted models of IDEA \cite{bosman2000}, AMaLGaM \cite{bosman2007,bosman2013} and its recent successors \cite{scholman2025}, and recent work on Gaussian EDAs still modifies one Gaussian \cite{wang2023}. One Gaussian has one peak, and the peak sits where most selected individuals are. The place where most selected individuals are is not always the place where the best solutions are.

A double-funnel landscape makes the difference visible. It has two wide basins (funnels) that lie far apart; one holds the global optimum and the other is shallower but wider. Early in a run the wider funnel supplies most of the selected individuals, the fitted Gaussian moves towards it, and the deeper funnel receives fewer samples in every generation until none is left. Lunacek \emph{et al.} \cite{lunacek2008} built such functions and showed that CMA-ES \cite{hansen2001} with its default population finds the global optimum in fewer than 10\% of the runs on the hard variant. Double funnels are not an artefact of test functions. The energy landscape of the 38-atom Lennard-Jones cluster has a narrow, deep funnel that ends in the global minimum and a wider funnel that ends in the second-lowest minimum, and relaxation usually leads into the wider one \cite{doye1999}. Recent landscape studies confirm that a small global funnel is among the features that best predict failure of CMA-ES and differential evolution \cite{mizuta2026}, and that algorithms with fast global information flow become trapped in one funnel \cite{blackwell2023}.

The natural remedy, a Gaussian mixture, has been in EDAs for twenty-five years \cite{bosman2001,thierens2001,lu2005}, and niching EDAs run one model per cluster \cite{maree2017,maree2018,yang2017,liang2018}. These methods differ in how they discover their components, how they allocate the samples among them, how they model each component and what they remember between generations (Section~\ref{sec:related}); what most of them share is that a component receives samples in proportion to the number of selected individuals it holds, either through a fitted mixing weight or through a sub-population whose size follows the cluster. We call this \emph{head-count coupling}. The deeper funnel, which has few individuals early on, then receives few samples, and whether it survives depends on how long its early disadvantage lasts. Section~\ref{sec:analysis} shows, in a model of two regions, that head-count coupling multiplies the odds of the deeper funnel in every generation by its current selection disadvantage, whereas an allocation that depends on the shapes of the components only pays the disadvantage once per generation, and it gives a bound on the probability of losing the funnel under explicit assumptions on the allocation, the coverage and the selection ratio.

The central idea of this paper concerns the estimation of the distribution itself: \emph{the distribution of the selected individuals is not estimated as a single probability distribution but as a composition of several}. The selected individuals are decomposed in the form of membership functions, each membership function is joined to its own local distribution, and the distributions are composed again when new solutions are generated. The construction is taken from neuro-fuzzy systems \cite{jang1993,jang1997}. An adaptive-network fuzzy inference system writes a complicated input--output relation with a few Gaussian membership functions on an input, each joined to its own consequent through the normalised firing strength. A membership function differs from a mixture component in one respect that matters here: it is \emph{normal}, that is, its height is one whatever the number of data points under it. We build the search distribution of an EDA in this way. In every generation the algorithm (i) searches a small set of candidate directions, the principal directions and the direction that joins the two means of the selected individuals, for one along which they form two separated groups; (ii) on that direction describes them by one or by two Gaussian membership functions, the number being decided by the Bayesian information criterion (BIC) \cite{schwarz1978} together with a separation test and a minimum mass; (iii) joins each membership function to a local full-covariance Gaussian by the membership degrees of the individuals; and (iv) samples from the normalised union of the membership functions. With one membership function the model is the ordinary full-covariance Gaussian, and the algorithm is its single-Gaussian baseline. With two, each funnel that the data reveal is represented at full height, and the share of the samples it receives follows its width, not its head count. The number of distributions in use, one or two, is decided from the selected individuals in every generation: a run may keep one throughout, keep two throughout, or alternate as the evidence changes. The model is not a general mixture, and the paper does not claim more. We call the algorithm \ALG{} (membership-function-joined Gaussian EDA).

The membership form is what makes the composition work on a double funnel. Because a membership function has height one, the share of a funnel in the composition is the area under its function and does not follow the number of individuals in it, whereas the mixing weight of a fitted mixture does (Section~\ref{sec:analysis}). An ablation composes the same membership functions by other rules (the fitted weights of the mixture, equal shares, a width-weighted sum, and fitted weights with a minimum share), each with soft and hard assignment and everything else equal, and shows which part of the composition carries the gain.

The contributions are the following.
\begin{enumerate}
\item \emph{Estimation of the distribution as a composition.} The selected individuals are decomposed by one or two Gaussian membership functions on a data-chosen direction, each joined to a local full-covariance Gaussian by membership degrees, and new solutions are generated from the normalised union; the number of distributions is decided from the data in every generation, and with one the algorithm is its single-Gaussian baseline.
\item \emph{An analysis of why the composition keeps the deeper funnel.} A two-region selection model keeps apart the fitted weight of a component, its sampling share and the mass it actually places on a region, and gives a geometric-decay identity for head-count coupling and a survival bound for shape coupling under explicit assumptions (Proposition~\ref{prop:odds}). The normalised union of two membership functions has a closed-form joint density, an exact acceptance rate $1/(1+\omega)$, and label shares that are within twice the overlap $\omega$ of the width shares (Proposition~\ref{prop:union}). Traces of real runs report where the assumptions hold, and the overlap and the acceptance are reported from the runs.
\item \emph{A detector of two groups that is calibrated, not assumed.} The candidate directions include the two-means direction, found and tested on disjoint halves of the selection so that choosing the direction does not bias the test; a candidate is admissible only if it passes a BIC gain with a search allowance, a separation test and a minimum effective mass; and the false-positive and detection rates of the whole pipeline are measured over dimension, sample size, conditioning and imbalance.
\item \emph{A component, tested with paired inference and stated limits.} The same steps are added to two single-Gaussian EDAs, a rank-weighted Gaussian EDA and AMaLGaM, with the integration into AMaLGaM stated mechanism by mechanism, and both are reported. The experiments cover double-funnel families with varying depth, position, width, rotation, conditioning and roughness, the BBOB suite $f_1$--$f_{24}$ through the official COCO code, and public CMA-ES with IPOP and BIPOP restarts among the baselines. They use paired tests on matched initial populations, risk differences with confidence intervals, expected running times with bootstrap intervals, success curves against the budget, and equivalence intervals for the claim of no loss on single-funnel landscapes, and they report the computing cost per generation, the evaluation cost above which the extra time is paid back, and the cases in which the method fails.
\end{enumerate}

Section~\ref{sec:related} states the problem and reviews related work along four axes. Section~\ref{sec:analysis} explains why a model that follows the head count loses the deeper funnel. Section~\ref{sec:method} describes the method. Section~\ref{sec:exp} reports the experiments, Section~\ref{sec:disc} discusses limits, computing time and future work, and Section~\ref{sec:conc} concludes.

\section{Problem and Related Work}\label{sec:related}

\subsection{Problem}
We minimise $f:[l,u]^D\to\mathbb{R}$ within a budget of fitness evaluations, where $D$ is the number of variables; in all experiments $[l,u]=[-5,5]$. The cost of an algorithm is the number of fitness evaluations it uses. A run succeeds when it finds $\vect{x}$ with $f(\vect{x})-f^\ast\le10^{-8}$, where $f^\ast$ is the global minimum. A funnel is a region in which the local minima are arranged so that the better ones lie closer to the funnel bottom \cite{lunacek2008}; a double funnel has two such regions. Lunacek \emph{et al.} \cite{lunacek2008} wrote the simplest one as
\begin{equation}
g_2(\vect{x})=\min\Big\{\sum_{i=1}^{D}(x_i-\mu_1)^2,\; w_2 D+s_2\sum_{i=1}^{D}(x_i-\mu_2)^2\Big\},
\label{eq:g2}
\end{equation}
where the first funnel holds the global optimum at $\mu_1\mathbf{1}$, and the second is lifted by $w_2 D$ and has width factor $s_2$. With $s_2<1$ the second funnel is flatter and wider than the first, and a larger share of the box drains into it. We call the first funnel the \emph{deeper} funnel and the second the \emph{wider} funnel. Rotated, ill-conditioned and rough variants of (\ref{eq:g2}) are defined in Section~\ref{sec:exp}.

\subsection{Gaussian EDAs}
EMNA \cite{larranaga2002} fits one Gaussian to the selected individuals by maximum likelihood. Kern \emph{et al.} \cite{kern2004} compared such models with evolution strategies and found that the variance of the maximum-likelihood Gaussian is not controlled and shrinks from generation to generation; in their experiments clustering the selected individuals into a Gaussian mixture gave no clear advantage. IDEA \cite{bosman2000} and its parameter-free successor AMaLGaM \cite{bosman2007,bosman2013} add adaptive variance scaling (AVS) \cite{grahl2006}, the standard-deviation-ratio trigger (SDR) \cite{bosman2007sdr} and the anticipated mean shift (AMS) to stop premature convergence of the maximum-likelihood Gaussian; AMaLGaM sets its population size from $D$ alone, and its authors state that it is designed for unimodal landscapes \cite{bosman2007}. Its Gaussian sampling is the basis of RV-GOMEA, which learns from fitness differences which variables depend on each other \cite{olieman2021} and now also conditional models for overlapping dependencies \cite{andreadis2024}, and for which Scholman \emph{et al.} \cite{scholman2025} recently introduced incremental estimation of the distribution. These developments refine the dependency structure of one Gaussian. Natural evolution strategies \cite{wierstra2014} update the parameters of a search distribution along the natural gradient of the expected fitness; their authors note that Gaussian mixtures are possible in principle and that the default settings are weak on some highly multimodal functions, for which they suggest larger populations and restarts. Wang \emph{et al.} \cite{wang2023} modify a single Gaussian with an archive of earlier generations, a shift of the sampling mean towards several leading solutions and a triggered shrinkage; they note that a unimodal Gaussian has difficulty on multimodal problems and address it without a mixture. Soloviev \emph{et al.} \cite{soloviev2024} leave the Gaussian assumption altogether and let a structure search decide, node by node of a Bayesian network, between a linear Gaussian and a kernel density estimate; they report that the model is slower to learn than a Gaussian network. The survey of Larra\~naga and Bielza \cite{larranaga2024} lists, as a direction for continuous EDAs, letting the complexity of the model follow the landscape during a run. \ALG{} does so in the smallest possible step, from one distribution to two.

CMA-ES \cite{hansen2001,hansen2016} samples from one Gaussian whose covariance and step size are adapted along evolution paths, and is the usual reference for continuous optimisation. On multimodal functions it relies on a larger population or on restarts with an increasing population (IPOP) \cite{auger2005}, or on two interlaced restart regimes with large and small populations (BIPOP) \cite{hansen2009bipop}. On the BBOB double funnel with Rastrigin-type local optima, $f_{24}$, BIPOP reached the target in 3 of 15 trials at $D=20$ \cite{hansen2009bipop}, and Loshchilov \emph{et al.} \cite{loshchilov2012} obtained a speed-up of two to three by also decreasing the initial step size at each restart. Adapting the population size during a run is an alternative to restarts whose step-size correction can fail on weakly structured multimodal functions \cite{fernando2025}. A current library documents that CMA-ES with its default population is often trapped on the 40-dimensional Rastrigin function \cite{nomura2026}; the restart strategies remain modules of the current modular framework \cite{denobel2026}, and Nomura \emph{et al.} \cite{nomura2024} show that lowering the learning rate has an effect similar to enlarging the population. The mixture in the evolution strategy of He \emph{et al.} \cite{he2020} serves another purpose: all its components are centred at the same point and together approximate one Gaussian cheaply in high dimension. The public implementation of CMA-ES with IPOP and BIPOP restarts is among the baselines of Section~\ref{sec:exp}.

\subsection{Mixture and niching models along four axes}
Methods with more than one component differ along four axes: (A) how the components are \emph{discovered} and how many there are; (B) how the samples are \emph{allocated} among them; (C) how each component is \emph{modelled} locally; and (D) what is \emph{remembered} between generations and whether restarts are part of the method. Table~\ref{tab:axes} places the methods closest to ours on these axes, from our reading of the papers; a dash means that the paper does not make the point explicit or that we could not determine it.

\begin{table*}[t]\centering\caption{Related methods along four axes: (A) discovery of the components, (B) allocation of the samples, (C) local model, (D) memory and restarts.}\label{tab:axes}
\footnotesize\setlength{\tabcolsep}{4pt}\renewcommand{\arraystretch}{1.08}
\begin{tabular}{@{}>{\raggedright\arraybackslash}p{2.45cm}>{\raggedright\arraybackslash}p{5.6cm}>{\raggedright\arraybackslash}p{3.3cm}>{\raggedright\arraybackslash}p{3.1cm}>{\raggedright\arraybackslash}p{2.5cm}@{}}
\toprule
Method & (A) Discovery & (B) Allocation & (C) Local model & (D) Memory, restarts \\
\midrule
Mixture IDEA \cite{bosman2001} & clustering/EM in $\mathbb{R}^D$; number set by the user & mixing weights of the fitted mixture & full Gaussian per cluster & none \\
Leader-clustering IDEA \cite{thierens2001} & leader clustering in $\mathbb{R}^D$; number from a distance threshold & per cluster & Gaussian per cluster & none \\
Lu and Yao \cite{lu2005} & rival-penalised competitive learning in $\mathbb{R}^D$; number learned & per cluster & Gaussian per cluster & none \\
CAMaLGaM \cite{maree2017} & hierarchy of clusters on a nearest-better tree, level chosen by density--fitness correlation; can be one & one AMaLGaM per cluster & AMaLGaM per cluster & -- \\
HillVallEA \cite{maree2018} & hill-valley test between individuals (costs evaluations) & one optimiser per niche & local optimiser per niche & archive of optima, restarts \\
Yang \emph{et al.} \cite{yang2017} & niches of random size, every generation & one niche per group & univariate Gaussian/Cauchy, local search & none \\
Liang \emph{et al.} \cite{liang2018} & clustering by fitness and relative distance & per niche & covariance from recent selections & recent selected sets \\
MAPS \cite{yang2015} & detection of promising areas & per sub-model & several sub-models & -- \\
RS-CMSA \cite{ahrari2017} & sub-populations repelled by taboo regions; number set from the optima wanted & equal-size sub-populations & covariance per sub-population & archive of minima, restarts \\
Le Minh \emph{et al.} \cite{leminh2025} & number fixed in advance & learned mixture weights, annealed & Gaussian per component & none \\
\ALG{} (this paper) & one-dimensional detector on $D+1$ candidate directions; one or two, every generation, no evaluations & width shares of the union, independent of the head count & full covariance with shrinkage, conditional on the detected coordinate & none (stateless) \\
\bottomrule
\end{tabular}
\end{table*}

On axis (A), Bosman and Thierens \cite{bosman2001} introduced mixture distributions into continuous IDEAs with clustering and expectation--maximisation (EM); the mixture IDEA of \cite{thierens2001} uses randomised leader clustering, so that the number of clusters follows from a distance threshold set by the user; Lu and Yao \cite{lu2005} cluster the selected individuals with rival-penalised competitive learning, which learns the number of clusters and does not need it in advance; CAMaLGaM \cite{maree2017} builds in every generation a hierarchy of clusters along a nearest-better tree, scores each level by the rank correlation between density and fitness, and falls back to one cluster when all scores are negative, so its number of components is adaptive and can be one; HillVallEA \cite{maree2018} replaces the hierarchy by the hill-valley test, which pays fitness evaluations for points on the segment between two individuals; the multimodal EDA of Yang \emph{et al.} \cite{yang2017} draws the niche size at random in each generation; Liang \emph{et al.} \cite{liang2018} cluster by fitness and relative distance; MAPS \cite{yang2015} detects promising areas; RS-CMSA \cite{ahrari2017} sets the number of sub-populations from the number of minima wanted; and Le Minh \emph{et al.} \cite{leminh2025} fix the number of components above the expected number of modes. The number of components is thus adaptive in several of these methods; what distinguishes \ALG{} on this axis is that the detection is one-dimensional, between one and two, on a few candidate directions, with no fitness evaluations and no distance threshold, and that its error rates are measured.

On axis (B), the fitted mixtures \cite{bosman2001,leminh2025} sample a component with its learned mixing weight, which is the share of the selected individuals it explains; the niching methods give each niche its own sub-population, whose size follows the cluster in some methods and is equal for all niches in others, as in RS-CMSA \cite{ahrari2017}; and Le Minh \emph{et al.} \cite{leminh2025} temper the mixture by annealing. Equal quotas and tempered or floored weights are also ways of decoupling the allocation from the head count; Section~\ref{sec:abl} compares them with the union, with the rest of the algorithm fixed. On axis (C), the methods that cluster in $\mathbb{R}^D$ fit a full Gaussian per cluster and therefore need many individuals per cluster as $D$ grows; \cite{yang2017} uses univariate models; \ALG{} fits full local Gaussians with a shrinkage that depends on the number of individuals of the component. On axis (D), HillVallEA and RS-CMSA keep archives of optima and restart; \ALG{} carries nothing from one generation to the next apart from the variance multiplier of its host, and the restarts of protocol B in Section~\ref{sec:exp} are applied to all algorithms alike. Two recent reviews of niching \cite{matanga2024,chauhan2025} agree on the open points: most methods need a niche radius or a number of clusters that is hard to set, newer methods try to estimate it from the data, and behaviour above about ten dimensions is largely untested. Niching variants of CMA-ES continue to appear \cite{karunarathne2024}, the successor of RS-CMSA remains the base optimiser of recent competition entries \cite{chauhan2026}, Landais \emph{et al.} \cite{landais2026} run several EDA agents that repel each other in parameter space, for combinatorial problems, and report that too many agents dilute the budget, and Zhang \emph{et al.} \cite{zhang2024} derive an evolutionary algorithm from the denoising step of diffusion models, in which every individual moves towards a weighted average of the better ones. These methods are mostly evaluated on the task of finding many optima, without a report on the single-component case; \ALG{} is evaluated on both.

\subsection{Membership functions and EDAs}
A fuzzy set is defined by a membership function \cite{zadeh1965}. In the adaptive-network fuzzy inference system \cite{jang1993,jang1997} Gaussian membership functions on the inputs produce firing strengths, the normalised firing strengths weight the consequents of the rules, and the parameters of the membership functions and of the consequents are learned from data. Recent work on such systems still fixes the number of rules by hand and studies how distinguishable adjacent membership functions should be \cite{khaled2026}; the separation test of Section~\ref{sec:number} plays a similar role here. An earlier EDA of our group \cite{liu2009} used a fuzzy histogram, a fuzzy partition of each variable, as its univariate model; it does not join membership functions into one distribution and does not choose their number. Among the works we read in full we found no EDA whose search distribution is a union of membership functions joined to local models by membership degrees; our search was not exhaustive, and we make no claim of priority beyond this statement.

\subsection{Funnels in landscape analysis}
Recent landscape studies look at funnels directly. Blackwell \cite{blackwell2023} generates double-funnel landscapes from barrier trees and finds that a downhill walker enters a funnel in proportion to its surface, and that global-best particle swarms and differential evolution reach the optimal basin far less often than methods that slow the flow of information. Mizuta \emph{et al.} \cite{mizuta2026} generate landscapes with diverse local-optima networks and report that the size of the global funnel is among the features most strongly correlated with the success rate of CMA-ES and differential evolution. Thomson \emph{et al.} \cite{thomson2024} record where runs of CMA-ES and differential evolution stall. Benchmark generators now control such global structure on purpose: adding a sphere component to a BBOB function eases it for CMA-ES gradually \cite{vermetten2023}, and problems with a prescribed contrast between global and local optima or prescribed basin sizes can be generated automatically \cite{skvorc2026}. Whether a function has a funnel structure can be predicted from a sample of points with a macro-F1 score above 0.99 on the BBOB functions \cite{seiler2024}, which is consistent with the premise of this paper that a sample of individuals carries enough information to tell one funnel from two. These studies describe when single-model algorithms fail; \ALG{} changes the model.
\section{Why a Head-Count Model Loses the Deeper Funnel}\label{sec:analysis}

This section states, with a model of two regions, what goes wrong when the allocation of samples to a region follows the number of selected individuals in it, and under which assumptions an allocation that is decoupled from the head count keeps the deeper funnel alive. The model is an idealisation: it replaces the fitting of a distribution by a few quantities that describe where the samples go and how selection treats them. Section~\ref{sec:mech} measures these quantities in real runs and reports where the idealisation holds and where it does not.

\subsection{A two-region selection model}
Let the landscape have two funnels, $A$ (deeper, with the global optimum) and $B$ (wider). In generation $t$ the search distribution consists of $J_t$ labelled components; a new individual is drawn with label $k$ with probability $\beta_{k,t}$ (the \emph{sampling share} of component $k$), and an individual drawn with label $k$ falls in $A$ with probability $T_{kA,t}$ (the \emph{coverage} of $A$ by component $k$). The probability that a new individual falls in $A$ is therefore
\begin{equation}
a_t=\sum_{k=1}^{J_t}\beta_{k,t}\,T_{kA,t},
\label{eq:at}
\end{equation}
and the number of new individuals in $A$ among the $N-1$ sampled ones (the elite is copied) is binomial with parameters $N-1$ and $a_t$; in particular no sample falls in $A$ with probability $(1-a_t)^{N-1}$. Three quantities are thus kept apart: the fitted weight of a component, which the model does not use; its sampling share $\beta_{k,t}$, which the algorithm chooses; and the mass $T_{kA,t}$ that the component actually puts on the region.

Selection is described by $q_{A,t}$ and $q_{B,t}$, the probabilities that an individual in $A$ and in $B$ is among the $n=\lfloor\tau N\rfloor$ selected ones; their ratio $r_t=q_{A,t}/q_{B,t}$ is the current advantage ($r_t>1$) or disadvantage ($r_t<1$) of the deeper funnel under selection. In the large-population limit the share of $A$ among the selected individuals is
\begin{equation}
\sigma_{t+1}=\frac{a_tq_{A,t}}{a_tq_{A,t}+(1-a_t)q_{B,t}},\;\;\text{i.e.}\;\;
\frac{\sigma_{t+1}}{1-\sigma_{t+1}}=\frac{a_t}{1-a_t}\,r_t ,
\label{eq:share}
\end{equation}
and for a finite population we take the number $M_{t+1}$ of selected individuals from $A$ to be binomial with parameters $n$ and $\sigma_{t+1}$.

The ratio $r_t$ is not constant during a run. While the samples are spread widely, the funnel that is flatter gives the better fitness values; once the samples have contracted around the funnel bottoms, the funnel that is deeper does. For (\ref{eq:g2}), if the samples of both funnels are spread around their centres with per-coordinate variance $v$, the mean fitness is $Dv$ in $A$ and $w_2D+s_2Dv$ in $B$, so that $B$ is better on average as long as $v>w_2/(1-s_2)$ when $s_2<1$. With $w_2=0.5$ and $s_2=0.87$ the threshold is a standard deviation of $1.96$, and the uniform initial population on $[-5,5]$ has a standard deviation of $2.89$ per coordinate. A run can thus start with a phase in which $r_t<1$, and the deeper funnel has to survive this phase; the traces in Section~\ref{sec:mech} show such a phase.

\subsection{Head-count coupling versus shape coupling}
A Gaussian mixture fitted by maximum likelihood gives each component a mixing weight equal to the share of the selected individuals it explains and samples the component with that weight; with components that cover their regions this means $a_{t+1}=\sigma_{t+1}$, which we call \emph{head-count coupling}. A single Gaussian has no explicit weights. Its mass on $A$ is $T_{1A,t}$, which is not $\sigma_{t}$ in general; but its mean and covariance are weighted averages over the selected individuals, so this mass decreases with the head count of $A$ and vanishes with it. We use the head-count recurrence as the idealisation of both models. The alternative is \emph{shape coupling}: the share $\beta_{k,t}$ is a function of the shape of the components only, not of their head counts.

\begin{proposition}\label{prop:odds}
In the two-region model let $o_t=a_t/(1-a_t)$ be the odds of sampling in $A$.
\begin{enumerate}
\item[(a)] (Head-count coupling.) If $a_{t+1}=\sigma_{t+1}$ for $t=0,\dots,T-1$, then $o_{T}=o_0\prod_{t=0}^{T-1}r_t$. If $r_t\le r<1$ for $t<T_0$, then $o_{T_0}\le o_0\,r^{T_0}$, and the expected number of new individuals in $A$ falls below one as soon as $T_0\ge\ln\!\big((N-1)o_0\big)/\ln(1/r)$.
\item[(b)] (Shape coupling.) If $a_t=b_t$, where $b_t$ does not depend on $\sigma_t$, then $\sigma_{t+1}/(1-\sigma_{t+1})=\frac{b_t}{1-b_t}\,r_t$ in every generation, with no product over earlier generations; if moreover $b_t\ge b_{\min}>0$, then $\sigma_{t+1}\ge b_{\min}r_t/(b_{\min}r_t+1-b_{\min})$.
\item[(c)] (Survival bound.) Suppose that during a phase of $T$ generations the model keeps a component on $A$ whose sampling share and coverage satisfy $\beta_{A,t}\ge\beta_{\min}$ and $T_{AA,t}\ge1-\epsilon$, and that $r_t\ge r_{\min}>0$. Put $a_{\min}=\beta_{\min}(1-\epsilon)$ and $\sigma_{\min}=a_{\min}r_{\min}/(a_{\min}r_{\min}+1-a_{\min})$. Then $\sigma_{t+1}\ge\sigma_{\min}$ throughout the phase, and for every $m<n\sigma_{\min}$
\begin{equation}
\Pr\Big(\min_{t<T}M_{t+1}<m\Big)\le T\exp\!\big(-2n(\sigma_{\min}-m/n)^2\big).
\label{eq:survival}
\end{equation}
\end{enumerate}
\end{proposition}
\begin{proof}
(a) Insert $a_{t+1}=\sigma_{t+1}$ in (\ref{eq:share}): $o_{t+1}=o_tr_t$, and iterate; $a_t\le o_t$ gives the last claim. (b) Insert $a_t=b_t$ in (\ref{eq:share}); the odds on the right are increasing in $b_t$. (c) By (\ref{eq:at}), $a_t\ge\beta_{A,t}T_{AA,t}\ge a_{\min}$, and (\ref{eq:share}) is increasing in $a_t$ and in $r_t$, so $\sigma_{t+1}\ge\sigma_{\min}$. For a binomial variable with parameters $n$ and $\sigma_{t+1}\ge\sigma_{\min}$, Hoeffding's inequality \cite{hoeffding1963} gives $\Pr(M_{t+1}<m)\le\exp(-2n(\sigma_{\min}-m/n)^2)$; a union bound over the $T$ generations gives (\ref{eq:survival}).
\end{proof}

Part (a) says that head-count coupling turns a disadvantage that lasts several generations into a geometric decay of the odds, and in a finite population the decay ends with a generation in which no sample falls in $A$. With $N=285$ ($D=20$), $o_0=1$ and $r=0.6$, the bound in (a) gives $T_0\ge12$ generations. The decay is not irreversible in principle: a Gaussian with non-degenerate tails can put a sample back into $A$ later; but once the component on $A$ has been dropped, such a sample comes from the tail of a Gaussian centred in $B$, it does not contract around the bottom of $A$, and $r_t$ stays below one for it. Section~\ref{sec:mech} counts how often $A$ is recovered after it has been lost.

Part (b) says that a shape-coupled model pays the disadvantage $r_t$ once per generation, not cumulatively, and part (c) turns this into a survival probability under three explicit assumptions: a minimum share, a minimum coverage and a selection ratio bounded away from zero. For instance, with $n=99$ ($D=20$), $\beta_{\min}=0.3$, $\epsilon=0.1$, $r_{\min}=0.5$ and $m=5$ selected individuals, $\sigma_{\min}=0.156$ and the right-hand side of (\ref{eq:survival}) is $0.11\,T$; it is $0.0019\,T$ for $r_{\min}=0.8$. The bound is crude (a union bound over the generations), but it makes the point: the probability of losing the deeper funnel grows at most linearly with the length of the phase, not geometrically. Under head-count coupling no such bound exists, because $\sigma_{\min}$ would have to be zero. The three assumptions are not guaranteed by the two-region model; they are properties of the algorithm that implements the shape coupling. For \ALG{}, Proposition~\ref{prop:union} shows that $\beta_{A,t}$ equals the width share $s_A/(s_A+s_B)$ of the membership function on $A$ up to twice the overlap of the two functions, so $\beta_{\min}$ is the smallest width ratio during the phase. This ratio is not bounded below by construction, since the component on the deeper funnel contracts as its individuals improve; Section~\ref{sec:mech} reports the measured shares, and the ablation of Section~\ref{sec:abl} includes a variant with an explicit minimum share. The coverage $T_{AA,t}$ and the persistence of the detection are measured in the same traces.
\section{Proposed Method}\label{sec:method}

\subsection{Overview}
\begin{figure*}[t]
\centering
\includegraphics[width=0.95\textwidth]{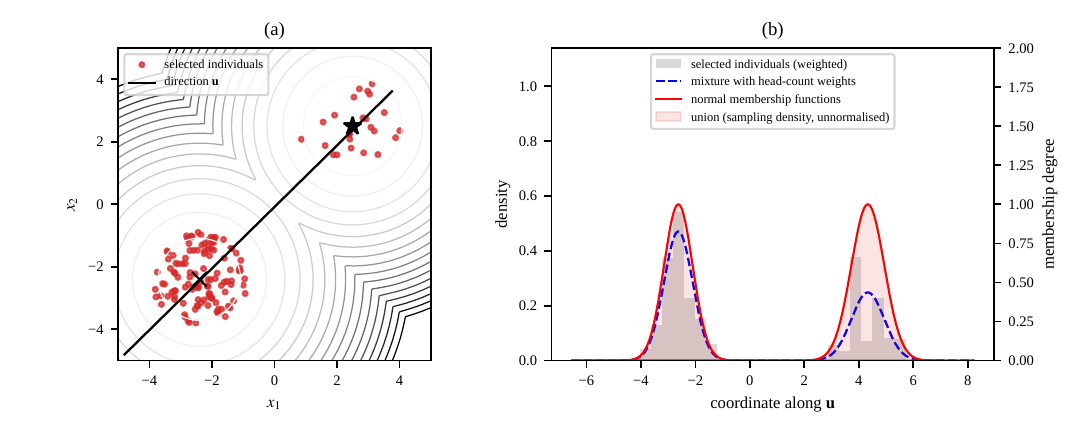}
\caption{Construction of the search distribution on a two-dimensional instance of (\ref{eq:g2}) (schematic). (a) Selected individuals of an early generation, the deeper funnel ($\star$), the wider funnel ($\times$) and the direction $\vect{u}$ found by the search of Section~\ref{sec:dir}. (b) Along $\vect{u}$ the wider funnel holds most of the selected individuals. A maximum-likelihood mixture (dashed) gives each component a mass equal to its head count. The two normal membership functions (solid) have height one; their normalised union is the sampling density, so the deeper funnel receives a share of the samples that follows its width, not its head count.}
\label{fig:principle}
\end{figure*}

\ALG{} is a Gaussian EDA in which four steps are inserted between the estimation of the Gaussian and the sampling: a search over a small set of candidate directions for one along which the selected individuals form two separated groups; a decision, on that direction, between one and two membership functions; the joining of each membership function to a local Gaussian by membership degrees; and the sampling of the new population from the normalised union of the membership functions. Fig.~\ref{fig:principle} shows the construction on a two-dimensional double funnel. The steps follow the layers of an adaptive-network fuzzy inference system \cite{jang1993}: membership functions on an input (layer 1), firing strengths and their normalisation (layers 2 and 3), one consequent per membership function (layer 4) and the combination of the consequents (layer 5). Here the input is the coordinate along the chosen direction, the consequent of a membership function is a local Gaussian in the other directions, and the combination is the sampling of a new individual. The number of membership functions, one or two, is decided anew in every generation; the model does not grow beyond two.

\subsection{Selection and the single Gaussian}
Each generation evaluates a population of $N$ individuals. The best $n=\lfloor\tau N\rfloor$ of them are selected, where $\tau\in(0,1)$ is the selection fraction and $\lfloor\cdot\rfloor$ rounds down. The selected individuals, sorted by fitness as $\vect{x}_1,\dots,\vect{x}_n$, receive the log-rank weights
\begin{equation}
w_i=\frac{\ln(n+\tfrac12)-\ln i}{\sum_{k=1}^{n}\big(\ln(n+\tfrac12)-\ln k\big)},\qquad i=1,\dots,n .
\end{equation}
The weighted mean and covariance are $\vect{m}=\sum_i w_i\vect{x}_i$ and $\mat{C}=\sum_i w_i(\vect{x}_i-\vect{m})(\vect{x}_i-\vect{m})^{\top}$, and $\mat{C}=\mat{V}\mat{\Lambda}\mat{V}^{\top}$ is the eigen-decomposition with eigenvectors $\vect{v}_1,\dots,\vect{v}_D$ and eigenvalues $\lambda_1\ge\dots\ge\lambda_D$ (a ridge of $10^{-12}$ times the mean eigenvalue is added to the diagonal before the decomposition, and eigenvalues below $10^{-30}$ are raised to that value). The eigen-coordinates of individual $i$ are $\vect{z}_i=\mat{V}^{\top}(\vect{x}_i-\vect{m})$ with components $z_{il}$. Sampling $\vect{z}$ with independent components $z_l\sim\mathcal N(0,\alpha\lambda_l)$ and setting $\vect{x}=\vect{m}+\mat{V}\vect{z}$, clipped to the box, gives the single-Gaussian baseline, which we call CUR-G; $\alpha\in[1,10]$ is the adaptive variance scaling of \cite{grahl2006}, divided by $0.9$ when the best fitness improved in the last generation and multiplied by $0.9$ otherwise, and the best individual found so far is copied into the next population. The order of exact ties in fitness is left to the sorting routine, and all random numbers of a run come from one generator seeded by the function, the dimension and the run index, so that CUR-G and \ALG{} start from the same population.

\subsection{Candidate directions}\label{sec:dir}
If the selected individuals come from two funnels, they are bimodal along the line that joins the funnels. The principal directions are natural candidates for this line, but they do not always contain it: a two-dimensional example with two equal components of covariance $\big(\begin{smallmatrix}1.1&-1\\-1&1.2\end{smallmatrix}\big)$ at $\pm(1,1)^{\top}$ has a mixture covariance that is diagonal, so that both principal directions are the coordinate axes, on which the two groups are not separated, whereas along $(1,1)^{\top}/\sqrt2$ they are $2\sqrt2$ apart with standard deviation $0.39$. \ALG{} therefore tests $K=D+1$ candidates: the $D$ principal directions and the \emph{two-means direction}, the normalised difference of the two centres found by a weighted two-means clustering of the selected individuals in the eigen-coordinates (Lloyd's iterations, at most ten, started from the split of the first principal coordinate at its weighted median). The two-means direction is chosen to separate the data, so the test applied to it would be biased if it used the same individuals; \ALG{} finds the direction on the individuals of even rank and tests it on those of odd rank; only when the test passes is the model refitted on all individuals along the now fixed direction, and the refit is kept only if its two centres remain separated by the criterion (\ref{eq:open2}) below. The gain attributed to the two-means direction is that of its test half. The search uses only the selected individuals and costs no fitness evaluations.

\subsection{One or two membership functions}\label{sec:number}
On a candidate direction $\vect{u}$, with $y_i=\vect{u}^{\top}\vect{z}_i$, the detector fits $J$ Gaussians,
\begin{equation}
p_J(y)=\sum_{j=1}^{J}\hat\pi_j\,\mathcal N(y;\hat c_j,\hat s_j^{2}),
\label{eq:em}
\end{equation}
for $J=1$ and $J=2$, where $\hat\pi_j$, $\hat c_j$ and $\hat s_j^{2}$ are the weight, centre and variance of component $j$ and $\mathcal N(y;c,s^2)$ is the normal density with mean $c$ and variance $s^2$. The detector gives every individual the same weight, because bimodality is a property of where the individuals are, not of their ranks; the rank weights enter only the moments of the local models below. The one-component fit is the weighted mean and variance. The two-component fit is by expectation--maximisation (EM) \cite{dempster1977}, started from the weighted quartiles as centres, equal weights and a quarter of the total variance as variances, iterated until the relative change of the log-likelihood is below $10^{-7}$ (at most 100 iterations), with the E-step computed by log-sum-exp, the weights floored at $10^{-12}$ and the variances floored at $10^{-6}$ times the variance of the data on $\vect{u}$ plus $10^{-12}\lambda_1$. A single start can converge to a scale mixture, two components with nearly the same centre and different widths, when a small group lies far from a large one; the likelihood of this solution is high but it is not the location mixture the detector looks for. Therefore, when the solution of the first start beats the allowance of (\ref{eq:open1}) but its centres fail the separation (\ref{eq:open2}), EM is rerun from two further starts with a minority component on either side (centres at the weighted quantiles $0.05$ and $0.55$, respectively $0.45$ and $0.95$, weights $0.1$ and $0.9$), and the separated solution of largest likelihood is kept, if there is one; otherwise the first solution stands. The reruns are rare under one component, because they require a gain above the allowance. The score is
\begin{equation}
\mathrm{BIC}_J=-2n_{\mathrm{eff}}\sum_{i=1}^{n}\tilde w_i\ln p_J(y_i)+(3J-1)\ln n_{\mathrm{eff}},
\label{eq:bic}
\end{equation}
with the detector weights $\tilde w_i$ and the effective sample size $n_{\mathrm{eff}}=1/\sum_i\tilde w_i^2$, which equals $n$ for equal weights; the gain of the direction is $G_{\vect u}=\mathrm{BIC}_1-\mathrm{BIC}_2$. A finite Gaussian mixture is not a regular model under the one-component hypothesis, so $G_{\vect u}$ has no chi-square distribution, and the directions are not independent; the score is used as a ranking statistic, and its false-positive rate is measured, not derived (Section~\ref{sec:calib}).

A candidate is \emph{admissible} when all of the following hold:
\begin{align}
G_{\vect u}&>c\ln K, \label{eq:open1}\\
|\hat c_1-\hat c_2|&>2\sqrt{(\hat s_1^{2}+\hat s_2^{2})/2}, \label{eq:open2}\\
\hat\pi_j\,n_{\mathrm{eff}}&\ge2,\qquad j=1,2, \label{eq:open3}
\end{align}
with $c=2$. Condition (\ref{eq:open1}) is the BIC decision with an allowance for the search over $K$ candidates. Condition (\ref{eq:open2}) asks that the two fitted Gaussians be separated by more than twice their root-mean-square standard deviation, the classical condition for two peaks when weights and variances are equal \cite{ashman1994}. It is not sufficient with unequal weights: the mixture $0.99\,\mathcal N(0,1)+0.01\,\mathcal N(2.1,1)$ satisfies it and has one peak; in the composition of tests such a mixture is stopped by (\ref{eq:open1}), because a component of one percent at that distance does not pay for its three parameters, and, for small selections, by condition (\ref{eq:open3}), which requires an effective mass of at least two individuals per component (Section~\ref{sec:calib} measures the rate at which this mixture is split). A fourth test, a dip of the kernel density estimate between the two centres below a fraction $\delta$ of the lower peak, was part of an earlier version of the component; it lowers the false-positive rate on a single funnel further but also rejects true detections when the two groups are unequal and only moderately separated, which cost the method \nm{abl50_dip_lost} of its \nm{abl50_full} successful runs at $D=50$ (Table~\ref{tab:abl}). It is therefore not part of the method; Section~\ref{sec:abl} reports it as a variant and Section~\ref{sec:calib} its calibration, so that the trade-off is visible. The model uses two membership functions on the admissible candidate with the largest gain, and one membership function when no candidate is admissible. Taking the largest gain among the admissible candidates, rather than testing only the candidate of largest gain, keeps a valid second direction from being masked by a direction with a large gain that fails the separation test. The decision is taken anew in every generation and no state is carried over: the model returns to one membership function as soon as one funnel has emptied or the tests fail.

\subsection{Joining by membership degrees}\label{sec:join}
When two membership functions are opened on direction $\vect{u}$, the membership degree of individual $i$ to function $j$ is the posterior probability of component $j$ in the last E-step of the fit,
\begin{equation}
\rho_{ij}=\frac{\hat\pi_j\,\mathcal N(y_i;\hat c_j,\hat s_j^{2})}{\sum_{k=1}^{2}\hat\pi_k\,\mathcal N(y_i;\hat c_k,\hat s_k^{2})},\qquad y_i=\vect{u}^{\top}\vect{z}_i .
\label{eq:rho}
\end{equation}
It has the form of the normalised firing strength of layer~3 of the adaptive-network fuzzy inference system, except that each Gaussian carries its normalising factor $1/\hat s_j$; it is therefore the responsibility of the fitted mixture. Let $\mat{B}$ be an orthonormal basis of $\mathbb{R}^D$ whose first vector is $\vect{u}$ (completed from the coordinate axes by Gram--Schmidt), and $\vect{t}_i=\mat{B}\vect{z}_i$ the coordinates of individual $i$ in this basis, so that $t_{i1}=y_i$. Each membership function is joined to a local Gaussian with the membership-weighted moments
\begin{equation}
\vect{m}_j=\frac{\sum_iw_i\rho_{ij}\vect{t}_i}{\sum_iw_i\rho_{ij}},\quad
\mat{S}_j=\frac{\sum_iw_i\rho_{ij}(\vect{t}_i-\vect{m}_j)(\vect{t}_i-\vect{m}_j)^{\top}}{\sum_iw_i\rho_{ij}},
\label{eq:conseq}
\end{equation}
a full covariance in the basis $\mat{B}$, estimated from an effective number $n_j=(\sum_iw_i\rho_{ij})^2/\sum_i(w_i\rho_{ij})^2$ of individuals that is a fraction of $n$. The off-diagonal entries of $\mat{S}_j$ are shrunk by the factor $1-\gamma_j$ with $\gamma_j=D/(n_j+D)$, so that a component with few individuals is close to diagonal and a component with many keeps its correlations; the diagonal is floored as in the detector. No individual is assigned to a cluster; each contributes to both local Gaussians in proportion to its membership degrees. The first coordinate of $\vect{m}_j$ and the first diagonal entry of $\mat{S}_j$ are the centre and width of the membership function, $c_j$ and $s_j^2$, and the function is taken normal,
\begin{equation}
\mu_j(y)=\exp\!\Big(-\frac{(y-c_j)^2}{2\alpha_{\mathrm e} s_j^2}\Big),\qquad j=1,2 ,
\label{eq:mf}
\end{equation}
with the effective variance multiplier
\begin{equation}
\alpha_{\mathrm e}=\min\Big\{\alpha,\;\max\Big\{1,\Big(\frac{|c_1-c_2|}{2\bar s}\Big)^{2}\Big\}\Big\},\qquad \bar s^2=\frac{s_1^2+s_2^2}{2}.
\label{eq:cap}
\end{equation}
The cap in (\ref{eq:cap}) keeps the two membership functions separated at the sampling scale: the separation test (\ref{eq:open2}) is made on the unscaled widths, and a multiplier $\alpha$ of up to 10 would otherwise reduce a separation of $2.1\bar s$ to $0.66\bar s$ at sampling time and merge the two functions. The fitted weights $\hat\pi_j$ are used to compute the membership degrees and are then discarded; discarding them is the step that distinguishes the model from a mixture.

\subsection{Sampling from the union}\label{sec:sampling}
A new individual is drawn in three steps. (i) Choose $k\in\{1,2\}$ with probability $s_k/(s_1+s_2)$ and draw $y\sim\mathcal N(c_k,\alpha_{\mathrm e}s_k^2)$. (ii) Accept $y$ with probability $\max_j\mu_j(y)/\sum_j\mu_j(y)$; otherwise return to (i). (iii) For the accepted $y$ draw the other $D-1$ coordinates from the conditional distribution of the local Gaussian of $k$ given its first coordinate, $\vect{t}_{\perp}\sim\mathcal N\big(\vect{m}_{k\perp}+\vect{r}_k(y-c_k),\,\alpha_{\mathrm e}\mat{S}_{k\perp|1}\big)$, where $\vect{r}_k=\mat{S}_{k,\perp1}/s_k^2$ are the regression coefficients of the other coordinates on the first and $\mat{S}_{k\perp|1}=\mat{S}_{k\perp\perp}-\vect{r}_k\mat{S}_{k,1\perp}$ is the Schur complement (its eigenvalues are floored at $10^{-12}\lambda_1$); then set $\vect{z}=\mat{B}^{\top}\vect{t}$ and $\vect{x}=\vect{m}+\mat{V}\vect{z}$, clipped to the box.

\begin{proposition}\label{prop:union}
Let $U(y)=\max_j\mu_j(y)$, $S(y)=\sum_j\mu_j(y)$, $Z=\int U(y)\,dy$, $\omega=\int\min_j\mu_j(y)\,dy\big/Z$, $R_k=\{y:\mu_k(y)\ge\mu_j(y),\,j\ne k\}$, and let $g_k(\vect{t}_\perp\mid y)$ be the conditional density of step (iii).
\begin{enumerate}
\item[(a)] The accepted $y$ has the density $U(y)/Z$, the normalised fuzzy union of the two membership functions.
\item[(b)] Given $y$, the local Gaussian of membership function $k$ is used with probability $\mu_k(y)/S(y)$, the normalised firing strength, and the joint density of a sample is
\begin{equation}
p(y,\vect{t}_\perp)=\frac{U(y)}{Z}\sum_{k=1}^{2}\frac{\mu_k(y)}{S(y)}\,g_k(\vect{t}_\perp\mid y).
\label{eq:joint}
\end{equation}
\item[(c)] A proposal is accepted with probability exactly $1/(1+\omega)\ge\tfrac12$, so that $Z=\sqrt{2\pi\alpha_{\mathrm e}}\,(s_1+s_2)/(1+\omega)$, and the expected number of proposals per sample is $1+\omega\le2$.
\item[(d)] The probability $\beta_k$ that a sample carries label $k$ and the mass $P_k$ of the dominance region $R_k$ are
\begin{equation}
\beta_k=\frac1Z\int\frac{U(y)\mu_k(y)}{S(y)}\,dy,\qquad
P_k=\frac1Z\int_{R_k}\mu_k(y)\,dy ,
\label{eq:beta}
\end{equation}
where $P_k$ is a sum of at most two terms $\sqrt{2\pi\alpha_{\mathrm e}}\,s_k[\Phi(b)-\Phi(a)]/Z$ over the intervals $[c_k+a\sqrt{\alpha_{\mathrm e}}s_k,\,c_k+b\sqrt{\alpha_{\mathrm e}}s_k]$ that make up $R_k$, whose end points are the roots of $(y-c_1)^2/s_1^2=(y-c_2)^2/s_2^2$, and $\Phi$ is the standard normal distribution function.
\item[(e)] $|\beta_k-P_k|\le\omega$ and $|P_k-s_k/(s_1+s_2)|\le\omega$; hence $|\beta_k-s_k/(s_1+s_2)|\le2\omega$. Neither $\beta_k$ nor $P_k$ depends on $\hat\pi_k$.
\item[(f)] For $s_1=s_2=s$, $\omega=(1-\Phi(d/2))/\Phi(d/2)$ with $d=|c_1-c_2|/(\sqrt{\alpha_{\mathrm e}}\,s)$; under the cap (\ref{eq:cap}) $d\ge2$ and $\omega\le0.19$.
\end{enumerate}
\end{proposition}
\begin{proof}
The proposal density of step (i) is $\sum_k\frac{s_k}{s_1+s_2}\mathcal N(y;c_k,\alpha_{\mathrm e}s_k^2)=S(y)/(\sqrt{2\pi\alpha_{\mathrm e}}(s_1+s_2))$. Multiplying by the acceptance probability $U/S$ gives a density proportional to $U$, which is (a). The probability that an accepted proposal came from $k$, given $y$, is $\mu_k(y)/S(y)$, because the acceptance depends on $y$ only; (\ref{eq:joint}) follows, which is (b). The acceptance probability is $\int\frac{S}{\sqrt{2\pi\alpha_{\mathrm e}}(s_1+s_2)}\frac{U}{S}\,dy=Z/(\sqrt{2\pi\alpha_{\mathrm e}}(s_1+s_2))$, and $\int S=\int U+\int\min_j\mu_j=Z(1+\omega)$, which gives (c) because $\omega\le1$. Integrating (\ref{eq:joint}) over $\vect{t}_\perp$ and over $y$ for a fixed $k$ gives $\beta_k$; on $R_k$, $U=\mu_k$, and $\mu_k$ is a scaled normal density, which gives $P_k$ and (d). For (e), $\beta_k-P_k=\frac1Z\big[\int_{R_j}\frac{\mu_j\mu_k}{S}-\int_{R_k}\frac{\mu_k\mu_j}{S}\big]$ with $0\le\mu_1\mu_2/S\le\min_j\mu_j$, so both integrals lie between $0$ and $\omega Z$ and their difference is at most $\omega Z$ in absolute value; and $P_k=\frac1Z\big[\sqrt{2\pi\alpha_{\mathrm e}}\,s_k-\int_{R_j}\mu_k\big]$ with $0\le\int_{R_j}\mu_k\le\int\min_j\mu_j=\omega Z$ and $\sqrt{2\pi\alpha_{\mathrm e}}\,s_k=Z(1+\omega)s_k/(s_1+s_2)$, so $P_k-s_k/(s_1+s_2)$ lies between $-\omega$ and $\omega$. For (f), with equal widths the dominance regions are the half-lines on either side of the midpoint, $\int\min_j\mu_j=2\sqrt{2\pi\alpha_{\mathrm e}}\,s\,(1-\Phi(d/2))$ and $Z=2\sqrt{2\pi\alpha_{\mathrm e}}\,s\,\Phi(d/2)$.
\end{proof}

Proposition~\ref{prop:union} states in which sense the search distribution is a union and not a sum, and what the share of a membership function is. The share that a funnel receives is the label probability $\beta_k$ of its membership function; it is a closed-form integral of normal functions, it is within $2\omega$ of the width share $s_k/(s_1+s_2)$, and it is the width share exactly when the two functions have equal widths or do not overlap. The overlap $\omega$ and the acceptance rate are computed in every generation with two membership functions and reported in Section~\ref{sec:mech}. Along the chosen direction every detected funnel is thus covered by a membership function of height one, and the number of selected individuals in a funnel influences where its membership function lies and how wide it is, not how tall it is. Whether the mass $\beta_k$ falls on the funnel is a separate question, that of the coverage $T_{kA}$ of Section~\ref{sec:analysis}; it depends on how well the local Gaussian fits the funnel and is measured, not derived.

Algorithm~\ref{alg:main} lists one generation. With one membership function steps 7--9 reduce to the sampling of CUR-G, and \ALG{} and CUR-G draw the same random numbers in those generations; the two algorithms therefore coincide, sample by sample, until the first generation in which a candidate is admissible, and differ afterwards.

\begin{algorithm}[t]
\caption{One generation of \ALG{}}\label{alg:main}
\begin{algorithmic}[1]
\STATE evaluate the $N$ individuals; keep the best-so-far $\vect{x}^{\mathrm{best}}$; update $\alpha$
\STATE select the best $n=\lfloor\tau N\rfloor$; weights $w_i$; mean $\vect{m}$; covariance $\mat{C}=\mat{V}\mat{\Lambda}\mat{V}^{\top}$; eigen-coordinates $\vect{z}_i$
\STATE candidates: the $D$ principal directions and the two-means direction found on the individuals of even rank (Section~\ref{sec:dir})
\FOR{each candidate $\vect{u}$}
\STATE fit one and two Gaussians to $y_i=\vect{u}^{\top}\vect{z}_i$ (equal weights; the two-means direction on the individuals of odd rank); $G_{\vect u}\leftarrow\mathrm{BIC}_1-\mathrm{BIC}_2$; test admissibility (\ref{eq:open1})--(\ref{eq:open3})
\ENDFOR
\STATE \textbf{if} some candidate is admissible \textbf{then} take the admissible one with the largest gain (refitted on all individuals if it is the two-means direction); $J\leftarrow2$: membership degrees (\ref{eq:rho}), local Gaussians (\ref{eq:conseq}) with shrinkage, membership functions (\ref{eq:mf}) with the cap (\ref{eq:cap}) \textbf{else} $J\leftarrow1$
\STATE \textbf{if} $J=2$: sample $N-1$ individuals from the union, the coordinate along $\vect{u}$ by rejection and the other coordinates from the conditional local Gaussian of the chosen function (Section~\ref{sec:sampling})
\STATE \textbf{if} $J=1$: sample $N-1$ individuals from $\mathcal N(\vect{m},\alpha\mat{C})$
\STATE add $\vect{x}^{\mathrm{best}}$
\end{algorithmic}
\end{algorithm}

\subsection{Cost}\label{sec:cost}
With $C_f$ the cost of one fitness evaluation and $I$ the number of EM iterations, one generation costs
\begin{equation}
O\big(NC_f+N\log N+(n+N)D^2+D^3+IKn+Kn\log n\big),
\label{eq:cost}
\end{equation}
where the terms are the evaluations, the sort, the covariance and the projections together with the change of coordinates of the samples, the eigen-decomposition, the $K=D+1$ one-dimensional fits and the sorts that start them. The two-means clustering adds $O(nD)$ per iteration, and the local Gaussians, when two functions are open, $O(nD^2+D^3)$ once. The sampling needs $1+\omega\le2$ proposals per individual (Proposition~\ref{prop:union}(c)). No step evaluates the fitness function. The orders of $N$ and $n$ coincide only because $n=\lfloor\tau N\rfloor$ with fixed $\tau$. The order hides a constant: the $K$ fits with $I\le100$ iterations each evaluate up to $200\,nK$ Gaussian densities per generation, which exceeds the $nD^2$ operations of the covariance for all practical $D$; Section~\ref{sec:time} reports the measured time per generation and the total time, and gives the evaluation cost above which the extra time is paid back by the evaluations saved.

\subsection{The steps as a component of AMaLGaM}\label{sec:amf}
Steps 3--7 of Algorithm~\ref{alg:main} use only the selected individuals, their weights and the eigen-decomposition of their covariance, and can be inserted into any Gaussian EDA that has these quantities. We inserted them into AMaLGaM \cite{bosman2007}, whose population size $N=17+3D^{1.5}$, selection fraction $\tau=0.35$ and maximum-likelihood Gaussian with equal weights are kept. In each generation \AMF{} (i) computes AMaLGaM's mean, covariance $\mat{C}$ and distribution multiplier $c$, (ii) runs the direction search and the detector on the eigen-coordinates of the selection with equal weights, (iii) when two functions are open, draws every new individual except the elite from the union of Section~\ref{sec:sampling} with $c$ in the place of $\alpha$ (so that the cap applies to $c$), and otherwise from $\mathcal N(\vect{m},c\mat{C})$ as AMaLGaM does, and (iv) applies AMaLGaM's three mechanisms unchanged: the anticipated mean shift adds $2c(\vect{m}-\vect{m}^{\mathrm{prev}})$, the shift of the global mean, to the first $\lfloor\tau(N-1)/2\rfloor$ new individuals whatever their label; the standard-deviation ratio is the largest coordinate of the average improving individual in the Cholesky coordinates of the global sampling covariance $c\mat{C}$, and $c$ is divided by $0.9$ when it exceeds one; and after a generation without improvement the multiplier is reduced by the factor $0.9$ when it exceeds one, and below one only after $25+D$ such generations, as in \cite{bosman2007}. The mean shift therefore moves a fraction of the samples of both functions in the direction of the global mean, which can work against the allocation early in a run; we kept it because the goal is to report the component on an unchanged host. The algorithm stops an attempt when the population has collapsed or the multiplier has fallen below $10^{-10}$, and under protocol B of Section~\ref{sec:exp} restarts from a new uniform population. When one function is open, \AMF{} and AMaLGaM draw the same random numbers and run identically.

\subsection{Differences from existing methods}
\begin{enumerate}
\item \emph{How the samples are allocated.} Mixture and niching EDAs sample a component with a learned weight or give each cluster a sub-population. \ALG{} samples from the union of normal membership functions, so that the share of a function follows its width (Proposition~\ref{prop:union}). Section~\ref{sec:abl} compares, with everything else fixed, this allocation with the fitted weights, with equal shares, with a width-weighted mixture and with fitted weights under a minimum share, each with soft and with hard assignment.
\item \emph{Where the components are formed.} CAMaLGaM, HillVallEA, the mixture IDEAs and the niching EDAs cluster in the full $D$-dimensional space. \ALG{} detects on one direction among $D+1$ candidates and keeps one local Gaussian per membership function; a one-dimensional detection needs few individuals per group and no pairwise distances, and the local Gaussians are full-covariance with shrinkage, so that correlations within a funnel are kept when the group is large enough to estimate them.
\item \emph{How the number is chosen.} Between one and two, in every generation, from the selected individuals alone, by three tests whose false-positive rate is measured on the actual pipeline (Section~\ref{sec:calib}). CAMaLGaM scores a hierarchy of clusterings by a density--fitness correlation; HillVallEA pays fitness evaluations for the hill-valley test; the multimodal EDA of \cite{yang2017} draws the niche size at random; \cite{leminh2025} fixes the number in advance.
\item \emph{What is claimed.} The claim has two sides: more successes on double-funnel landscapes and no change on single-funnel landscapes, at equal numbers of fitness evaluations; and the ablation of Section~\ref{sec:abl} composes the same membership functions by other rules, so that the part of the composition that carries the gain is identified.
\end{enumerate}
\section{Experiments}\label{sec:exp}

The experiments test the two-sided claim by comparing each algorithm with and without the component (\ALG{} with CUR-G, \AMF{} with AMaLGaM) on matched initial populations: on double-funnel landscapes the component raises the number of runs that reach the global optimum (Sections~\ref{sec:family}, \ref{sec:lung} and~\ref{sec:hard}), and on landscapes with one funnel or one global structure it changes neither the number of successful runs nor, beyond a stated margin, the evaluations they need (Section~\ref{sec:noloss}). Section~\ref{sec:mech} compares the runs with the model of Section~\ref{sec:analysis}; Section~\ref{sec:abl} composes the same membership functions by other rules and removes the other elements one at a time; Section~\ref{sec:calib} calibrates the detector; Section~\ref{sec:popn} varies the population size independently of the dimension; Section~\ref{sec:limits} reports where the method stops helping; and Section~\ref{sec:time} measures the computing time and the evaluation cost at which it is paid back.

\subsection{Setup}\label{sec:setup}
\emph{Programs and platforms.} All algorithms are implemented in C99, with the exceptions of the BBOB functions, which are evaluated through the official COCO C code \cite{hansen2021coco}, the public C implementation of CMA-ES \cite{ccmaes} with IPOP and BIPOP restarts, and HillVallEA, for which we use the authors' public C++ code without modification (repository SCMaree/HillVallEA on GitHub, commit 1a68ab73) with a driver that counts the evaluations itself. The sources, the job lists, the per-run result files (one line per run with the evaluations to each of eight targets, the best-so-far error at seven checkpoints, the final error and the diagnostics of the model) and the scripts that produce every table and figure are released with the paper. \nm{platform_note} The cost measure is the number of fitness evaluations; the computing time of the model is reported in Section~\ref{sec:time}.

\emph{Protocols.} A run succeeds when the best-so-far error is at most $10^{-8}$. Protocol A runs an algorithm once, until the budget is spent or the error falls below $10^{-10}$. Protocol B re-initialises the population uniformly whenever it has collapsed (all fitness values within $10^{-12}$) and continues until the budget is spent, which puts every algorithm on the same footing as the restarting HillVallEA; IPOP and BIPOP restart by their own rules under both protocols. Both are reported.

\emph{Parameters.} \ALG{} and CUR-G use $N=\max\{115,\lfloor17+3D^{1.5}+\tfrac12\rfloor\}$ individuals, that is, 115 at $D=10$, 285 at $D=20$, 510 at $D=30$ and 1078 at $D=50$, and the selection fraction $\tau=0.35$; the lower bound keeps at least 40 selected individuals. The constants of the method, the allowance factor $c=2$, the separation factor 2, the minimum effective mass 2, the shrinkage $\gamma_j=D/(n_j+D)$ and the EM tolerance, are the same in every experiment; Section~\ref{sec:calib} varies $c$ and, for the dip-test variant, its threshold $\delta$ (Table~\ref{tab:pen}). \ALG{} and CUR-G, and likewise \AMF{} and AMaLGaM, start every run from the same initial population and draw the same random numbers until the first generation with two membership functions, so that the comparison of a host with its component is paired; all other comparisons are between independent runs.

\emph{Development and test seeds.} The method was developed on one set of seeds (runs 0--49 of the Lunacek functions, instances 1--15 of BBOB). Two complete sets of test runs preceded the final one, and each led to one change of the detector that was then checked on the development seeds only. The first set (runs 1000--1049) used a single EM start; inspecting its failed runs on G2-S at $D=30$ revealed the scale-mixture solutions, and the reruns of Section~\ref{sec:number} were added (development seeds, $D=30$: 45 to 48 of 50 successes). The second set (runs 3000--3049, 4000--4049 and BBOB instances 16--30) used a detector with the density-dip test of Section~\ref{sec:number}; it succeeded in 33 of 50 runs on G2-S at $D=50$, fewer than the component of the first version of this paper on the same seeds (45), and the diagnosis on the development seeds (13 of 20 runs with the test, 20 of 20 without, every other element unchanged) showed that the test rejected the true detections of the first generations. Because the failed test runs had been inspected both times, the final experiments use a third set of seeds that were never looked at: runs numbered from 5000 for the Lunacek, rotated, rough, swept and multi-funnel functions and for the ablation and sensitivity runs, from 6000 for the double-funnel family, and instances 31--45 of BBOB; the HillVallEA runs are those of the first version of this paper (runs 1000--1049), which are independent of the other algorithms.

\emph{Comparison algorithms.} AMaLGaM with its population size $17+3D^{1.5}$, $\tau=0.35$, AVS, SDR and AMS, re-implemented from \cite{bosman2007}; \AMF{}; the $(\mu/\mu_w,\lambda)$-CMA-ES with the default $\lambda=4+\lfloor3\ln D\rfloor$, $\sigma_0=2$ and the initial mean uniform in the box, re-implemented from \cite{hansen2016}; the public C code of CMA-ES \cite{ccmaes} with the IPOP restart rule (population doubled at each restart) \cite{auger2005} and with the BIPOP rule (large and small restart regimes interlaced) \cite{hansen2009bipop}, each restart with $\sigma_0=2$ and an initial mean uniform in $[-4,4]^D$ (the BBOB convention), until the budget is spent, and, as our CMA-ES re-implementation, without a box constraint on the samples (the box is the initialisation region; the test functions are defined on $\mathbb{R}^D$, the COCO functions carry their own boundary penalty, and the global optimum lies inside the box), whereas the EDAs clip their samples to the box; a re-implementation of CAMaLGaM from \cite{maree2017}, for which we found no public code, with the choices that the paper leaves open made by us (merged edge length equal to the distance between cluster means, inheritance from the cluster with the nearest mean, equal cluster weights, global selection) and the population size of \ALG{}; and the authors' HillVallEA with hill-valley clustering and either its default local optimiser, univariate AMaLGaM, or full-covariance AMaLGaM. The AMaLGaM and CMA-ES re-implementations were checked for convergence and evaluation counts on the single funnel but not line by line against the authors' code; the CAMaLGaM re-implementation represents our reading of the paper and is not evidence about the original method, and the public CMA-ES with restarts is the strong baseline against which the evaluation efficiency of \ALG{} should be judged. One correction to the first version of this paper: its CAMaLGaM re-implementation seeded its generator with an affine function of the run number whose slope coincided with the increment of the generator, so that the random streams of consecutive runs were the same stream shifted by one step and the 50 runs of a cell were strongly correlated; its counts (for example 11 of 50 on G2-S at $D=10$) are therefore not independent evidence. The seeding is corrected here (the state is the run number itself, as in the other programs), and all CAMaLGaM counts in this paper are from the corrected program.

\emph{Statistics.} The primary endpoint is success within the budget. For a host and its component (paired runs) we report the $2\times2$ table of successes, the exact McNemar test on its discordant pairs, and the risk difference with the Agresti--Min 95\% interval \cite{agresti2005}; for independent runs, Fisher's exact test and the Newcombe hybrid-score interval \cite{newcombe1998}. Evaluation efficiency is reported as the expected running time \cite{hansen2021coco}, $\mathrm{ERT}=(\text{evaluations to the target in the successful runs}+\text{evaluations spent in the failed runs})/\text{number of successes}$, with a 95\% bootstrap percentile interval (2000 resamples of the runs), and as success curves against the budget; medians of the successful runs alone are not used for inference, because they are selected. The claim of no loss on single-structure landscapes is tested as equivalence: the risk difference of successes with its interval, and the geometric mean of the ratio of evaluations to the target, component over host, across the runs in which both succeeded, with its 90\% interval, which must lie within $[0.90,1.11]$ (a margin of $\pm10\%$). On BBOB, where the target is seldom reached, the endpoint is the final best-so-far error, compared by the paired signed-rank test \cite{wilcoxon1945} on matched initial populations; the checkpoints are used only for the success curves, because the tests at different checkpoints are dependent. The families for Holm's correction \cite{holm1979} at level $0.05$ are fixed in advance: the 36 settings of the double-funnel family, the 24 cells of the Lunacek functions for each host, the 36 multi-funnel settings, the ablation tests of Table~\ref{tab:abl}, and the 96 function--dimension pairs of BBOB. Clustered summaries over settings (the mean risk difference with a $t$-interval across settings) are reported in addition to pooled counts, because the settings of a family are not homogeneous trials.

\subsection{Double funnels with varying depth gap and position}\label{sec:family}
The first family is (\ref{eq:g2}) in the form used by BBOB $f_{24}$ \cite{hansen2009} without its Rastrigin term and rotation: $\mu_2=-\sqrt{(\mu_1^2-\delta)/s}$ with $s=1-1/(2\sqrt{D+20}-8.2)$, and the second funnel lifted by $\delta D$. The depth gap takes the values $\delta\in\{0.25,0.5,0.75,1\}$, the position of the global funnel $\mu_1\in\{2,2.5,3\}$ and $D\in\{10,20,30\}$: 36 settings, 50 paired runs each, budget $10^4D$, protocol A. Table~\ref{tab:funnel} gives the success counts and the risk differences.

\begin{table*}[t]\centering\caption{Double-funnel family with depth gap $\delta$ and global-funnel position $\mu_1$, budget $10^4D$, 50 paired runs per setting (same initial populations): successful runs of CUR-G / \ALG{}, and the risk difference \ALG{}$-$CUR-G in percentage points with its 95\% interval. $^\dagger$: exact McNemar test significant at $0.05$ after Holm correction over the 36 settings.}\label{tab:funnel}
\small\setlength{\tabcolsep}{4pt}\begin{tabular}{cc ccc}\toprule
$\delta$ & $\mu_1$ & $D=10$ & $D=20$ & $D=30$\\\midrule
0.25 & 2.0 & 7\,/\,34\ +54 [38, 66]$^\dagger$ & 0\,/\,48\ +96 [84, 100]$^\dagger$ & 0\,/\,50\ +100 [90, 103]$^\dagger$\\
0.25 & 2.5 & 9\,/\,39\ +60 [44, 72]$^\dagger$ & 1\,/\,50\ +98 [87, 102]$^\dagger$ & 1\,/\,50\ +98 [87, 102]$^\dagger$\\
0.25 & 3.0 & 10\,/\,44\ +68 [52, 79]$^\dagger$ & 2\,/\,50\ +96 [84, 100]$^\dagger$ & 0\,/\,50\ +100 [90, 103]$^\dagger$\\
0.5 & 2.0 & 25\,/\,40\ +30 [16, 42]$^\dagger$ & 17\,/\,48\ +62 [46, 73]$^\dagger$ & 9\,/\,50\ +82 [67, 91]$^\dagger$\\
0.5 & 2.5 & 25\,/\,46\ +42 [27, 54]$^\dagger$ & 15\,/\,50\ +70 [54, 81]$^\dagger$ & 13\,/\,50\ +74 [58, 84]$^\dagger$\\
0.5 & 3.0 & 23\,/\,45\ +44 [26, 58]$^\dagger$ & 21\,/\,50\ +58 [42, 70]$^\dagger$ & 16\,/\,50\ +68 [52, 79]$^\dagger$\\
0.75 & 2.0 & 35\,/\,46\ +22 [9, 33]$^\dagger$ & 32\,/\,49\ +34 [19, 46]$^\dagger$ & 31\,/\,50\ +38 [23, 50]$^\dagger$\\
0.75 & 2.5 & 33\,/\,46\ +26 [13, 37]$^\dagger$ & 31\,/\,50\ +38 [23, 50]$^\dagger$ & 28\,/\,50\ +44 [28, 56]$^\dagger$\\
0.75 & 3.0 & 36\,/\,47\ +22 [7, 35]$^\dagger$ & 34\,/\,50\ +32 [18, 44]$^\dagger$ & 31\,/\,50\ +38 [23, 50]$^\dagger$\\
1 & 2.0 & 41\,/\,47\ +12 [2, 21] & 40\,/\,49\ +18 [6, 28]$^\dagger$ & 45\,/\,50\ +10 [1, 18]\\
1 & 2.5 & 39\,/\,49\ +20 [8, 31]$^\dagger$ & 38\,/\,50\ +24 [11, 35]$^\dagger$ & 42\,/\,50\ +16 [5, 26]$^\dagger$\\
1 & 3.0 & 39\,/\,49\ +20 [8, 31]$^\dagger$ & 42\,/\,50\ +16 [5, 26]$^\dagger$ & 41\,/\,50\ +18 [6, 28]$^\dagger$\\
\midrule total & & 322\,/\,532 of 600 & 273\,/\,594 of 600 & 257\,/\,600 of 600\\
\bottomrule\end{tabular}\end{table*}

CUR-G reaches the global optimum in \nm{fam_curg} of the \nm{fam_runs} runs and \ALG{} in \nm{fam_mf}; the exact McNemar test is significant in \nm{fam_sig} of the 36 settings after Holm correction, and in no setting is the difference in favour of CUR-G. Treated as 36 clusters, the risk difference \ALG{}$-$CUR-G is \nm{fam_rdmean} percentage points on average, with a 95\% $t$-interval of $[\nm{fam_rdlo},\nm{fam_rdhi}]$ across settings and a range from \nm{fam_rdmin} to \nm{fam_rdmax}. The gain is largest where the single Gaussian is weakest: when the two funnels have nearly the same depth ($\delta=0.25$) CUR-G succeeds in \nm{fam_curg_d025} of 450 runs and \ALG{} in \nm{fam_mf_d025}; when the gap is large ($\delta=1$) the counts are \nm{fam_curg_d1} and \nm{fam_mf_d1}. The 36 settings share the same spherical, unrotated structure and differ in depth, position and dimension, not in mechanism; the rotated, ill-conditioned and rough variants are in Section~\ref{sec:hard}.

\subsection{Lunacek funnels and comparison algorithms}\label{sec:lung}
\begin{figure*}[t]
\centering
\includegraphics[width=0.95\textwidth]{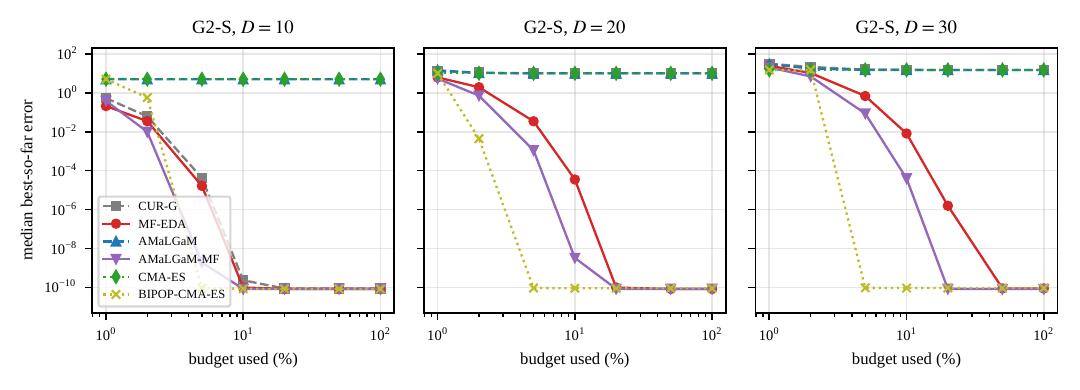}
\caption{Median best-so-far error over 50 runs against the share of the budget used, on G2-S under protocol A. A curve that ends at $D/2$ means that the median run finishes in the sub-optimal funnel.}
\label{fig:curves}
\end{figure*}

\begin{figure*}[t]
\centering
\includegraphics[width=0.95\textwidth]{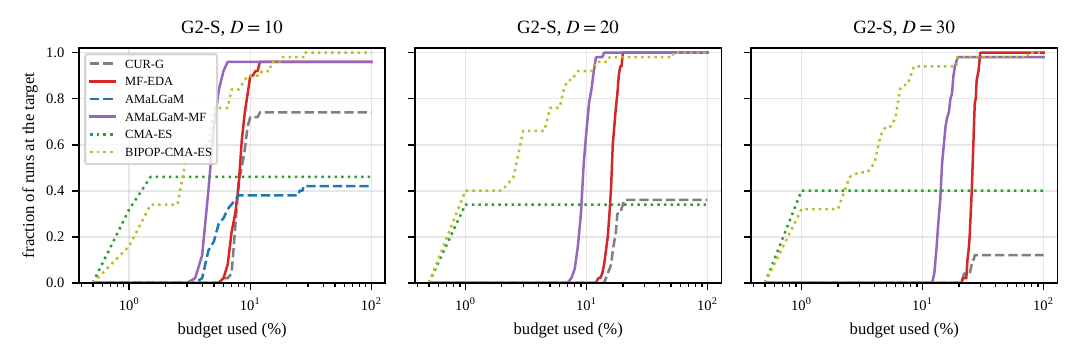}
\caption{Fraction of the 50 runs that have reached the target $10^{-8}$ against the share of the budget used, on G2-S under protocol A (success curves; the value at 100\% is the success count of Table~\ref{tab:lungA}).}
\label{fig:succ}
\end{figure*}

The second family uses the original parameters of \cite{lunacek2008}: $\mu_1=2.5$, $\mu_2=-2.398179$, $w_2=0.5$, and $s_2=0.87$ (G2-S, the global funnel is the narrower one) or $s_2=1.15$ (G2-L, the global funnel is the wider one), together with the single funnel G1 $=\sum_i(x_i-2.5)^2$; 50 runs, budget $15\,000D$, $D\in\{10,20,30,50\}$. Tables~\ref{tab:lungA} and~\ref{tab:lungB} give the success counts and the expected running times under the two protocols, Table~\ref{tab:lungpair} the paired comparisons of each host with its component, Fig.~\ref{fig:curves} plots the median best-so-far error against the budget used on G2-S, and Fig.~\ref{fig:succ} the fraction of runs that have reached the target against the budget.

\begin{table*}[t]\centering\caption{Lunacek G-type funnels, protocol A (no restart): successful runs out of 50 (error $10^{-8}$ within $1.5\times10^4D$ evaluations) and, in parentheses, the expected running time ERT in thousands of evaluations ($\infty$ when no run succeeded). \AMF{}\,(v2) is the component of the first version of this paper on the same host; IPOP and BIPOP are the public CMA-ES with restarts. A dash marks a dimension at which the algorithm was not run.}\label{tab:lungA}
\scriptsize\setlength{\tabcolsep}{1.8pt}\begin{tabular}{llccccccccc}\toprule
Function & $D$ & CUR-G & \ALG{} & AMaLGaM & \AMF{}\,(v2) & \AMF{} & CMA-ES & IPOP & BIPOP & CAMaLGaM-re\\\midrule
G1 (single) & 10 & 50 (12.1) & 50 (12.3) & 50 (6.6) & 50 (6.6) & 50 (6.8) & 50 (1.5) & 50 (1.5) & 50 (1.5) & 50 (7.3)\\
 & 20 & 50 (51.6) & 50 (52.2) & 50 (27.5) & 50 (27.6) & 50 (28.5) & 50 (2.8) & 50 (2.8) & 50 (2.8) & 50 (30.9)\\
 & 30 & 50 (123) & 50 (122) & 50 (63.3) & 50 (63.6) & 50 (63.0) & 50 (4.1) & 50 (4.1) & 50 (4.1) & 50 (67.1)\\
 & 50 & 50 (341) & 50 (347) & 50 (171) & 50 (171) & 50 (173) & 50 (6.4) & 50 (6.3) & 50 (6.3) & --\\
\addlinespace
G2-L (double, global wider) & 10 & 49 (14.8) & 50 (12.0) & 50 (6.7) & 50 (6.6) & 50 (6.9) & 33 (2.7) & 50 (4.7) & 50 (3.7) & 50 (7.5)\\
 & 20 & 50 (49.0) & 50 (50.2) & 50 (26.7) & 50 (28.0) & 50 (29.6) & 30 (5.6) & 50 (7.9) & 50 (5.9) & 50 (28.7)\\
 & 30 & 50 (114) & 50 (116) & 50 (60.1) & 50 (62.8) & 50 (64.4) & 31 (7.9) & 50 (11.3) & 50 (8.9) & 50 (64.4)\\
 & 50 & 50 (329) & 50 (344) & 50 (169) & 50 (180) & 50 (186) & 36 (10.0) & 50 (13.6) & 50 (11.4) & --\\
\addlinespace
G2-S (double, global narrower) & 10 & 37 (64.8) & 48 (18.5) & 21 (25.7) & 48 (7.5) & 48 (7.4) & 23 (4.2) & 50 (8.8) & 50 (7.1) & 38 (12.6)\\
 & 20 & 18 (584) & 50 (49.0) & 0 ($\infty$) & 49 (29.8) & 50 (28.9) & 17 (10.9) & 46 (39.6) & 50 (13.2) & 15 (130)\\
 & 30 & 6 (3409) & 50 (117) & 0 ($\infty$) & 47 (70.1) & 49 (68.9) & 20 (13.3) & 46 (64.3) & 50 (25.1) & 1 (4572)\\
 & 50 & 0 ($\infty$) & 49 (350) & 0 ($\infty$) & 41 (224) & 49 (185) & 13 (32.6) & 46 (121) & 50 (36.3) & --\\
\addlinespace
\bottomrule\end{tabular}\end{table*}

\begin{table*}[t]\centering\caption{Lunacek G-type funnels, protocol B (restart after collapse; the restart strategies of IPOP and BIPOP are their own): successful runs out of 50 (error $10^{-8}$ within $1.5\times10^4D$ evaluations) and, in parentheses, the expected running time ERT in thousands of evaluations ($\infty$ when no run succeeded). \AMF{}\,(v2) is the component of the first version of this paper on the same host; IPOP and BIPOP are the public CMA-ES with restarts; HVEA-u and HVEA-f are HillVallEA with its univariate and its full-covariance local optimiser. A dash marks a dimension at which the algorithm was not run.}\label{tab:lungB}
\scriptsize\setlength{\tabcolsep}{1.8pt}\begin{tabular}{llccccccccccc}\toprule
Function & $D$ & CUR-G & \ALG{} & AMaLGaM & \AMF{}\,(v2) & \AMF{} & CMA-ES & IPOP & BIPOP & CAMaLGaM-re & HVEA-u & HVEA-f\\\midrule
G1 (single) & 10 & 50 (12.1) & 50 (12.3) & 50 (6.6) & 50 (6.6) & 50 (6.8) & 50 (1.5) & 50 (1.5) & 50 (1.5) & 50 (7.3) & 50 (2.3) & 50 (7.9)\\
 & 20 & 50 (51.6) & 50 (52.2) & 50 (27.5) & 50 (27.6) & 50 (28.5) & 50 (2.8) & 50 (2.8) & 50 (2.8) & 50 (30.9) & 50 (5.7) & 50 (30.6)\\
 & 30 & 50 (123) & 50 (122) & 50 (63.3) & 50 (63.6) & 50 (63.0) & 50 (4.1) & 50 (4.1) & 50 (4.1) & 50 (67.1) & 50 (9.1) & 50 (61.7)\\
 & 50 & 50 (341) & 50 (347) & 50 (171) & 50 (171) & 50 (173) & 50 (6.4) & 50 (6.3) & 50 (6.3) & -- & -- & --\\
\addlinespace
G2-L (double, global wider) & 10 & 50 (12.2) & 50 (12.0) & 50 (6.7) & 50 (6.6) & 50 (6.9) & 50 (3.1) & 50 (4.7) & 50 (3.7) & 50 (7.5) & 50 (3.2) & 50 (11.2)\\
 & 20 & 50 (49.0) & 50 (50.2) & 50 (26.7) & 50 (28.0) & 50 (29.6) & 50 (5.5) & 50 (7.9) & 50 (5.9) & 50 (28.7) & 50 (7.3) & 50 (36.6)\\
 & 30 & 50 (114) & 50 (116) & 50 (60.1) & 50 (62.8) & 50 (64.4) & 50 (7.9) & 50 (11.3) & 50 (8.9) & 50 (64.4) & 50 (10.5) & 50 (70.1)\\
 & 50 & 50 (329) & 50 (344) & 50 (169) & 50 (180) & 50 (186) & 50 (10.8) & 50 (13.6) & 50 (11.4) & -- & -- & --\\
\addlinespace
G2-S (double, global narrower) & 10 & 50 (18.4) & 50 (13.2) & 50 (19.9) & 50 (7.4) & 50 (7.3) & 50 (4.0) & 50 (8.8) & 50 (7.1) & 50 (13.1) & 50 (3.9) & 50 (13.9)\\
 & 20 & 46 (143) & 50 (49.0) & 21 (563) & 50 (29.7) & 50 (28.9) & 50 (10.0) & 46 (39.6) & 50 (13.2) & 45 (145) & 50 (10.0) & 50 (49.7)\\
 & 30 & 13 (1512) & 50 (117) & 1 (22487) & 50 (69.8) & 50 (68.9) & 50 (16.0) & 46 (64.3) & 50 (25.1) & 7 (2966) & 50 (19.2) & 50 (131)\\
 & 50 & 0 ($\infty$) & 49 (350) & 0 ($\infty$) & 49 (233) & 50 (185) & 50 (33.4) & 46 (121) & 50 (36.3) & -- & -- & --\\
\addlinespace
\bottomrule\end{tabular}\end{table*}

On G2-S under protocol A the component raises the successes of CUR-G from \nm{A_G2S_D10_curg}, \nm{A_G2S_D20_curg}, \nm{A_G2S_D30_curg} and \nm{A_G2S_D50_curg} of 50 at $D=10$, 20, 30 and 50 to \nm{A_G2S_D10_mf}, \nm{A_G2S_D20_mf}, \nm{A_G2S_D30_mf} and \nm{A_G2S_D50_mf}, and those of AMaLGaM from \nm{A_G2S_D10_ama}, \nm{A_G2S_D20_ama}, \nm{A_G2S_D30_ama} and \nm{A_G2S_D50_ama} to \nm{A_G2S_D10_amf}, \nm{A_G2S_D20_amf}, \nm{A_G2S_D30_amf} and \nm{A_G2S_D50_amf}; the risk differences and their intervals are in Table~\ref{tab:lungpair}, and the expected running times fall by the factors given there. Under protocol B the hosts recover part of the gap through restarts, and the component still \nm{B_summary}. On G1 and G2-L, where one funnel dominates, the component leaves the success counts and, within the stated margin, the evaluations of its host unchanged (Section~\ref{sec:noloss}).

Among the other algorithms, the public CMA-ES with restarts is the most efficient on these smooth double funnels: BIPOP reaches the target in \nm{bipop_g2s_summary}, with expected running times of \nm{bipop_ert_summary}, against \nm{mf_ert_summary} for \ALG{}, which has no restarts and a population of a different order; the single-run CMA-ES without restarts succeeds in \nm{cma_g2s_summary}. The comparison with CMA-ES is one of mechanisms rather than of evaluation counts: \ALG{} keeps the deeper funnel within one population and one run, whereas restarts find it by starting again with a larger population. Our re-implementation of CAMaLGaM \nm{cam_summary}, and HillVallEA under protocol B \nm{hv_summary}; we draw no conclusion about the original CAMaLGaM from the re-implementation.

\subsection{Comparison with the two-region model}\label{sec:mech}
\begin{figure*}[t]
\centering
\includegraphics[width=0.98\textwidth]{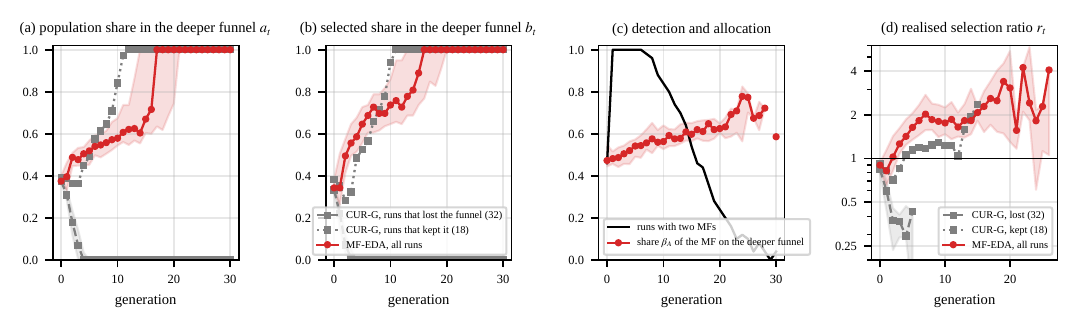}
\caption{Traces of 50 runs on G2-S, $D=20$, protocol A (median and quartiles over the runs; a generation is shown while at least three runs contribute; the CUR-G runs are split into those that lost the deeper funnel and those that kept it). (a) Share of the population in the deeper funnel, $a_t$. (b) Share of the selected individuals in the deeper funnel, $b_t$. (c) Fraction of the runs with two membership functions and, over those runs, the sampling share $\beta_A$ of the membership function on the deeper funnel. (d) Realised selection ratio $r_t$; values below one are the phase of disadvantage.}
\label{fig:mech}
\end{figure*}

Fig.~\ref{fig:mech} traces the quantities of Section~\ref{sec:analysis} in the 50 runs on G2-S at $D=20$ under protocol A, for CUR-G and \ALG{} from the same initial populations. The region of a point is the branch of (\ref{eq:g2}) that is active there, the funnel whose expression is the smaller; it is not the nearest centre. Panel (a) shows the share $a_t$ of the population in the deeper funnel, panel (b) the share $b_t$ of the selected individuals, panel (c) the fraction of runs in which two membership functions are open and the sampling share $\beta_A$ of the function whose members are mostly in the deeper funnel (the label probability of Proposition~\ref{prop:union}, computed from the fitted widths), and panel (d) the realised selection ratio $r_t$, the odds of the deeper funnel among the selected individuals divided by its odds among the sampled ones.

The traces show the mechanism of Proposition~\ref{prop:odds}. The initial population puts about \nm{tr_a0}\% of its individuals in the deeper funnel, and the realised selection ratio is below one in the first generations for both algorithms (median \nm{tr_rt0}, \nm{tr_rt1} and \nm{tr_rt2} in generations 0--2 for CUR-G): the wider funnel gives the better fitness values while the population is spread. CUR-G ends in the wider funnel in \nm{tr_cg_lost} of the 50 runs. In these runs the share $a_t$ decays geometrically, from \nm{tr_a_cgl1}\% at generation 1 to \nm{tr_a_cgl3}\% at generation 3, and reaches zero at generation \nm{tr_zero_med} in the median run (range \nm{tr_zero_min} to \nm{tr_zero_max}), while the selection ratio stays well below one (median \nm{tr_rtl1} at generation 1 and \nm{tr_rtl3} at generation 3); \nm{tr_recover} of them puts a sample back into the deeper funnel afterwards. In the other \nm{tr_cg_kept} runs the deeper funnel happens to hold enough of the selection for the ratio to rise above one by generation \nm{tr_rtk_cross}, and the single Gaussian then contracts into it. Under \ALG{} the second membership function is open in \nm{tr_frac2_g1}\% of the runs at generation 1 and \nm{tr_frac2_g2}\% at generation 2, the share $\beta_A$ of its function lies between \nm{tr_pA_q1} and \nm{tr_pA_q3} (quartiles over all generations with two functions; minimum \nm{tr_pA_min}), and $a_t$ rises to \nm{tr_a_mf5}\% by generation 5 while the selection ratio crosses one at generation \nm{tr_rt_cross} in the median run, because the samples of the deeper funnel contract around its bottom; from then on $b_t$ grows and the deeper funnel takes over the selected set. The labels of the membership functions agree with the active branches of the selected individuals in \nm{tr_agree}\% of the cases, and the overlap $\omega$ of the two functions at the sampling scale has a median of \nm{tr_ovl_med} (maximum \nm{tr_ovl_max}), so the label shares are within a few percent of the width shares. The assumptions of Proposition~\ref{prop:odds}(c) thus hold in these runs with $\beta_{\min}\approx\nm{tr_pA_min}$ and a coverage close to one. The detection, however, is not persistent: the second function closed and reopened at least once in \nm{tr_reopen} of the 50 runs, which costs a generation of single-Gaussian sampling each time. \nm{tr_lost_sentence}

\subsection{Ablation: how the membership functions are composed}\label{sec:abl}
Table~\ref{tab:abl} changes one element of \ALG{} at a time on G2-S under protocol A, with the detector, the fitted components, the initial populations and the random streams unchanged, so that the comparison with the full method is paired. The first block keeps the two membership functions and their local Gaussians and changes only the rule by which they are composed at sampling time, crossed with two assignments. The rules are the union with width shares (the full method), the fitted weights $\hat\pi_j$ of the mixture (head-count coupling), equal shares, a width-weighted mixture, which samples the component $k$ with probability $s_k/(s_1+s_2)$ but without the rejection step, so that the density is the sum $S(y)$ instead of the union $U(y)$, and the fitted weights floored at $1/4$; for the rules other than the width shares the union is taken of the share-weighted functions $\pi_k\mathcal N(y;c_k,\alpha_{\mathrm e}s_k^2)$, which reduces to the union of Section~\ref{sec:sampling} for the width shares. The assignment is soft (membership degrees $\rho_{ij}$ in the local moments) or hard (each individual counted only in the function of larger membership). The second block changes the detector or the local model.

\begin{table*}[t]\centering\caption{Ablation on G2-S, protocol A, 50 paired runs per cell (same initial populations and random streams as the full method; the detector, the fitted components and the budget are the same in every row, only the named element changes): successful runs (ERT in thousands) and the risk difference variant$-$full in percentage points with its 95\% interval. $^\dagger$: exact McNemar test significant at $0.05$ after Holm correction over all 60 tests. The first block crosses the rule by which the two membership functions are composed with the assignment (soft membership degrees, or hard labels given by the larger membership degree); the second block changes one element of the detector or of the local model.}\label{tab:abl}
\footnotesize\setlength{\tabcolsep}{3pt}\begin{tabular}{l cccc cccc}\toprule
 & \multicolumn{4}{c}{successes (ERT)} & \multicolumn{4}{c}{$\Delta$ successes vs full}\\\cmidrule(lr){2-5}\cmidrule(lr){6-9}
Variant & $D=10$ & $D=20$ & $D=30$ & $D=50$ & $D=10$ & $D=20$ & $D=30$ & $D=50$\\\midrule
\ALG{} (union, width shares, soft) & 48 (18.5) & 50 (49.0) & 50 (117) & 49 (350) &  &  &  & \\
\multicolumn{9}{l}{\emph{Composition rule $\times$ assignment}}\\
fitted weights $\hat\pi_j$ (head count), soft & 45 (29.2) & 49 (57.0) & 48 (137) & 48 (363) & -6 [-13, 2] & -2 [-7, 3] & -4 [-10, 3] & -2 [-7, 3]\\
fitted weights, hard & 45 (28.9) & 48 (64.1) & 49 (125) & 48 (363) & -6 [-13, 2] & -4 [-10, 3] & -2 [-7, 3] & -2 [-7, 3]\\
equal shares, soft & 49 (15.3) & 50 (49.3) & 50 (115) & 49 (349) & +2 [-3, 7] & +0 [-4, 4] & +0 [-4, 4] & +0 [-4, 4]\\
equal shares, hard & 49 (15.8) & 50 (49.6) & 50 (116) & 49 (347) & +2 [-3, 7] & +0 [-4, 4] & +0 [-4, 4] & +0 [-4, 4]\\
width-weighted mixture (sum), soft & 48 (18.2) & 50 (49.3) & 50 (117) & 49 (352) & +0 [-4, 4] & +0 [-4, 4] & +0 [-4, 4] & +0 [-4, 4]\\
width-weighted mixture, hard & 48 (18.3) & 50 (49.0) & 50 (120) & 49 (347) & +0 [-4, 4] & +0 [-4, 4] & +0 [-4, 4] & +0 [-4, 4]\\
fitted weights floored at $1/4$, soft & 47 (21.5) & 50 (50.3) & 50 (116) & 49 (354) & -2 [-7, 3] & +0 [-4, 4] & +0 [-4, 4] & +0 [-4, 4]\\
fitted weights floored at $1/4$, hard & 48 (18.8) & 50 (50.6) & 50 (117) & 49 (346) & +0 [-4, 4] & +0 [-4, 4] & +0 [-4, 4] & +0 [-4, 4]\\
union, hard assignment & 48 (18.8) & 50 (49.9) & 50 (117) & 49 (352) & +0 [-4, 4] & +0 [-4, 4] & +0 [-4, 4] & +0 [-4, 4]\\
\multicolumn{9}{l}{\emph{Detector and local model}}\\
principal directions only (no two-means direction) & 47 (22.2) & 50 (49.8) & 50 (117) & 49 (351) & -2 [-7, 3] & +0 [-4, 4] & +0 [-4, 4] & +0 [-4, 4]\\
with density-dip test ($\delta=0.5$) & 47 (21.9) & 48 (62.9) & 47 (146) & 34 (681) & -2 [-7, 3] & -4 [-10, 3] & -6 [-13, 2] & -30 [-42, -16]$^\dagger$\\
rank weights in the detector & 47 (21.9) & 50 (49.6) & 50 (116) & 49 (353) & -2 [-7, 3] & +0 [-4, 4] & +0 [-4, 4] & +0 [-4, 4]\\
diagonal local Gaussians (no correlations) & 48 (18.2) & 50 (49.4) & 50 (114) & 49 (351) & +0 [-4, 4] & +0 [-4, 4] & +0 [-4, 4] & +0 [-4, 4]\\
no cap of $\alpha$ at the sampling scale & 48 (18.6) & 50 (48.4) & 50 (115) & 49 (350) & +0 [-4, 4] & +0 [-4, 4] & +0 [-4, 4] & +0 [-4, 4]\\
single-start EM, 40 iterations & 48 (18.4) & 50 (50.8) & 47 (143) & 24 (1148) & +0 [-4, 4] & +0 [-4, 4] & -6 [-13, 2] & -50 [-62, -34]$^\dagger$\\
\bottomrule\end{tabular}\end{table*}

\nm{abl_alloc_summary} \nm{abl_union_verdict} On G2-S the phase of disadvantage lasts only a few generations (Section~\ref{sec:mech}), so even the head-count weights keep the deeper funnel in most runs once it has been detected; the difference is largest in the expected running time at $D=10$. Table~\ref{tab:ablx} repeats the comparison where the phase lasts longer: the flatter wider funnels of the width sweep ($s_2=0.8$, 0.7 and 0.6) and the smallest depth gap of the double-funnel family ($\delta=0.25$), at $D=20$. \nm{ablx_summary} This is the pattern that Proposition~\ref{prop:odds} predicts: the longer the disadvantage, the more a share that follows the head count costs, and any share that does not follow it, the union or equal shares, performs alike. Among the detector and model elements, \nm{abl_det_summary}. \nm{abl50_summary}

\begin{table}[t]\centering\caption{Composition rules where the phase of disadvantage lasts longer: the width sweep of Fig.~\ref{fig:sweep} and the $\delta=0.25$ settings of Table~\ref{tab:funnel}, $D=20$, protocol A, 50 paired runs per cell. Successes (ERT in thousands) of CUR-G, of \ALG{} (union) and of the variants with fitted weights (head count) and with equal shares, and the difference in successes of each variant against the union with its 95\% interval; $\dagger$: exact McNemar test significant at 0.05 after Holm correction over the 12 tests.}\label{tab:ablx}
\scriptsize\setlength{\tabcolsep}{2.5pt}\begin{tabular}{@{}l cc cc cc@{}}\toprule
 & & & \multicolumn{2}{c}{fitted weights} & \multicolumn{2}{c}{equal shares}\\\cmidrule(lr){4-5}\cmidrule(lr){6-7}
Setting & CUR-G & union & succ. (ERT) & $\Delta$ vs union & succ. (ERT) & $\Delta$ vs union\\\midrule
\multicolumn{7}{@{}l}{\emph{G2-S, width ratio $s_2$ (Fig.~\ref{fig:sweep})}}\\
$s_2=0.8$ & 4 (3498) & 44 (90.1) & 38 (145) & -12 [-21, -2] & 44 (90.8) & +0 [-4, 4]\\
$s_2=0.7$ & 0 ($\infty$) & 24 (375) & 17 (633) & -14 [-23, -3] & 24 (375) & +0 [-4, 4]\\
$s_2=0.6$ & 0 ($\infty$) & 2 (7250) & 1 (14753) & -2 [-7, 3] & 2 (7254) & +0 [-4, 4]\\
\multicolumn{7}{@{}l}{\emph{double-funnel family, $\delta=0.25$ (Table~\ref{tab:funnel})}}\\
$\mu_1=2$ & 0 ($\infty$) & 48 (58.3) & 38 (112) & -20 [-31, -8]$^\dagger$ & 48 (58.1) & +0 [-4, 4]\\
$\mu_1=2.5$ & 1 (9853) & 50 (51.1) & 44 (77.0) & -12 [-21, -2] & 50 (50.2) & +0 [-4, 4]\\
$\mu_1=3$ & 2 (4853) & 50 (50.8) & 44 (77.7) & -12 [-21, -2] & 50 (49.9) & +0 [-4, 4]\\
\bottomrule\end{tabular}\end{table}

The cost of the component on a single funnel is in Table~\ref{tab:ablg1}: on G1 the full method opens a second membership function in \nm{ablg1_full_D10}\% and \nm{ablg1_full_D20}\% of the generations at $D=10$ and 20, and the ratio of its evaluations to the target to those of CUR-G at $D=20$ is \nm{ablg1_full_ratio_D20} [90\% interval \nm{ablg1_full_ci_D20}]. The dip-test variant of Section~\ref{sec:number} trades the two sides: it lowers the false openings on G1 to \nm{ablg1_nodip_D10}\% and \nm{ablg1_nodip_D20}\%, but it costs successes on G2-S (Table~\ref{tab:abl}), most of all at $D=50$ (\nm{abl_dip_D50_text}), because it rejects the unequal and moderately separated groups of the first generations; since a false opening costs almost no evaluations (the ratios above) and a rejected true detection loses the funnel, the test is not part of the method.

\subsection{Calibration of the detector}\label{sec:calib}
The detector is a composition of tests whose joint distribution under the one-component hypothesis is not available in closed form (Section~\ref{sec:number}), so its false-positive and detection rates are measured. Table~\ref{tab:calib} applies it to samples of known distributions with equal weights, as the detector sees them. Under one normal component the rate is \nm{cal_null_10_40}\% at $D=10$, $n=40$ and at most \nm{cal_null_max100}\% at $n=100$; anisotropy up to a condition number of 1000 does not change it, because the detector works in the eigen-coordinates; heavy tails do: a multivariate $t_3$ component is split in \nm{cal_t3_10_40}\% of the samples at $n=40$. Two equal components at a distance of four standard deviations are detected in \nm{cal_d4_10_40}\% of the samples at $n=40$ and \nm{cal_d4_10_100}\% at $n=100$, at five standard deviations in \nm{cal_d5_10_40}\% and \nm{cal_d5_10_100}\%, and at six in nearly all; a minority component with weight 0.1 at five standard deviations is detected in \nm{cal_w01d5_10_40}\% and \nm{cal_w01d5_10_100}\%. The correlated pair of Section~\ref{sec:dir}, invisible on the principal directions, is detected through the two-means direction in \nm{cal_corr_10_100}\% of the samples at $D=10$, $n=100$, and in \nm{cal_corr_20_100}\% at $D=20$. The rates on the actual pipeline, where the selected set is a truncated sample rather than a Gaussian one, are of the same order: on G1 under protocol A the second function opens in \nm{pipe_null_D10}\%, \nm{pipe_null_D20}\%, \nm{pipe_null_D30}\% and \nm{pipe_null_D50}\% of the generations at $D=10$, 20, 30 and 50, \nm{pipe_null_alpha}, and with the condition numbers of Section~\ref{sec:hard} in \nm{pipe_null_cond}. The mixture $0.99\,\mathcal N(0,I)+0.01\,\mathcal N(2.1\,\mathbf e_1,I)$, which has one peak although it has two components, is treated as one component: the second function opens in at most \nm{cal_ce_max40}\% of the samples at $n=40$ and \nm{cal_ce_max377}\% at $n=377$, the selected set of $D=50$, as under one normal component. The density-dip test (last block of Table~\ref{tab:calib}) buys robustness to heavy tails, cutting the false positives under $t_3$ from \nm{cal_t3_10_40}\% to \nm{cal_dip_t3_10_40}\% at $D=10$, $n=40$, but it rejects true detections as well: at $d=4$ the detection rate at $n=100$ falls from \nm{cal_d4_10_100}\% to \nm{cal_dip_d4_10_100}\%, and for the minority component with weight 0.1 at $d=5$ from \nm{cal_w01d5_10_100}\% to \nm{cal_dip_w01d5_10_100}\%. On G2-S at $D=50$ the first admissible detections (the first two generations of ten development runs) have separations of three to five standard deviations, masses of about 0.8/0.2 and a density between the groups of 0.2 to 1.0 of the lower peak, around the threshold $\delta=0.5$: the test rejected the detection in the first generation of every run and in the second generation of some, and where the rejection persisted the funnel was lost (13 of 20 development runs succeeded with the test, 20 of 20 without), which is the loss of Section~\ref{sec:abl} at $D=50$. The selected sets of the test functions are not heavy-tailed, and the final method leaves the test out. The two thresholds trade false openings against detections in the expected direction (Table~\ref{tab:pen}): \nm{pen_summary}

\begin{table}[t]\centering\caption{Calibration of the detector on samples of known distributions ($n$ points in $D$ dimensions, equal weights; 2000 samples for the rates under one component, 500 otherwise): percentage of samples on which a second membership function is opened. One component: standard normal, anisotropic normal with condition number $\kappa$, and multivariate $t_3$. Two components: isotropic normals at distance $d$ (in standard deviations) with equal weights or with minority weight $w$, and the correlated pair of Section~\ref{sec:dir} with equal weights (corr.) or weights $0.9/0.1$ (corr.~0.1). The mixture $0.99\,\mathcal N(0,I)+0.01\,\mathcal N(2.1\,\mathbf e_1,I)$ has one peak and is listed with the one-component cases. The last block repeats the key rows for the variant with the density-dip test of Section~\ref{sec:number} ($\delta=0.5$).}\label{tab:calib}
\footnotesize\setlength{\tabcolsep}{4pt}\begin{tabular}{@{}l cccc@{}}\toprule
 & \multicolumn{2}{c}{$D=10$} & \multicolumn{2}{c}{$D=20$}\\\cmidrule(lr){2-3}\cmidrule(lr){4-5}
Sample & $n=40$ & $n=100$ & $n=40$ & $n=100$\\\midrule
\multicolumn{5}{l}{\emph{One component (false positives)}}\\
normal & 4.4 & 0.2 & 5.0 & 0.8\\
anisotropic, $\kappa=10$ & 4.3 & 0.4 & 4.8 & 1.1\\
anisotropic, $\kappa=100$ & 3.8 & 0.9 & 4.9 & 0.8\\
anisotropic, $\kappa=1000$ & 3.5 & 0.5 & 4.3 & 0.7\\
multivariate $t_3$ & 29.2 & 31.6 & 36.7 & 40.4\\
$0.99/0.01$ at $d=2.1$ (one peak) & 3.9 & 0.7 & 5.1 & 0.4\\
\addlinespace\multicolumn{5}{l}{\emph{Two components (detections)}}\\
equal weights, $d=3$ & 12.8 & 26.6 & 8.2 & 18.8\\
equal weights, $d=4$ & 49.6 & 97.8 & 36.2 & 92.8\\
equal weights, $d=5$ & 94.0 & 100.0 & 87.6 & 100.0\\
equal weights, $d=6$ & 100.0 & 100.0 & 100.0 & 100.0\\
$w=0.2$, $d=4$ & 45.0 & 92.4 & 26.6 & 89.0\\
$w=0.2$, $d=5$ & 84.0 & 100.0 & 78.2 & 100.0\\
$w=0.2$, $d=6$ & 99.2 & 100.0 & 98.8 & 100.0\\
$w=0.1$, $d=4$ & 22.4 & 73.8 & 14.4 & 56.0\\
$w=0.1$, $d=5$ & 58.0 & 97.2 & 44.6 & 94.4\\
$w=0.1$, $d=6$ & 76.4 & 99.6 & 70.0 & 99.2\\
corr. & 34.6 & 66.2 & 10.4 & 45.2\\
corr.~0.1 & 5.4 & 16.8 & 5.2 & 1.8\\
\addlinespace\multicolumn{5}{l}{\emph{With the density-dip test ($\delta=0.5$)}}\\
normal & 1.7 & 0.1 & 2.6 & 0.1\\
multivariate $t_3$ & 11.1 & 11.2 & 15.9 & 16.4\\
$0.99/0.01$ at $d=2.1$ (one peak) & 1.6 & 0.1 & 2.4 & 0.0\\
equal weights, $d=4$ & 38.0 & 47.0 & 28.2 & 40.4\\
equal weights, $d=5$ & 86.6 & 97.0 & 82.6 & 97.8\\
$w=0.1$, $d=4$ & 12.8 & 22.6 & 7.8 & 11.8\\
$w=0.1$, $d=5$ & 40.4 & 69.8 & 31.8 & 58.2\\
corr. & 28.4 & 53.0 & 6.0 & 26.6\\
\bottomrule\end{tabular}\end{table}

\subsection{Rotated, ill-conditioned and rough double funnels}\label{sec:hard}
The 36 settings of Section~\ref{sec:family} and the Lunacek functions are spherical and unrotated. Table~\ref{tab:rot} rotates (\ref{eq:g2}) and makes it ill-conditioned: $f(\vect{x})=\min\{q(\mat{R}_1(\vect{x}-\mu_1\mathbf 1)),\,w_2D+s_2\,q(\mat{R}_2(\vect{x}-\mu_2\mathbf 1))\}$ with $q(\vect{y})=\sum_i\kappa^{(i-1)/(D-1)}y_i^2$, random orthogonal matrices fixed by a seed, and either $\mat{R}_2=\mat{R}_1$ (both funnels elongated along the same axes) or an independent $\mat{R}_2$ (the two funnels elongated along different axes, so that neither the global principal directions nor one local covariance fit both). Table~\ref{tab:rough} adds the Rastrigin-type term $A\sum_i(1-\cos2\pi(x_i-\mu_1))$ of Lunacek's F type with $A\in\{1,10\}$, which fills both funnels with local optima; $A=10$ is the original bi-Rastrigin function and the structure of BBOB $f_{24}$. For the rough functions the table also counts the runs whose final best point lies in the global funnel, because the target of $10^{-8}$ is seldom reached by any algorithm.

\begin{table*}[t]\centering\caption{Rotated and ill-conditioned double funnels (G2-S structure, protocol A, 50 runs): $\kappa$ is the condition number of each funnel, and the second funnel is rotated with the same matrix as the first (same) or with its own (own). Successful runs (ERT in thousands); \ALG{}-ax is \ALG{} restricted to the principal directions.}\label{tab:rot}
\footnotesize\setlength{\tabcolsep}{3pt}\begin{tabular}{lll ccccccc}\toprule
$D$ & $\kappa$ & rotation & CUR-G & \ALG{} & \ALG{}-ax & AMaLGaM & \AMF{} & CMA-ES & BIPOP-CMA-ES\\\midrule
10 & 1 & same & 37 (64.8) & 48 (18.5) & 47 (22.2) & 21 (25.7) & 48 (7.4) & 23 (4.2) & 50 (7.0)\\
10 & 1 & own & 37 (64.8) & 48 (18.5) & 47 (22.2) & 21 (25.7) & 48 (7.4) & 23 (4.2) & 50 (6.8)\\
10 & 10 & same & 23 (190) & 36 (71.3) & 36 (71.3) & 9 (56.5) & 38 (12.1) & 22 (5.0) & 50 (9.9)\\
10 & 10 & own & 22 (205) & 45 (29.4) & 45 (29.8) & 22 (22.3) & 41 (10.6) & 28 (3.7) & 50 (5.1)\\
10 & 100 & same & 11 (547) & 29 (122) & 30 (114) & 7 (78.0) & 24 (21.5) & 20 (6.8) & 50 (14.0)\\
10 & 100 & own & 20 (239) & 32 (98.3) & 32 (98.3) & 30 (16.1) & 34 (14.6) & 27 (4.8) & 50 (7.9)\\
20 & 1 & same & 18 (584) & 50 (49.0) & 50 (49.8) & 0 ($\infty$) & 50 (28.9) & 17 (10.9) & 50 (13.1)\\
20 & 1 & own & 18 (584) & 50 (49.0) & 50 (49.8) & 0 ($\infty$) & 50 (28.9) & 17 (10.9) & 50 (13.1)\\
20 & 10 & same & 0 ($\infty$) & 46 (79.1) & 46 (79.4) & 0 ($\infty$) & 49 (35.4) & 21 (10.1) & 50 (15.5)\\
20 & 10 & own & 0 ($\infty$) & 29 (270) & 28 (289) & 0 ($\infty$) & 17 (117) & 17 (12.8) & 50 (23.6)\\
20 & 100 & same & 2 (7263) & 46 (82.4) & 46 (82.2) & 0 ($\infty$) & 48 (42.7) & 24 (12.2) & 50 (24.2)\\
20 & 100 & own & 0 ($\infty$) & 5 (2767) & 5 (2767) & 0 ($\infty$) & 0 ($\infty$) & 15 (20.5) & 50 (43.2)\\
\bottomrule\end{tabular}\end{table*}

\begin{table*}[t]\centering\caption{Rough double funnels, Lunacek F type with cosine amplitude $A$ (G2-S structure), 50 runs: successful runs at $10^{-8}$ / runs whose final best point lies in the global funnel / median final error.}\label{tab:rough}
\scriptsize\setlength{\tabcolsep}{2.5pt}\begin{tabular}{lll cccccccc}\toprule
$A$ & $D$ & Prot. & CUR-G & \ALG{} & AMaLGaM & \AMF{} & CMA-ES & CAMaLGaM-re & IPOP & BIPOP\\\midrule
1 & 10 & A & 21 / 37 / 0.95 & 19 / 46 / 0.95 & 7 / 17 / 5.1 & 22 / 49 / 0.95 & 8 / 23 / 5.1 & 21 / 36 / 0.95 & 49 / 50 / 8e-11 & 50 / 50 / 8.3e-11\\
1 & 10 & B & 50 / 50 / 8.3e-11 & 43 / 50 / 7.9e-11 & 45 / 50 / 7.7e-11 & 48 / 50 / 7.1e-11 & 49 / 50 / 8.2e-11 & 46 / 50 / 7.9e-11 & 49 / 50 / 8e-11 & 50 / 50 / 8.3e-11\\
1 & 20 & A & 15 / 19 / 10 & 33 / 50 / 9.3e-11 & 0 / 0 / 10 & 25 / 48 / 0.48 & 0 / 12 / 13 & 4 / 15 / 10 & 42 / 45 / 9.3e-11 & 49 / 50 / 8.6e-11\\
1 & 20 & B & 40 / 47 / 9.2e-11 & 43 / 50 / 8.8e-11 & 5 / 9 / 10 & 45 / 50 / 8.8e-11 & 18 / 50 / 0.95 & 12 / 42 / 1.9 & 42 / 45 / 9.3e-11 & 49 / 50 / 8.6e-11\\
10 & 10 & A & 1 / 24 / 7.7 & 0 / 32 / 7.9 & 0 / 9 / 8.5 & 0 / 13 / 8.2 & 0 / 1 / 14 & 0 / 10 / 8.2 & 32 / 50 / 9.7e-11 & 11 / 48 / 0.99\\
10 & 10 & B & 2 / 28 / 3 & 0 / 27 / 3 & 0 / 44 / 3 & 0 / 41 / 3.5 & 0 / 32 / 5 & 0 / 47 / 3 & 32 / 50 / 9.7e-11 & 11 / 48 / 0.99\\
10 & 20 & A & 0 / 18 / 15 & 0 / 42 / 9.9 & 0 / 0 / 16 & 0 / 7 / 10 & 0 / 0 / 39 & 0 / 2 / 16 & 10 / 32 / 3 & 4 / 33 / 4\\
10 & 20 & B & 0 / 19 / 7 & 0 / 41 / 7 & 0 / 2 / 13 & 0 / 15 / 7 & 0 / 0 / 17 & 0 / 7 / 13 & 10 / 32 / 3 & 4 / 33 / 4\\
\bottomrule\end{tabular}\end{table*}

\nm{rot_summary} \nm{rough_summary}

\subsection{Single-structure landscapes: no loss}\label{sec:noloss}
The second side of the claim is tested on G1 and G2-L (Tables~\ref{tab:lungA} and \ref{tab:lungB}, with the paired comparisons in Table~\ref{tab:lungpair}) and on the BBOB suite. On G1 and G2-L under both protocols the success counts of host and component \nm{noloss_counts}, and the geometric mean ratio of the evaluations to the target, component over host, lies between \nm{noloss_gmr_min} and \nm{noloss_gmr_max} with \nm{noloss_inside}; the component and its host draw the same random numbers until the first generation with two membership functions, and on these functions such generations are \nm{noloss_frac2}\% of all generations.

\begin{figure*}[t]
\centering
\includegraphics[width=0.98\textwidth]{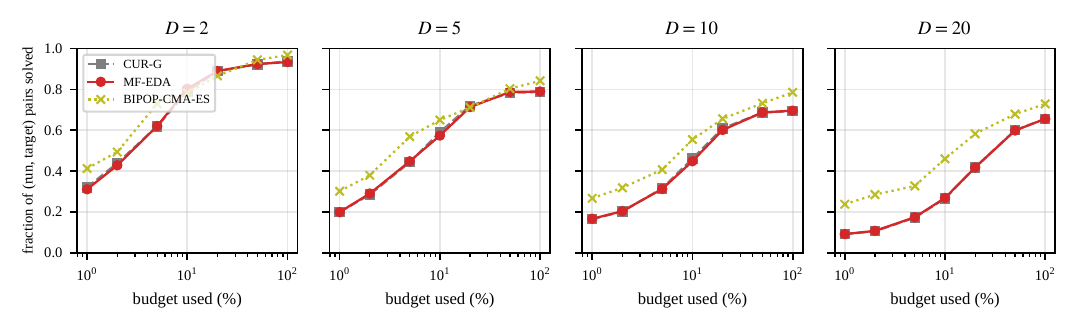}
\caption{BBOB $f_1$--$f_{24}$ (COCO, instances 31--45): fraction of (run, target) pairs solved against the budget used, over the 24 functions, 15 runs and the eight targets $10^{2},10^{1},10^{0},10^{-1},10^{-2},10^{-4},10^{-6},10^{-8}$, for CUR-G, \ALG{} and BIPOP-CMA-ES.}
\label{fig:ecdf}
\end{figure*}

Table~\ref{tab:bbob} covers the 24 BBOB functions at $D\in\{2,5,10,20\}$ with 15 instances each (instances 31--45), through the official COCO code, with CUR-G and \ALG{} on matched initial populations and BIPOP-CMA-ES as reference, and Fig.~\ref{fig:ecdf} shows the fraction of (run, target) pairs solved against the budget, over the eight targets $10^{2},10^{1},10^{0},10^{-1},10^{-2},10^{-4},10^{-6},10^{-8}$. Of the 96 paired comparisons of the final error, \nm{bbob_wins} are significant in favour of \ALG{} and \nm{bbob_losses} against it after Holm correction\nm{bbob_sig_tail}. The success counts are \nm{bbob_mf} for \ALG{}, \nm{bbob_curg} for CUR-G and \nm{bbob_bipop} for BIPOP over the 1440 runs. On $f_{24}$, the rough double funnel, no algorithm reaches the target at $D\ge5$; the median final errors at $D=20$ are \nm{bbob_f24_D20} for CUR-G, \ALG{} and BIPOP. BBOB is a test of no harm on a wide range of structures, not evidence about double funnels, apart from $f_{24}$, which is covered together with the rough functions in Section~\ref{sec:hard}.

\begin{table}[t]\centering\caption{BBOB $f_1$--$f_{24}$ through the official COCO code (instances 31--45, one run per instance, budget $10^4D$), \ALG{} against CUR-G on matched initial populations: successful runs \ALG{}/CUR-G/BIPOP-CMA-ES, and the paired signed-rank test on the final best-so-far error ($+$: \ALG{} smaller, $-$: larger, blank: not significant at $0.05$ after Holm correction over the 96 pairs).}\label{tab:bbob}
\scriptsize\setlength{\tabcolsep}{2pt}\begin{tabular}{l cccc}\toprule
$f$ & $D=2$ & $D=5$ & $D=10$ & $D=20$\\\midrule
$f_{1}$ & 15/15/15 & 15/15/15 & 15/15/15 & 15/15/15\\
$f_{2}$ & 15/15/15 & 15/15/15 & 15/15/15 & 15/15/15\\
$f_{3}$ & 14/14/15 & 1/1/5 & 0/0/0 & 0/0/0\\
$f_{4}$ & 10/10/12 & 0/0/0 & 0/0/0 & 0/0/0\\
$f_{5}$ & 15/15/15 & 15/15/15 & 15/15/15 & 0/0/15\\
\addlinespace
$f_{6}$ & 15/15/15 & 15/15/15 & 15/15/15 & 0/0/15\\
$f_{7}$ & 15/15/15 & 15/15/15 & 14/15/15 & 14/14/15\\
$f_{8}$ & 15/15/15 & 15/15/15 & 13/13/15 & 15/15/15\\
$f_{9}$ & 15/15/15 & 15/15/15 & 15/15/15 & 15/15/15\\
\addlinespace
$f_{10}$ & 15/15/15 & 15/15/15 & 15/15/15 & 15/15/15\\
$f_{11}$ & 15/15/15 & 15/15/15 & 15/15/15 & 15/15/15\\
$f_{12}$ & 13/14/15 & 15/15/15 & 15/15/15 & 15/15/15\\
$f_{13}$ & 15/15/15 & 15/15/15 & 15/15/15 & 15/15/12\\
$f_{14}$ & 15/15/15 & 15/15/15 & 15/14/15 & 15/15/15\\
\addlinespace
$f_{15}$ & 15/15/15 & 4/5/13 & 0/0/6 & 0/0/0\\
$f_{16}$ & 10/14/15 & 9/11/15 & 2/1/11 & 1/1/4\\
$f_{17}$ & 15/15/15 & 15/15/15 & 6/4/15 & 5/5/12\\
$f_{18}$ & 15/14/15 & 5/8/14 & 2/2/14 & 0/2/3\\
$f_{19}$ & 11/12/14 & 1/0/0 & 0/0/0 & 0/0/0\\
\addlinespace
$f_{20}$ & 11/11/15 & 0/1/1 & 0/0/0 & 0/0/0\\
$f_{21}$ & 15/15/14 & 11/11/12 & 4/4/11 & 6/5/10\\
$f_{22}$ & 14/9/15 & 5/3/10 & 0/1/6 & 0/0/1\\
$f_{23}$ & 7/14/15 & 3/3/9 & 0/0/0 & 0/0/1\\
$f_{24}$ & 0/1/0 & 0/0/0 & 0/0/0 & 0/0/0\\
\bottomrule\end{tabular}\end{table}

\subsection{Population size and dimension}\label{sec:popn}
The default population grows with $D$, so that a comparison across dimensions changes the number of selected individuals, the detection sample, at the same time. Table~\ref{tab:popn} fixes $N$ at 115, 285 and 510 for $D=10$, 20 and 30 (the default sizes of the three dimensions) and reruns G2-S and G1 under protocol A. \nm{popn_summary} With $N=115$ at $D=20$ and $N\le285$ at $D=30$ neither algorithm succeeds, and with these sizes neither solves the single funnel G1 (\nm{popn_g1_small}), because a selection of $0.35N$ individuals is then too small to estimate one $D$-dimensional covariance, let alone two; where the population is large enough the component keeps its gain, and the single Gaussian profits from a larger population mainly at $D=10$, where the sample size, not the model, was part of its problem.

\subsection{Limits}\label{sec:limits}
\begin{figure}[t]
\centering
\includegraphics[width=0.9\columnwidth]{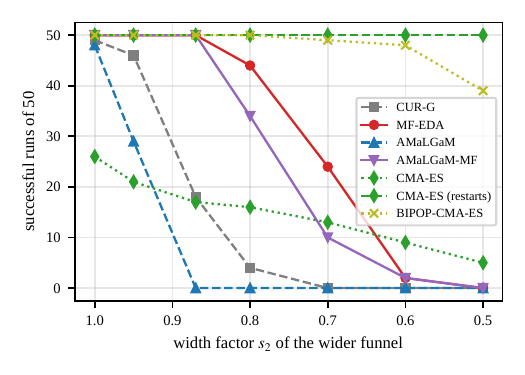}
\caption{Successful runs out of 50 against the width factor $s_2$ of the sub-optimal funnel, $D=20$, protocol A (CMA-ES also with restarts under protocol B, and BIPOP-CMA-ES). $s_2=0.87$ is G2-S.}
\label{fig:sweep}
\end{figure}

Two settings mark the limits of the method. Fig.~\ref{fig:sweep} varies the width factor $s_2$ of the wider funnel on G2-S at $D=20$. As the wider funnel becomes flatter the phase of disadvantage lengthens, and below $s_2=\nm{sweep_knee}$ \ALG{} also fails in most runs (\nm{sweep_mf_06} of 50 at $s_2=0.6$), as do \nm{sweep_others_06}; the restarted CMA-ES variants \nm{sweep_cma_06}. Table~\ref{tab:multi} replaces the two funnels by $K\in\{3,5\}$ random ones, one narrow global funnel and $K-1$ wider funnels lifted by $\delta D$, with centres that are not on one line. Here the component helps little: \nm{multi_curg} against \nm{multi_mf} successes of \nm{multi_runs} runs, a mean risk difference of \nm{multi_rdmean} points [\nm{multi_rdlo}, \nm{multi_rdhi}] across the 36 settings, and \nm{multi_sig} significant settings. A second membership function is opened \nm{multi_open} (in \nm{multi_frac2}\% of the generations on average), so the detector does find two groups; but with three or more funnels that are not on one line, two distributions along one direction cannot represent them, and a model with at most two distributions is the wrong model. Both limits are inherent to the construction, not to the parameters.

\subsection{Computing time and break-even}\label{sec:time}
Table~\ref{tab:time} gives the user CPU time per fitness evaluation of the compared algorithms, which includes the evaluation of the test function and all model computations, and the ratio of \ALG{} to CUR-G, which equals the ratio of their times per generation because both use the same population. \nm{time_summary} Evaluations of the test functions are nearly free, so these times are the cost of the models; they are reported as throughput and do not enter any comparison, which is made at equal numbers of evaluations. The extra time is paid back when the fitness function is expensive. With $t_X$ the model time per evaluation of algorithm $X$ and $c_f$ the cost of one evaluation, the expected time to solve a problem is $\mathrm{ERT}_X\,(t_X+c_f)$, and \ALG{} is faster than CUR-G whenever $c_f$ exceeds
\begin{equation}
c_f^\ast=\frac{\mathrm{ERT}_{\ALG{}}\,t_{\ALG{}}-\mathrm{ERT}_{\mathrm{CUR\text{-}G}}\,t_{\mathrm{CUR\text{-}G}}}{\mathrm{ERT}_{\mathrm{CUR\text{-}G}}-\mathrm{ERT}_{\ALG{}}},
\label{eq:breakeven}
\end{equation}
provided its expected running time is the smaller one. With the expected running times of Table~\ref{tab:lungA} (protocol A), the break-even cost on G2-S is \nm{be_D10}, \nm{be_D20}, \nm{be_D30} and \nm{be_D50} per evaluation at $D=10$, 20, 30 and 50; on G1, where the running times are equal, the component is never faster, and its time overhead of a factor \nm{time_ratio_G1} per generation is the price of the detection.

\begin{table}[t]\centering\caption{Computing time: user CPU time per fitness evaluation in microseconds, including the evaluation of the test function (5 runs per cell with a budget of $3000D$ evaluations on one core of the Linux machine of Section~\ref{sec:setup}, single-threaded, gcc~-O2; one measurement per cell). The ratio is \ALG{} over CUR-G and \AMF{} over AMaLGaM, which have the same population size and therefore the same number of evaluations per generation.}\label{tab:time}
\scriptsize\setlength{\tabcolsep}{2.1pt}\begin{tabular}{@{}lr rrr rrr rr@{}}\toprule
 & & \multicolumn{3}{c}{Rank-weighted EDA} & \multicolumn{3}{c}{AMaLGaM host} & \multicolumn{2}{c}{CMA-ES}\\\cmidrule(lr){3-5}\cmidrule(lr){6-8}\cmidrule(lr){9-10}
 & $D$ & CUR-G & \ALG{} & ratio & AMaLGaM & \AMF{} & ratio & single & BIPOP\\\midrule
G1 & 10 & 0.60 & 12.5 & 20.6 & 0.49 & 12.4 & 25.3 & 5.02 & 0.77\\
 & 20 & 1.25 & 27.7 & 22.2 & 0.88 & 28.3 & 32.1 & 27.7 & 1.02\\
 & 30 & 2.16 & 44.0 & 20.4 & 1.55 & 42.7 & 27.6 & 82.2 & 1.61\\
 & 50 & 4.24 & 75.7 & 17.8 & 2.75 & 74.7 & 27.1 & 327 & 3.32\\
\midrule
G2-S & 10 & 0.54 & 12.3 & 22.9 & 0.44 & 12.2 & 27.7 & 4.80 & 0.30\\
 & 20 & 1.27 & 27.9 & 22.1 & 0.91 & 29.7 & 32.6 & 28.3 & 0.82\\
 & 30 & 2.08 & 42.8 & 20.6 & 1.40 & 42.2 & 30.1 & 82.1 & 1.50\\
 & 50 & 4.26 & 77.5 & 18.2 & 2.89 & 75.1 & 26.0 & 330 & 3.28\\
\bottomrule\end{tabular}\end{table}

\section{Discussion}\label{sec:disc}
\textit{What the component changes.} The component changes one thing in the search distribution: when the selected individuals form two separated groups along one of $D+1$ candidate directions, the two groups are kept as two local Gaussians whose sampling shares follow their widths instead of their head counts. The two-region model of Section~\ref{sec:analysis} says why this matters on a double funnel, the traces of Section~\ref{sec:mech} show the geometric decay of the single Gaussian and the bounded share of the union in real runs, and the ablation of Section~\ref{sec:abl} shows which part of the composition carries the gain: \nm{disc_union}. Elsewhere the component changes little: on the single funnel G1 a second membership function is opened in \nm{pipe_null_D10}\% of the generations at $D=10$ and in \nm{pipe_null_D50}\% at $D=50$, and the evaluations stay within or at the equivalence margin, and on the BBOB suite \nm{disc_bbob}.

\textit{What the propositions do and do not say.} Proposition~\ref{prop:odds} is a statement about an idealised model, not about the trajectories of the algorithm: it keeps the fitted weights, the sampling shares and the coverage of a region apart, and its survival bound holds under a minimum share, a minimum coverage and a selection ratio bounded away from zero, assumptions that the traces support on the Lunacek funnel but that the two-region model cannot guarantee. Proposition~\ref{prop:union} is exact for the sampler: the union is its density, $1/(1+\omega)$ its acceptance rate, and the width share is the label share up to twice the overlap; it says nothing about where the mass of a membership function falls on the landscape, which is the coverage and is measured. The detector is a composition of tests on a non-regular model whose error rates are calibrated by simulation, not derived.

\textit{Limits.} Section~\ref{sec:limits} has shown where the component stops helping: a much narrower global funnel ($s_2\le\nm{sweep_knee}$) and more than two funnels that are not on one line. \nm{disc_hard} Restarted CMA-ES solves the smooth double funnels with fewer evaluations than any EDA tested, because its single runs are cheap; the component keeps the deeper funnel within one run and one population, which is a different mechanism and does not replace restarts. The comparison with CAMaLGaM rests on our re-implementation and says nothing about the original method. Heavy-tailed samples raise the false-positive rate of the detector considerably (\nm{cal_t3_lo}\% to \nm{cal_t3_hi}\% for the multivariate $t_3$ samples of Table~\ref{tab:calib}), and the detection of a minority group of a tenth of the selection needs a separation of five to six standard deviations.

\textit{Computing time.} The component costs no fitness evaluations but costs computing time: a generation takes \nm{disc_time_auto} times as long as a generation of the baseline at $D=10$ to 50 (Section~\ref{sec:time}), which the operation count of Section~\ref{sec:cost} attributes mainly to the $D+1$ one-dimensional fits. By (\ref{eq:breakeven}) the extra time is paid back on G2-S when an evaluation costs more than \nm{be_D20} at $D=20$; on a single funnel it is never paid back. In absolute terms the model adds tens of microseconds per evaluation (Table~\ref{tab:time}), which is negligible when an evaluation costs milliseconds or more, as in most applications, and dominates the run time on cheap test functions.

\textit{Future work.} Three directions follow. First, the detector can be made cheaper by testing the two-means direction and the first few principal directions only, which the ablation suggests loses little; and the EM fits can share their starting points across generations. Second, membership functions on two or three directions, with the same admissibility tests, would describe more than two funnels; they will need more individuals per component, and the minimum-share variant of Section~\ref{sec:abl} is the natural way to protect a small group among several. Third, the component can be inserted into a faster single-Gaussian method, such as CMA-ES with restarts, to combine cheap runs with width-based shares.

\section{Conclusion}\label{sec:conc}
This paper has presented \ALG{}, a Gaussian estimation-of-distribution algorithm that estimates the distribution of the selected individuals as a composition of one or two distributions: in every generation the selected individuals are decomposed by normal membership functions on a data-chosen direction, each membership function is joined to a local Gaussian, and new solutions are generated from their union, so that the share of samples a funnel receives follows the width of its membership function and not the number of selected individuals in it. A two-region model with explicit assumptions explains why a model whose shares follow the head count loses the deeper funnel and gives a survival bound for the composition; the sampler has a closed-form density, an exact acceptance rate and label shares within twice the overlap of the width shares; the detector is calibrated by simulation. On seeds not used in development and on matched initial populations, the component raised the successful runs of its baseline from \nm{fam_curg} to \nm{fam_mf} of \nm{fam_runs} on 36 double-funnel settings, and on the Lunacek G2-S funnel at $D=50$ from \nm{A_G2S_D50_curg} to \nm{A_G2S_D50_mf} of 50 for the rank-weighted EDA and from \nm{A_G2S_D50_ama} to \nm{A_G2S_D50_amf} for AMaLGaM, with \nm{noloss_conc}, and \nm{conc_bbob} on the BBOB suite. The ablation shows that the gain over the single Gaussian comes from keeping the second group as its own local Gaussian with a share that does not follow its head count: composing the same membership functions with the head-count weights of a mixture loses successes and evaluations where the disadvantage of the deeper funnel lasts, whereas the union, equal shares and a width-weighted sum perform alike; the limits are a much narrower global funnel, more than two funnels off one line, the evaluation efficiency of restarted CMA-ES on smooth double funnels, and a longer computing time per generation that is paid back only when evaluations are expensive.
\appendices
\section*{Supplementary Material}
The following appendices give the constants of the method, the paired comparisons on the Lunacek functions, the cost of the component on a single funnel, the sensitivity to the thresholds, the runs with the population size fixed across dimensions, the multi-funnel landscapes and the reproducibility details. All tables are generated by the released scripts from the per-run result files.

\section{Constants of the method}\label{supp:const}
Table~\ref{tab:const} lists every constant of \ALG{} and of the hosts, with the value used in all experiments and the experiment in which it is varied, if any.

\begin{table*}[t]\centering\caption{Constants of \ALG{} and of its hosts, with the value used in all experiments and the experiment in which the constant is varied.}\label{tab:const}
\footnotesize\begin{tabular}{@{}>{\raggedright\arraybackslash}p{1.1cm}>{\raggedright\arraybackslash}p{5.2cm}>{\raggedright\arraybackslash}p{7.6cm}>{\raggedright\arraybackslash}p{2.6cm}@{}}\toprule
Symbol & Meaning & Value & Varied in\\\midrule
$N$ & population size & $\max\{115,\lfloor17+3D^{1.5}+\tfrac12\rfloor\}$ & Section~\ref{sec:popn}\\
$\tau$ & selection fraction & 0.35 & --\\
$\alpha$ & variance multiplier (AVS) & $[1,10]$, factor $0.9$ & --\\
$c$ & allowance factor, threshold $c\ln K$ & 2 & Table~\ref{tab:pen}\\
$K$ & number of candidate directions & $D+1$ & Table~\ref{tab:abl}\\
-- & separation factor (\ref{eq:open2}) & 2 & --\\
-- & minimum effective mass (\ref{eq:open3}) & 2 & --\\
$\delta$ & dip threshold of the dip-test variant (not part of the method) & 0.5 & Tables~\ref{tab:abl}, \ref{tab:pen}\\
-- & dip bandwidth of that variant & $0.9\,\bar s\,(n_{\mathrm{eff}}/2)^{-1/5}$, floor $0.3\,\mathrm{sd}$ & --\\
-- & EM tolerance, maximum iterations & $10^{-7}$ relative, 100 & Table~\ref{tab:abl}\\
-- & EM starts & quartiles; reruns from minority starts $(0.05,0.55)$, $(0.45,0.95)$ with weights $0.1/0.9$ when the first solution beats the allowance but is not separated & Table~\ref{tab:abl} (single start)\\
-- & variance floors & $10^{-6}$ relative $+10^{-12}\lambda_1$ & --\\
$\gamma_j$ & shrinkage of the local covariance & $D/(n_j+D)$ & Table~\ref{tab:abl}\\
$\alpha_{\mathrm e}$ & cap of the multiplier (\ref{eq:cap}) & $\max\{1,(|c_1-c_2|/2\bar s)^2\}$ & Table~\ref{tab:abl}\\
-- & minimum share (variant) & $1/4$ & Table~\ref{tab:abl}\\
-- & two-means iterations & $\le10$ & --\\
\bottomrule\end{tabular}\end{table*}

\section{Paired comparisons on the Lunacek functions}
Table~\ref{tab:lungpair} gives, for every cell of Tables~\ref{tab:lungA} and~\ref{tab:lungB}, the paired comparison of each host with its component: the $2\times2$ table of successes, the exact McNemar test, the risk difference with its Agresti--Min interval, and the geometric mean ratio of the evaluations to the target with its 90\% interval, which is the equivalence test of Section~\ref{sec:noloss}.
\begin{table*}[t]\centering\caption{Paired comparisons of the component with its host on the Lunacek funnels (50 matched runs per cell): risk difference in percentage points [95\% interval], ratio of the expected running times host/component, and the geometric mean ratio of the evaluations to the target, component/host, over the runs in which both succeeded [90\% interval]. $^\dagger$: exact McNemar test significant at $0.05$ after Holm correction over the 24 cells of the family. A dash: no ratio, because one of the two never reached the target (ERT ratio) or fewer than five runs succeeded for both (evaluation ratio).}\label{tab:lungpair}
\footnotesize\setlength{\tabcolsep}{3pt}\begin{tabular}{lll ccc ccc}\toprule
 & & & \multicolumn{3}{c}{\ALG{} vs CUR-G} & \multicolumn{3}{c}{\AMF{} vs AMaLGaM}\\\cmidrule(lr){4-6}\cmidrule(lr){7-9}
Prot. & Function & $D$ & $\Delta$ succ. & ERT ratio & eval. ratio & $\Delta$ succ. & ERT ratio & eval. ratio\\\midrule
A & G1 & 10 & +0 [-4, 4] & 0.99 & 1.01 [0.97, 1.05] & +0 [-4, 4] & 0.96 & 1.04 [1.00, 1.09]\\
A & G1 & 20 & +0 [-4, 4] & 0.99 & 1.01 [0.98, 1.05] & +0 [-4, 4] & 0.96 & 1.04 [1.00, 1.07]\\
A & G1 & 30 & +0 [-4, 4] & 1.01 & 0.99 [0.97, 1.01] & +0 [-4, 4] & 1.01 & 1.00 [0.97, 1.02]\\
A & G1 & 50 & +0 [-4, 4] & 0.98 & 1.02 [1.00, 1.03] & +0 [-4, 4] & 0.99 & 1.01 [1.00, 1.03]\\
A & G2-L & 10 & +2 [-3, 7] & 1.23 & 1.02 [0.97, 1.07] & +0 [-4, 4] & 0.98 & 1.02 [0.98, 1.06]\\
A & G2-L & 20 & +0 [-4, 4] & 0.97 & 1.03 [1.00, 1.06] & +0 [-4, 4] & 0.90 & 1.11 [1.06, 1.17]\\
A & G2-L & 30 & +0 [-4, 4] & 0.98 & 1.02 [0.99, 1.04] & +0 [-4, 4] & 0.93 & 1.07 [1.02, 1.13]\\
A & G2-L & 50 & +0 [-4, 4] & 0.96 & 1.05 [1.02, 1.07] & +0 [-4, 4] & 0.91 & 1.11 [1.06, 1.15]\\
A & G2-S & 10 & +22 [9, 33]$^\dagger$ & 3.50 & 1.00 [0.95, 1.06] & +54 [37, 67]$^\dagger$ & 3.49 & 0.78 [0.63, 0.96]\\
A & G2-S & 20 & +64 [48, 75]$^\dagger$ & 11.93 & 0.94 [0.88, 1.00] & +100 [90, 103]$^\dagger$ & -- & --\\
A & G2-S & 30 & +88 [74, 95]$^\dagger$ & 29.24 & 1.05 [0.98, 1.12] & +98 [87, 102]$^\dagger$ & -- & --\\
A & G2-S & 50 & +98 [87, 102]$^\dagger$ & -- & -- & +98 [87, 102]$^\dagger$ & -- & --\\
B & G1 & 10 & +0 [-4, 4] & 0.99 & 1.01 [0.97, 1.05] & +0 [-4, 4] & 0.96 & 1.04 [1.00, 1.09]\\
B & G1 & 20 & +0 [-4, 4] & 0.99 & 1.01 [0.98, 1.05] & +0 [-4, 4] & 0.96 & 1.04 [1.00, 1.07]\\
B & G1 & 30 & +0 [-4, 4] & 1.01 & 0.99 [0.97, 1.01] & +0 [-4, 4] & 1.01 & 1.00 [0.97, 1.02]\\
B & G1 & 50 & +0 [-4, 4] & 0.98 & 1.02 [1.00, 1.03] & +0 [-4, 4] & 0.99 & 1.01 [1.00, 1.03]\\
B & G2-L & 10 & +0 [-4, 4] & 1.01 & 1.00 [0.95, 1.06] & +0 [-4, 4] & 0.98 & 1.02 [0.98, 1.06]\\
B & G2-L & 20 & +0 [-4, 4] & 0.97 & 1.03 [1.00, 1.06] & +0 [-4, 4] & 0.90 & 1.11 [1.06, 1.17]\\
B & G2-L & 30 & +0 [-4, 4] & 0.98 & 1.02 [0.99, 1.04] & +0 [-4, 4] & 0.93 & 1.07 [1.02, 1.13]\\
B & G2-L & 50 & +0 [-4, 4] & 0.96 & 1.05 [1.02, 1.07] & +0 [-4, 4] & 0.91 & 1.11 [1.06, 1.15]\\
B & G2-S & 10 & +0 [-4, 4] & 1.39 & 0.80 [0.72, 0.89] & +0 [-4, 4] & 2.72 & 0.46 [0.39, 0.55]\\
B & G2-S & 20 & +8 [-0, 16] & 2.92 & 0.49 [0.42, 0.56] & +58 [42, 70]$^\dagger$ & 19.47 & 0.22 [0.18, 0.26]\\
B & G2-S & 30 & +74 [58, 84]$^\dagger$ & 12.97 & 0.59 [0.45, 0.79] & +98 [87, 102]$^\dagger$ & 326.27 & --\\
B & G2-S & 50 & +98 [87, 102]$^\dagger$ & -- & -- & +100 [90, 103]$^\dagger$ & -- & --\\
\bottomrule\end{tabular}\end{table*}

\section{Cost of the component on a single funnel}
Table~\ref{tab:ablg1} reports, on the single funnel G1, the share of generations in which the full method and its ablated variants open a second membership function (false openings) and the ratio of their evaluations to the target to those of CUR-G, with the 90\% interval used for the equivalence margin.
\begin{table*}[t]\centering\caption{Cost of false positives on the single funnel G1, protocol A: share of generations with two membership functions (\%), and the geometric mean ratio of the evaluations to the target against CUR-G over the paired runs [90\% interval].}\label{tab:ablg1}
\footnotesize\setlength{\tabcolsep}{3pt}\begin{tabular}{l ccc ccc}\toprule
 & \multicolumn{3}{c}{two-MF generations (\%)} & \multicolumn{3}{c}{evaluation ratio vs CUR-G}\\\cmidrule(lr){2-4}\cmidrule(lr){5-7}
Variant & $D=10$ & $D=20$ & $D=30$ & $D=10$ & $D=20$ & $D=30$\\\midrule
\ALG{} (full) & 9.6 & 2.9 & 1.1 & 1.012 [0.972, 1.053] & 1.013 [0.981, 1.046] & 0.990 [0.975, 1.006]\\
principal directions only & 8.3 & 2.6 & 1.0 & 1.027 [0.992, 1.064] & 1.001 [0.969, 1.034] & 0.986 [0.970, 1.001]\\
with density-dip test ($\delta=0.5$) & 4.3 & 0.2 & 0.1 & 1.033 [0.993, 1.076] & 1.002 [0.991, 1.013] & 1.001 [0.997, 1.004]\\
rank weights in the detector & 7.5 & 1.7 & 1.0 & 1.020 [0.972, 1.071] & 0.997 [0.969, 1.024] & 0.982 [0.968, 0.996]\\
diagonal local Gaussians & 9.1 & 2.8 & 1.1 & 1.055 [1.012, 1.099] & 0.997 [0.969, 1.025] & 0.986 [0.965, 1.008]\\
\bottomrule\end{tabular}\end{table*}

\section{Sensitivity to the thresholds}
Table~\ref{tab:pen} varies the allowance factor $c$ of (\ref{eq:open1}) and, with the dip-test variant switched on, its threshold $\delta$, one at a time with everything else at its default, and reports the successes on G2-S and the false openings on G1; Section~\ref{sec:calib} summarises it.
\begin{table}[t]\centering\caption{Sensitivity to the allowance factor $c$ (threshold $c\ln K$ on the BIC gain), and the dip-test variant with threshold $\delta$ (off = the method as proposed, without the dip test), protocol A, 50 runs ($c=2$ in all other experiments). G2-S: successful runs (ERT in thousands); G1: share of generations with two membership functions (\%).}\label{tab:pen}
\footnotesize\setlength{\tabcolsep}{3pt}\begin{tabular}{ll ccc ccc}\toprule
 & & \multicolumn{3}{c}{G2-S: successes (ERT)} & \multicolumn{3}{c}{G1: two-MF generations (\%)}\\\cmidrule(lr){3-5}\cmidrule(lr){6-8}
 & & $D=10$ & $D=20$ & $D=30$ & $D=10$ & $D=20$ & $D=30$\\\midrule
$c$ & 0 & 46 (25.0) & 50 (48.1) & 50 (114) & 42.2 & 24.4 & 14.8\\
 & 1 & 46 (25.4) & 50 (50.1) & 50 (116) & 21.1 & 8.2 & 3.9\\
 & 2 & 48 (18.5) & 50 (49.0) & 50 (117) & 9.6 & 2.9 & 1.1\\
 & 3 & 48 (18.2) & 49 (55.4) & 50 (115) & 4.6 & 0.8 & 0.2\\
 & 4 & 47 (22.0) & 48 (61.9) & 49 (124) & 1.9 & 0.2 & 0.1\\
\addlinespace
$\delta$ & off & 48 (18.5) & 50 (49.0) & 50 (117) & 9.6 & 2.9 & 1.1\\
 & 0.3 & 45 (29.2) & 47 (70.1) & 42 (203) & 1.4 & 0.1 & 0.1\\
 & 0.4 & 46 (25.3) & 48 (63.0) & 43 (189) & 2.5 & 0.1 & 0.1\\
 & 0.5 & 47 (21.9) & 48 (62.9) & 47 (146) & 4.3 & 0.2 & 0.1\\
 & 0.6 & 47 (21.7) & 48 (62.3) & 47 (145) & 6.1 & 0.4 & 0.1\\
 & 0.7 & 47 (21.8) & 48 (62.6) & 49 (126) & 7.3 & 1.0 & 0.1\\
\addlinespace
\bottomrule\end{tabular}\end{table}

\section{Population size fixed across dimensions}
Table~\ref{tab:popn} reruns G2-S and G1 with the population size fixed at the default of each dimension in turn, so that the number of selected individuals, the detection sample, is the same across dimensions; Section~\ref{sec:popn} summarises it.
\begin{table}[t]\centering\caption{Population size $N$ fixed across dimensions (the default sizes of \ALG{}, $N=\max\{115,\lfloor17+3D^{1.5}+\tfrac12\rfloor\}$, are 115, 285 and 510 at $D=10$, 20 and 30), protocol A, 50 paired runs: G2-S successful runs CUR-G / \ALG{} (ERT of \ALG{} in thousands); G1: two-MF generations of \ALG{} (\%).}\label{tab:popn}
\scriptsize\setlength{\tabcolsep}{2.5pt}\begin{tabular}{@{}l ccc ccc@{}}\toprule
 & \multicolumn{3}{c}{G2-S: CUR-G / \ALG{} (ERT)} & \multicolumn{3}{c}{G1: two-MF generations (\%)}\\\cmidrule(lr){2-4}\cmidrule(lr){5-7}
$N$ & $D=10$ & $D=20$ & $D=30$ & $D=10$ & $D=20$ & $D=30$\\\midrule
115 & 37 / 48 (18.5) & 0 / 0 ($\infty$) & 0 / 0 ($\infty$) & 9.6 & 13.2 & 48.2\\
285 & 50 / 50 (30.3) & 18 / 50 (49.0) & 0 / 0 ($\infty$) & 3.8 & 2.9 & 3.0\\
510 & 50 / 50 (53.8) & 29 / 50 (92.7) & 6 / 50 (117) & 3.3 & 1.5 & 1.1\\
\bottomrule\end{tabular}\end{table}

\section{Multi-funnel landscapes}
Table~\ref{tab:multi} gives the paired comparison of CUR-G and \ALG{} on the 36 settings with three and five funnels of Section~\ref{sec:limits}; the generator of the centres is given in Appendix~\ref{supp:repro}.
\begin{table}[t]\centering\caption{Random multi-funnel landscapes with $K$ funnels (one narrow global funnel and $K-1$ wider ones lifted by $\delta D$; three layouts $L$ of the centres per $K$), budget $10^4D$, 25 paired runs per setting: successful runs CUR-G / \ALG{}. $^\dagger$: exact McNemar significant at $0.05$ after Holm correction over the 36 settings.}\label{tab:multi}
\small\setlength{\tabcolsep}{3pt}\begin{tabular}{ccc cc}\toprule
$K$ & $\delta$ & $L$ & $D=10$ & $D=20$\\\midrule
3 & 0.6 & 0 & 20\,/\,17 & 4\,/\,7\\
3 & 0.6 & 1 & 11\,/\,12 & 19\,/\,19\\
3 & 0.6 & 2 & 25\,/\,25 & 13\,/\,15\\
3 & 0.7 & 0 & 20\,/\,20 & 10\,/\,12\\
3 & 0.7 & 1 & 14\,/\,14 & 22\,/\,21\\
3 & 0.7 & 2 & 25\,/\,25 & 21\,/\,21\\
3 & 0.8 & 0 & 25\,/\,24 & 12\,/\,13\\
3 & 0.8 & 1 & 18\,/\,18 & 25\,/\,24\\
3 & 0.8 & 2 & 25\,/\,25 & 21\,/\,22\\
5 & 0.6 & 0 & 0\,/\,1 & 13\,/\,17\\
5 & 0.6 & 1 & 11\,/\,8 & 2\,/\,4\\
5 & 0.6 & 2 & 21\,/\,20 & 4\,/\,7\\
5 & 0.7 & 0 & 3\,/\,4 & 13\,/\,17\\
5 & 0.7 & 1 & 13\,/\,14 & 7\,/\,9\\
5 & 0.7 & 2 & 22\,/\,22 & 8\,/\,13\\
5 & 0.8 & 0 & 6\,/\,8 & 19\,/\,21\\
5 & 0.8 & 1 & 13\,/\,14 & 11\,/\,13\\
5 & 0.8 & 2 & 25\,/\,25 & 14\,/\,15\\
\midrule total & & & \multicolumn{2}{c}{535\,/\,566 of 900}\\
\bottomrule\end{tabular}\end{table}

\section{Reproducibility}\label{supp:repro}
\emph{Programs.} \texttt{mfeda3.c} (CUR-G and \ALG{}), \texttt{cmp\_algos.c} (AMaLGaM, \AMF{}, CMA-ES), \texttt{cma\_pub.c} (public CMA-ES with IPOP and BIPOP), \texttt{camalgam\_re.c}, the component \texttt{mfcomp.h}, the test functions \texttt{funnel\_fn.h}, the amalgamated COCO sources, the parallel job runner \texttt{runjobs.c}, the completeness check \texttt{checkdone.c}, the detector test \texttt{det\_test.c} and the statistics program \texttt{stats3.c}; all C99, compiled with \texttt{-std=c99 -O2}; and the C++ driver \texttt{drv\_funnel.cpp} that runs the authors' HillVallEA code. The job lists (911 jobs: the main list, the twelve allocation-ablation jobs of Table~\ref{tab:ablx}, the twelve ablation variants at $D=50$ and the 162 calibration calls), the expected line counts of every result file and the SHA-256 fingerprints of the executables that produced the results are part of the release. The BBOB result files come from a second build of \texttt{mfeda3} and \texttt{cma\_pub} that differs from the first only in the instance list passed to the COCO suite (instances 1--45 instead of 1--30, which does not contain the instances 31--45 used here); the fingerprints of both builds are listed.

\emph{Seeds.} A run of \texttt{mfeda3} with function number $f$ ($f=5$ for the Lunacek G type, $f=4$ for the multi-funnel function, $f=1$ for the double-funnel family; the COCO function $f_i$ is numbered $100+i$ and uses $i$), dimension $D$ and run index $r$ uses the seed $20261118+1000f+100D+r$ for a splitmix64 generator, from which all random numbers of the run are drawn, so that CUR-G and \ALG{} with the same $r$ start from the same population; \texttt{cmp\_algos} uses $777000+10000\ell+1000\lfloor10\sigma_0\rfloor+100D+r$ for every algorithm ($\ell=0$ for the default $\lambda$ of CMA-ES, $\sigma_0=2$), so that AMaLGaM, \AMF{} and the CMA-ES re-implementation with the same $r$ share the initial draws; \texttt{cma\_pub} uses the seed formula of \texttt{mfeda3}; \texttt{camalgam\_re} uses $777000+1000D+r$ (the state of the generator is the seed itself; see the correction in Section~\ref{sec:setup}). The run indices are $r=5000,\dots,5049$ (Lunacek, rotated, rough, swept, multi-funnel, ablation, sensitivity and fixed-population runs) and $r=6000,\dots,6049$ (double-funnel family); the BBOB run $r=0,\dots,14$ uses instance $r+31$. Multi-funnel centres $\vect{c}_0,\dots,\vect{c}_{K-1}$ are drawn uniformly in $[-3,3]^D$ from the seed $777+1000D+K+100000L$ for layout $L\in\{0,1,2\}$, and the function is $f(\vect{x})=\min\{\|\vect{x}-\vect{c}_0\|^2,\;\min_{k\ge1}(\delta D+s\|\vect{x}-\vect{c}_k\|^2)\}$ with the BBOB width factor $s=1-1/(2\sqrt{D+20}-8.2)$ and $\delta\in\{0.6,0.7,0.8\}$.

\emph{Per-run files.} Each line holds the algorithm, the function, $D$, the run index, the seed, success, the evaluations at the first hit of $10^{-8}$, the evaluations used, the final error, the mean number of membership functions, the share of generations with two functions, the CPU time, the best-so-far error at $1\%$, $2\%$, $5\%$, $10\%$, $20\%$, $50\%$ and $100\%$ of the budget, the evaluations at which the targets $10^{2},10^{1},10^{0},10^{-1},10^{-2},10^{-4},10^{-6},10^{-8}$ were first reached ($-1$ if never), and for \texttt{mfeda3} the fraction of two-means directions among the openings, the mean acceptance rate, the mean overlap $\omega$, the mean shrinkage, the number of two-function generations and the branch of (\ref{eq:g2}) at the final best point.\section*{Acknowledgment}
The original idea of this work was proposed by Professor ZHANG Jun. Under Professor Zhang's direction, an AI system, Claude (Anthropic), was used to implement the algorithms and the experiments, and help complete the manuscript. The other authors and students reviewed and proofread the code, the results and the text. The authors take full responsibility for the content.

\end{document}